\documentclass[12pt]{article}
\usepackage{amsmath}
\usepackage{times}
\usepackage{graphicx}
\usepackage{color}
\usepackage{multirow}
\usepackage{subcaption}
\usepackage{caption}
\usepackage[export]{adjustbox}

\usepackage[backend=biber,style=apa,uniquename=false]{biblatex}
\usepackage{amsfonts}

\usepackage{rotating}
\usepackage{bbm}
\usepackage{latexsym}
\usepackage{booktabs}
\usepackage{float}

\def\*#1{\mathbf{#1}}
\newcommand{\captionfonts}{\normalsize}
\newcommand{\E}{{\mathbb E}}

\makeatletter 
\long\def\@makecaption#1#2{%
  \vskip\abovecaptionskip
  \sbox\@tempboxa{{\captionfonts #1: #2}}%
  \ifdim \wd\@tempboxa >\hsize
    {\captionfonts #1: #2\par}
  \else
    \hbox to\hsize{\hfil\box\@tempboxa\hfil}%
  \fi
  \vskip\belowcaptionskip}
\makeatother 
\begin{document}
%\hspace{13.9cm}1

\vspace*{20mm}

\noindent {\LARGE Hierarchical Active Inference using Successor\\Representations}\\[12pt]
\noindent {\bf \large Prashant Rangarajan$^{1}$, Rajesh P. N. Rao$^{1,*}$}\\[4pt]
{$^{\displaystyle 1}$Paul G. Allen Center for Computer Science and Engineering, University of Washington, Seattle, USA}\\
{$^{\displaystyle *}$Corresponding author}\\[8pt]

\thispagestyle{empty}
%\markboth{}{NC instructions}
%
%\ \vspace{-0mm}\\
%
\vspace{5mm}
%Abstract
\begin{center} {\bf Abstract} \end{center}
Active inference, a neurally-inspired model for inferring actions based on the free energy principle (FEP), has been proposed as a unifying framework for understanding perception, action, and learning in the brain. Active inference has previously been used to model ecologically important tasks such as navigation and planning, but scaling it to solve complex large-scale problems in real-world environments has remained a challenge. Inspired by the existence of multi-scale hierarchical representations in the brain, we propose a model for planning of actions based on hierarchical active inference. Our approach combines a hierarchical model of the environment with successor representations for efficient planning. We present results demonstrating (1) how lower-level successor representations can be used to learn higher-level abstract states, (2) how planning based on active inference at the lower-level can be used to bootstrap and learn higher-level abstract actions, and (3) how these learned higher-level abstract states and actions can facilitate efficient planning. We illustrate the performance of the approach on several planning and reinforcement learning (RL) problems including a variant of the well-known four rooms task, a key-based navigation task, a partially observable planning problem, the Mountain Car problem, and PointMaze, a family of navigation tasks with continuous state and action spaces. Our results represent, to our knowledge, the first application of learned hierarchical state and action abstractions to active inference in FEP-based theories of brain function.
%%%%%%%%%%%

\section{Introduction}
Active Inference is a biologically-inspired theory for explaining how agents make decisions in dynamic environments \parencite{friston2015active}: an agent is modeled as being constantly engaged in a perception-action loop, with actions that can modify the state of the agent and the environment, causing new sensory observations to be received. In order to maintain a form of homeostasis of its internal states in relation to its environment, the agent is hypothesized to minimize its variational free energy, consistent with the free energy principle (FEP) \parencite{friston2010free}. In this view, the agent can be thought of as an inference engine constantly minimizing the ``surprise'' of its sensory observations using approximate Bayesian inference to infer not only the hidden sensory state of the environment, but also actions that serve to minimize surprise, and hence, the free energy. This inference process includes a bias \parencite{sharot2011optimism} towards the agent's preferences or prior beliefs. 

The FEP, through active inference, can thus be regarded as extending brain theories such as predictive coding \parencite{Rao-Ballard1999, huang2011predictive} and other Bayesian brain models \parencite{Rao_Olshausen_ProbabilisticModels_2002,knill2004bayesian, Doya_etal_BayesianBrain_2011,aitchison2017or} to action inference. The FEP treats action and perception as two different means of achieving the same objective: minimizing variational free energy to reduce uncertainty about the world.

Active inference algorithms use planning methods to choose actions that minimize the expected free energy of future observations with respect to the agent's preferred goal. In a discrete setting \parencite{da2020active}, the agent chooses among a set of actions to find the best sequence of actions that minimizes expected free energy, combining an intrinsic component (capturing epistemic value or expected information gain) and an extrinsic component (capturing instrumental value or expected utility, in terms of prior preferences or goals).

Active inference thus offers a natural solution to the exploration-exploitation problem in reinforcement learning \parencite{friston2015active,fountas2020deep}. However, the elegance of active inference comes at a cost: it is hard to scale active inference to large-scale environments. Here we address this problem by proposing a hierarchical version of active inference based on successor representations \parencite{dayan1993improving}. We provide several examples of its applicability to hierarchical planning problems.

\begin{figure}[!htbp]
\begin{minipage}[t]{\textwidth}
    \centering
    \begin{subfigure}{0.45\textwidth}
    \vspace{0pt}
        \includegraphics[width=\textwidth]{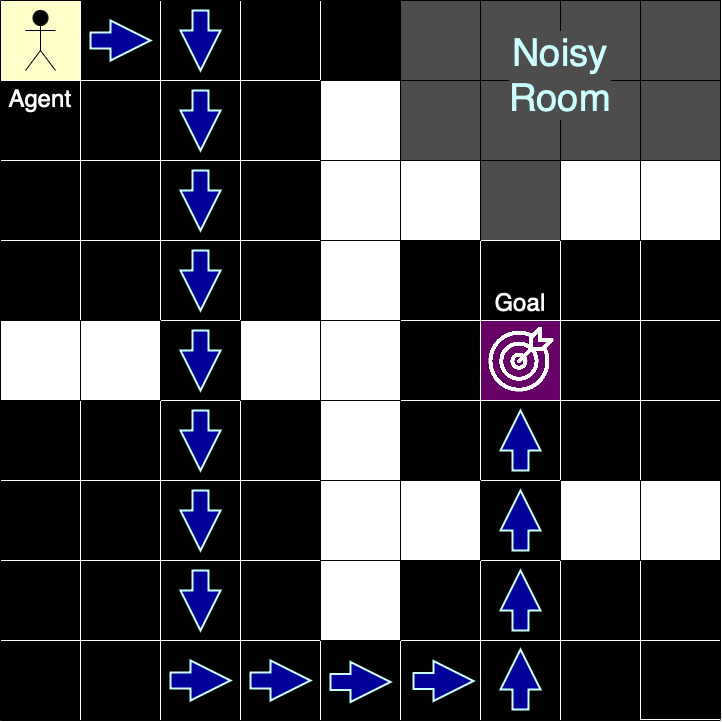} 
        \caption{}
        \label{fig:gridworld_schema}
    \end{subfigure}
    \hfill
    \begin{subfigure}{0.45\textwidth}
    \vspace{0pt}
        \includegraphics[width=\textwidth]{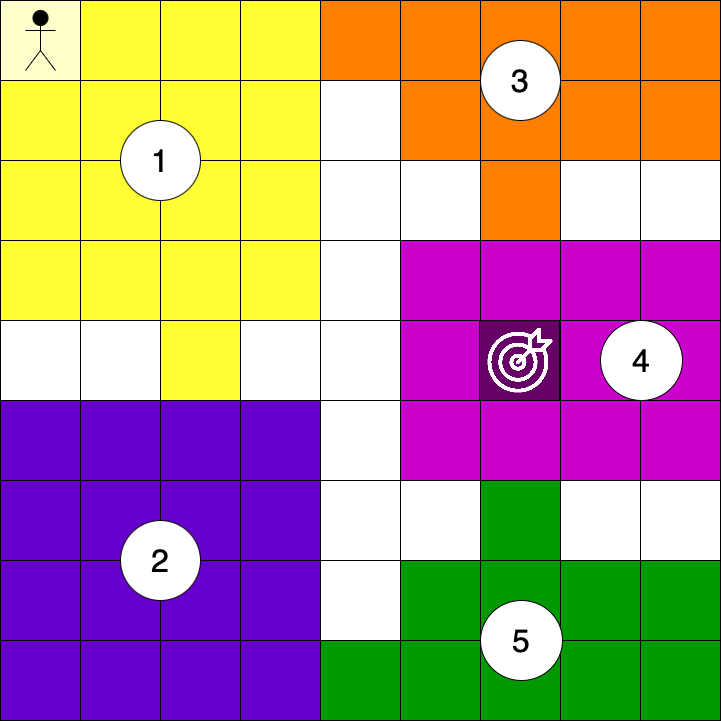} 
        \caption{}
        \label{fig:macro_state_schema}
    \end{subfigure}
    \begin{subfigure}{0.45\textwidth}
    \vspace{0pt}
        \includegraphics[width=\textwidth]{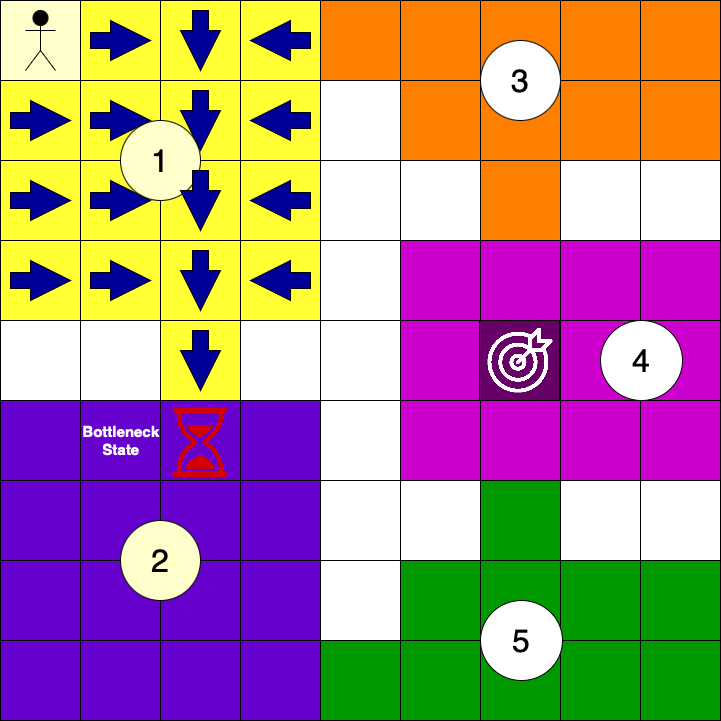}
        \caption{}
        \label{fig:macro_action_schema}
    \end{subfigure}
    \hfill
    \begin{subfigure}{0.45\textwidth}
    \vspace{0pt}
    \centering
        \includegraphics[width=\textwidth]{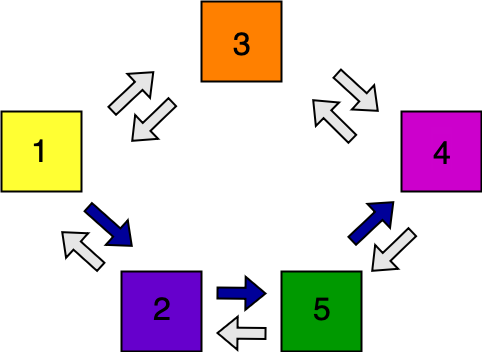}
        \caption{} 
        \label{fig:macro_plan_schema}
    \end{subfigure}
    \caption{{\bf Hierarchical Active Inference using Successor Representations.}\\(continued next page)}
\end{minipage}
\end{figure}
\addtocounter{figure}{-1}
\begin{figure}[!htbp]
\caption{ (continued)\\
    (a) A $9\times 9$ Gridworld with five rooms (separated by white walls with openings). The agent must navigate from the start state (top left) to the goal state (purple target symbol) to get a reward. In this example, one of the rooms has noisy observations. Active inference in this case would prefer an alternate (longer but less noisy) path to the goal.
    (b) By clustering of the successor matrix, the states of this Gridworld can be grouped into $5$ macro states (numbered) corresponding to the five rooms in the environment.
    (c) Example of a macro action learned by the model that corresponds to a policy for navigating from macro state $1$ to macro state $2$. The arrows represent the best action to take from each state in macro state $1$ to get to a bottleneck state in macro state $2$ (end point represented by the red hourglass).
    (d) Higher level planning. The arrows represent possible macro actions that the agent can take from one macro state to another. The blue colored arrows represent the optimal path chosen by applying active inference at the higher level to reach the room containing the goal location.}
\label{fig:schema}
\end{figure}

Figure~\ref{fig:schema} illustrates our proposed hierarchical active inference method based on successor representation using a simple ``Gridworld'' environment.
The central idea of our approach is to use a state and action hierarchy to reach a goal more efficiently. The state-action hierarchy is learned by first clustering the successor matrix to obtain ``macro states'' and then learning ``macro actions'' to navigate between these macro states. The macro actions allow the agent to plan at a higher level of abstraction, thereby enabling it to reach a goal more efficiently. The successor matrix is also used directly for planning at the lower level (instead of more traditional active inference-based planning), allowing the agent to significantly cut down the computational cost. 

\section{Related Work}
Friston and colleagues have explored how active inference can be applied to deep hierarchical Bayesian models for goal-directed behavior \parencite{friston2018deep,pezzulo2018hierarchical}. Related work has focused on potential neural implementations of such mechanisms, for example, in hippocampal-prefrontal circuits \parencite{proietti2023active, van2023bridging}. In these models, hierarchy emerges from an efficient factorization of the underlying Bayesian model. In contrast, our approach induces a hierarchy of states and actions from the successor representation.

Hierarchical planning has been extensively studied in the field of reinforcement learning (RL). In hierarchical reinforcement learning (HRL) \parencite{Pateria2021,Dayan1992}, macro actions called ``options'' \parencite{sutton1999between} have been proposed as temporally extended actions that enable efficient learning and planning in complex environments. In model-based RL, deep neural networks have been used to break down complex tasks efficiently for hierarchical planning \parencite{hafner2022deephierarchicalplanningpixels}. Other researchers in RL have explored the use of state abstraction for solving Markov Decision Processes (e.g., \cite{li2006towards}). Our model builds on these past approaches to leverage both state and action hierarchies for efficient planning.

Our approach uses the successor representation \parencite{dayan1993improving}, which has been proposed as a compromise between the computational inefficiencies of learning full-fledged world models for model-based RL and the inflexibility of model-free RL. The successor representation has also been used for modeling human behavior \parencite{momennejad2017successor}. Correspondences have been identified between place cells in the hippocampus and the successor representation \parencite{stachenfeld2017hippocampus, gershman2018successor, momennejad2020learning, stoewer2022neural} (see also \cite{whittington2020tolman}). 
In the context of active inference, the concept of inductive inference \parencite{Friston-et-al2025} bears some similarities with the successor representation: a known (or learned) series of transition matrices is collapsed over the action-dependent dimension of the transition tensor to provide a description of plausible transitions, which is used to implement a form of backwards induction. Other work has explored combining the successor representation with active inference \parencite{millidge2023active} to reduce computational cost, albeit without hierarchical states or actions. 

\section{Methods}
\subsection{Active Inference}
{\large {\it Notation and Formulation}}\\
In this work, we assume that an agent does not have complete visibility of the state of the world. At each time step, the agent uses the observations it gathers from sensors to refine its internal estimate of the state of the world. It also executes actions at each time step to actively learn more about the state and reduce its uncertainty about the environment. Specifically, we model the problem using a Partially Observable Markov Decision Process (POMDP), where $s_t$, $a_t$, and $o_t$ represent the hidden state, action and observation (or outcome), respectively, at time $t$ for $1\leq t\leq T$. Here $T$ represents the planning horizon, that is, the number of future time steps the agent considers for planning. At each time $t$, the agent maintains a posterior probability distribution (``belief state'') over the entire space of possible states as an estimate of its current state. We denote $b_t$ as this belief state of the agent at time $t$. Although the underlying process is Markovian, the belief state accumulates information from all past observations to compute the posterior probability of state. Note that a Markov Decision Process (MDP), which forms the basis for traditional RL, can be regarded as a special case of the POMDP where the true state of the environment is observed at each time step (i.e., no noise in the observations). The Bayesian generative model of the agent is a joint distribution of states, observations, and actions over all time steps, represented by $p_\theta(o_{1:T},s_{1:T}, a_{1:T-1})$ where $\theta$ represents the parameters.

%\vspace*{\baselineskip}\\
Using Bayesian inference to determine the exact posterior probability distribution of states and actions under a full generative model as above may be intractable in many settings. In such cases, active inference employs variational inference to estimate an approximate posterior by minimizing the variational free energy. Specifically, a variational distribution $q_\phi(\cdot)$ (parameterized by $\phi$) is used to approximate the posterior (note however that for the simple POMDP task we explore in this paper (Section~\ref{sec:task3}), the posterior can be computed exactly and we therefore do not need $\phi$).

The {\em surprisal} of an observation, given by $-\log p_\theta(o)$, quantifies how unlikely an observation is under a given generative model. Any agent that desires to remain in equilibrium (homeostasis) with its environment must minimize its surprisal over its set of observations. In accordance with the Free Energy Principle \parencite{friston2006free}, the agent minimizes its (variational) free energy as a proxy in order to remain in homeostasis. %can be used for perception and learning, i.e. to infer both the states as well as update any priors such as the likelihood. Let us assume that the agent has run for a while and the current time step is $t$. 
The variational free energy (VFE) at time step $t$ is defined as:
\begin{equation}
    F_\pi(t) = \mathbb{E}_{q_\phi(s_t|\pi)}[\log q_\phi(s_t|\pi) - \log p_\theta(s_t,o_t|\pi)] \label{eq:vfe-def}
%= D_{KL}(q_\phi(s_t|\pi)|| p_\theta(s_t,o_t|\pi))
\end{equation}
where $\pi$ denotes a ``policy,'' defined as a sequence of actions given by a plan (note that this definition of a policy differs from the definition of policy in reinforcement learning \cite{sutton-barto-book} as a state-to-action mapping).

Expanding the second log term in Equation~\ref{eq:vfe-def}, VFE can also be written as:
\begin{eqnarray}
        F_\pi(t) &=& -\log p_\theta(o_t) + \mathbb{E}_{q_\phi(s_t|\pi)}[\log q_\phi(s_t|\pi) - \log p_\theta(s_t|o_t,\pi)]\nonumber\\
    &=& -\log p_\theta(o_t) + D_{KL}(q_\phi(s_t|\pi)|| p_\theta(s_t|o_t,\pi))\nonumber\\
    &\geq& -\log p_\theta(o_t)\nonumber
\end{eqnarray}
%\[ \]
%\geq -\log p_\theta(o_t) \]
%\[F_\pi(t)   \]
where $D_{KL}$ denotes the Kullback-Leibler divergence. As evident above, minimizing VFE is equivalent to minimizing an upper bound on the surprisal of the agent (this is similar to maximizing the evidence lower bound (ELBO) in variational autoencoders \parencite{kingma2022autoencodingvariationalbayes}).

It is useful to consider an alternative way of expressing VFE, obtained by rearranging the terms in Equation~\ref{eq:vfe-def} as follows:
\begin{eqnarray}
    F_\pi(t) &=& \mathbb{E}_{q_\phi(s_t|\pi)}[\log q_\phi(s_t|\pi) - \log p_\theta(o_t,s_t|\pi)] \label{eq:vfe-def1}\\
    &=& \mathbb{E}_{q_\phi(s_t|\pi)}[\log q_\phi(s_t|\pi) - \log (p_\theta(o_t|s_t,\pi)p_\theta(s_t|\pi))] \nonumber\\
    &=& \mathbb{E}_{q_\phi(s_t|\pi)}[\log q_\phi(s_t|\pi) - \log p_\theta(o_t|s_t) - \log p_\theta(s_t|\pi)] \nonumber\\
    &=&-\underbrace{\E_{q_\phi(s_t|\pi)}[\log p_\theta(o_t|s_t)]}_{accuracy} + \underbrace{D_{KL}(q_\phi(s_t|\pi)||p_\theta(s_t|\pi))}_{complexity}
    \label{eq:vfe}
\end{eqnarray}
Thus, VFE can be viewed as quantifying the trade-off between accuracy and model complexity. Specifically, the first term in Equation~\ref{eq:vfe}, accuracy, measures how well the agent's model of the world predicts the sensory observations. The second term, representing the complexity of the model, quantifies how close the posterior state distribution remains to the prior. Thus, minimizing VFE can be understood as maximizing the accuracy of one's predictions of observed outcomes while at the same time trying not to increase the complexity of the model.

The VFE as defined above depends only on the current state and observation. In planning, the agent selects an optimal sequence of future actions which requires us to estimate the free energy for future actions and observations.
%minimizes uncertainty. 
This entails a formulation of free energy known as expected free energy (EFE), which incorporates future observations and actions, for determining the optimal policy \parencite{millidge2020deep}. 

\noindent The EFE at time $\tau$ ($t \leq \tau \leq T$) is defined as:
\begin{equation}
    G_\pi(\tau)= \mathbb{E}_{q_\phi(o_\tau, s_\tau|\pi)}[\log q_\phi(s_\tau|\pi) - \log p_\theta(o_\tau,s_\tau|\pi)]\label{eq:efe-def}
\end{equation}
The structure of EFE is very similar to VFE (compare the above equation to Equation~\ref{eq:vfe-def1}). However, unlike VFE, EFE is used for planning and the observations occur at future time steps. The observation $o_\tau$ at a future time step $\tau$ is therefore treated as a random variable, and the expectation operation in Equation~\ref{eq:efe-def} includes future observations in addition to future states. Active inference picks actions that minimize the EFE in the future. 
We can simplify the right side of Equation~\ref{eq:efe-def} further and get:
\begin{eqnarray}
        G_\pi(\tau) &=& \mathbb{E}_{q_\phi(o_\tau,s_\tau|\pi)}[\log q_\phi(s_\tau|\pi) - \log p_\theta(s_\tau|o_\tau,\pi)] - \E_{q_\phi(o_\tau|\pi)}\log p_\theta(o_\tau|\pi)] \nonumber \\
    G_\pi(\tau) &\approx& -\underbrace{\mathbb{E}_{q_\phi(o_\tau,s_\tau|\pi)}[\log q_\phi(s_\tau|o_\tau,\pi) - \log q_\phi(s_\tau|\pi)]}_{expected~information~gain 
} - \underbrace{\E_{q_\phi(o_\tau|\pi)}[\log \tilde{p}_\theta(o_\tau)}_{utility}]\label{eq:efe-tradeoff}
\end{eqnarray}
The first term in Equation~\ref{eq:efe-tradeoff} is the (negative) epistemic value, which is the expected information gain or reduction in uncertainty when conditioned on a future observation. This represents the information seeking (or intrinsic) motivation of the agent. The second term in Equation~\ref{eq:efe-tradeoff} is the pragmatic value, which is the expected utility or score of an observation. As discussed earlier, active inference selects actions that take the agent towards observations with high priors. In Equation~\ref{eq:efe-tradeoff} above, we use $\tilde{p}$ to represent the agent's prior probability distribution over sensory observations (instead of a probability distribution conditioned on policy $\pi$): this allows the EFE to capture the agent's prior biases for particular observations as being ``desirable'' (e.g., locations containing food or goal locations in navigation). Thus, the second term in the EFE equation above represents the reward seeking (or extrinsic) motivation of the agent to seek outcomes that match its prior preferences. 

Clearly the agent would like to minimize its uncertainty about the world through information gathering (first term in Equation~\ref{eq:efe-tradeoff}), but it would also like to maximize its reward (second term in Equation~\ref{eq:efe-tradeoff}). Thus, the EFE as defined in Equation~\ref{eq:efe-tradeoff} naturally captures the tradeoff between exploration and exploitation, a topic of considerable interest in reinforcement learning \parencite{sutton-barto-book}.

A more convenient formulation of EFE for the purpose of computation is as follows:
\begin{equation}
    G_\pi(\tau)= \underbrace{D_{KL}[q_\phi(o_\tau|\pi)||\tilde{p}_\theta(o_\tau)]}_{risk} + \underbrace{\E_{q_\phi(s_\tau|\pi)}[H(p_\theta(o_\tau|s_\tau))}_{ambiguity}]
    \label{eq:efe_risk}
\end{equation}
where $H$ denotes entropy. In the equation above, the first term represents the risk or expected cost/complexity, i.e., the KL-divergence between the predicted outcomes dictated by a policy $\pi$ and the preferred outcomes dictated by the agent's prior $\tilde{p}_\theta(o_\tau)$ capturing its preferences or goals. The lower this value, the higher the chance of achieving a rewarding outcome based on the policy $\pi$. 
The second term represents the ambiguity or the expected entropy of the likelihood distribution. The lower this value, the more precise are the agent's predictions about its observations given its beliefs about the true state of the world. Policies that direct agents to states that provide precise observations about the world give a better picture about the true underlying state, and thus are desirable. 
%This is the form that will be used in the methods section.
 
The path integral (or summation) of the EFE over all future time steps, given policy 
$\pi$, starting from the current time step $t$ and going up to the horizon $T$, is given by:
\begin{equation}
   G_\pi = \sum_{\tau=t}^T G_\pi(\tau) \label{eq:G-pi-sum} 
\end{equation}

By minimizing $G_\pi$ over the space of all possible policies, the agent can identify the optimal sequence of actions to execute. Alternatively, rather than using a greedy approach where the agent selects the policy or action sequence that minimizes $G_\pi$, the agent can estimate the probability distribution over the policy space and sample a policy from it instead.
 Let the distribution over the policies be $Q$. Then, we can use the EFE values to compute the probability of a policy $\pi$ as: $Q(\pi) = \sigma(-G_\pi)$ where $\sigma$ is the softmax function $\sigma(x)=\frac{exp(x)}{\sum_x exp(x)}$. 
 
 Note that there are exponentially many such policies (as a function of the planning horizon $T$), making exact minimization computationally challenging for long horizons. One way to address this issue is to evaluate $G_\pi$ over a shorter time horizon. Based on this evaluation, the agent can sample an action according to the marginal distribution $Q(a) = \sum_{\pi:a_t = a}Q(\pi)$, representing the probability of the action $a$ being executed at the current time step $t$ and the summation is over all policies (action sequences) beginning with $a$. After executing the sampled action, the agent can recompute $G_\pi$ once again and take the next action as described above, and repeat this process until the goal state is reached (or the episode times out).

\vspace*{.2in}
{\large {\it Bayesian Model for a Discrete POMDP}}\\
Consider the case of a POMDP which has the joint probability distribution $p_\theta(o_{1:T},s_{1:T}, a_{1:T-1})$ with discrete states, actions and observations. 
Let us consider a simple case with $n_s$ states, $n_o$ observations, and $n_a$ actions:
\begin{align*}
\mathcal{S}&=\{s^{(1)},s^{(2)},\ldots, s^{(n_s)}\},\\
\mathcal{O}&=\{o^{(1)},o^{(2)},\ldots, o^{(n_o)}\},\\
\mathcal{A}&=\{a^{(1)},a^{(2)},\ldots, a^{(n_a)}\}.
\end{align*}
We assume that the state and observation of an agent at time $t$ are represented by one-hot vectors, where $s_t\in \mathbb{R}^{n_s\times 1}$ and $o_t\in \mathbb{R}^{n_o\times 1}$.

Due to the underlying process being Markovian, we can factorize the joint distribution as follows:
\begin{equation}
    p_\theta(o_{1:T},s_{1:T}, a_{1:T-1}) = p_\theta(s_1)p_\theta(a_{1:T-1})\left(\prod_{t=1}^Tp_\theta(o_t|s_t)\right) \left(\prod_{t=2}^{T}p_\theta(s_t|s_{t-1},a_{t-1})\right)\label{eq:joint-factorized}
\end{equation}
This Bayesian model is visualized in Figure~\ref{fig:graph_schema}.

\vspace*{.2in}
\noindent {\large {\it Parameters of a Discrete POMDP and Parameter Learning}}\\
The POMDP in Equation~\ref{eq:joint-factorized} is defined by the set of parameters $\theta$ which includes the following components (Figure~\ref{fig:graph_schema}) which can be set a priori or learned:
\begin{itemize}
    \item The observation model or likelihood $p_\theta(o_t|s_t)$, which captures the underlying relationship between the observations (sensory measurements) and hidden states of the agent. It can be represented by a matrix $A\in \mathbb{R}^{n_o\times n_s}$, where each column represents the probabilities of different possible observations given a particular state. We assume that this probability distribution remains identical for all time steps and across all policies. In the case where there are multiple modalities, a separate $A$ matrix can be maintained for each modality. 
    %For instance, in a simple gridworld, the observation space might include two modalities, one indicating the location of the agent, and another signaling whether the agent has reached the goal (a reward observation). Both modalities in this case would have an individual $A$ matrix. 
  If the $A$ matrix is an identity matrix, then the model simplifies to an MDP, where the observation space and state space are identical. This is the case of our Tasks 1 and 2 below (Sections~\ref{sec:task1} and \ref{sec:task2}). For the POMDP Gridworld in Task 3 (Section~\ref{sec:task3}) in which the observations are noisy versions of the true state, we assume this noise (and hence $A$) is known. 
    \item The state-action transition probability distribution $p_\theta(s_{t}|s_{t-1},a_{t-1})$, which encodes the Markovian dynamics of the model. It can be represented by $B\in \mathbb{R}^{n_s\times n_s\times n_a}$, a tensor which contains a state transition matrix for each action. We can also denote $B_a$ as the $n_s\times n_s$ state transition matrix corresponding to action $a$. For the results reported in Section~\ref{sec:results}, we use a simple count-based learning procedure for learning $B$: for each occurrence of the triple $(s_t=s_i,s_{t-1}=s_j,a_{t-1}=a_k)$, we increment $B(i,j,k)$ by 1 (all entries of $B$ are initialized to zero before the triples are obtained from interacting with the environment). Note that the learning of $B$ could be formalized using Dirichlet distributions (e.g., \cite{Friston-dirichlet-2015}), but for simplicity, we use the count-based procedure above for our demonstrations in Section~\ref{sec:results}. 

\begin{figure}[!htbp]
\begin{minipage}[t]{\textwidth}
        \centering
        \includegraphics[width=\textwidth]{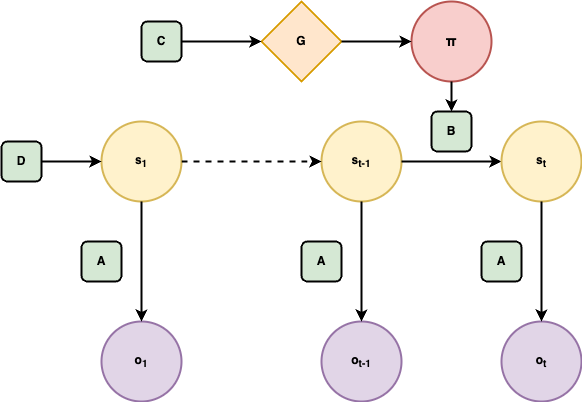} 
        \caption{{\bf Generative model for a Discrete POMDP.} The notation used here follows prior active inference literature (e.g., \cite{smith2022step}). The EFE $G$ depends on the prior $C$ and generates the probabilities of different policies $\pi$. The policy $\pi$ being a sequence of actions in turn affects the state transition matrix $B$. The observations are generated from the state using the observation model determined by the matrix $A$.}
        \label{fig:graph_schema}
\end{minipage}
\end{figure}

    \item Prior over observations, $\tilde{p}_\theta(o_t)$, represented by $C\in \mathbb{R}^{n_o\times 1}$. The agent seeks observations that have higher prior probability, i.e., are considered more rewarding. We assume this is known to the agent and therefore $C$ is assumed to be known for the experiments in Section~\ref{sec:results}.
    \item Prior over initial state $p_\theta(s_1)$, which can be denoted by the vector $D\in \mathbb{R}^{n_s\times 1}$. Since the state is encoded as a one-hot vector, the actual prior probability for a particular initial state $s_1$ is given by $p_\theta(s_1) = s_1^\top D$. $D$ is often assumed to be a uniform distribution over 
    $\mathcal{S}$ if the agent has no other prior information. We make this assumption for the results in Section~\ref{sec:results}.
    \item Prior over policies $p_\theta(\pi)$ or $p_\theta(a_{1:T-1})$. This is represented by a vector $E$ over the space of policies with the elements of the vector corresponding to the prior probability for each policy. Based on past actions or habits of the agent, the prior probability could be higher for certain policies. However, for the present work (and the results in Section~\ref{sec:results}), we assume a uniform prior over policies, leading to a constant factor for this term in the factorization in Equation~\ref{eq:joint-factorized}. 
\end{itemize}

Using the vector and matrix notation above, the factorization of the joint distribution in Equation~\ref{eq:joint-factorized} can be rewritten as:
\[p_\theta(o_{1:T},s_{1:T}, a_{1:T-1}) \propto (s_1^\top D)\left(\prod_{t=1}^To_t^\top As_t\right) \left(\prod_{t=2}^{T}s_t^\top B_{a_{t-1}} s_{t-1}\right)\]

To infer the agent's hidden state in a POMDP environment, the agent can estimate a belief state $b_t\in \mathbb{R}^{n_s\times 1}$, which represents the posterior distribution over its possible states at time $t$, as:
\begin{equation}
    b_t = \sigma\left(\ln A^\top o_t + \ln B_{a_{t-1}} b_{t-1}\right)
\end{equation}
where the functions $\sigma$ (the softmax function) and $\ln$ are applied to each element of their vector arguments individually. Note that in this case of a discrete POMDP where the exact posterior can be computed as above, we do not need an approximating distribution $q$ or separate parameters $\phi$ introduced in the {\em Notation and Formulation} section above. 

For planning using active inference, the EFE in Equations~\ref{eq:efe_risk} and \ref{eq:G-pi-sum} can be rewritten in vector-matrix notation as:
\begin{eqnarray}
    G_\pi(\tau) &=& \underbrace{D_{KL}[q_\phi(o_\tau|\pi)||\tilde{p}_\theta(o_\tau)]}_{risk} + \underbrace{\E_{q_\phi(s_\tau|\pi)}[H(p_\theta(o_\tau|s_\tau))}_{ambiguity}]\\
     &=& \left( \left(As_{\tau}\right)^\top \left( \ln As_{\tau} - \ln C_\tau \right) \right) 
- \operatorname{diag} \left( A^\top \ln A \right) ^\top s_{\tau}\nonumber\\
G_\pi &=& \sum_{\tau} \left( \left(As_{\tau}\right)^\top \left( \ln As_{\tau} - \ln C_\tau \right) \right) 
- \operatorname{diag} \left( A^\top \ln A \right)^\top s_{\tau}\nonumber
\label{eq:efe}
\end{eqnarray}
As discussed above, the agent selects a policy according to the probability distribution $Q(\pi) = \sigma(-G_\pi)$ where $\sigma$ is the softmax function. 
%Assuming the number of policies is $n_\pi$, all of the $G_\pi$ values can be represented using a vector $G\in\mathbb{R}^{n_\pi \times 1}$. The agent can then select a policy according to the distribution $\sigma(-G)$ where $\sigma$ is applie to each element of the vector 

\subsection{Successor Representation}
Active inference requires a search over a sequence of future actions (for computing EFE as above) and as a result, does not scale well as the time horizon $T$ increases. For problems that involve choosing actions to reach a desired goal state, the scaling problem for active inference can be ameliorated if we have an estimate of how close a particular state is to the goal state, in terms of reachability as defined by the environment's transition function. This notion of reachability can be quantified using the successor representation \parencite{dayan1993improving}, defined as a matrix $M$ whose entries $M(s,s')$ represent the expected future occupancy of a state $s'$ starting from the state $s$.
\vspace*{\baselineskip}\\
The successor representation was originally proposed in reinforcement learning as a way of quickly estimating the value function, which represents the expected cumulative future reward for any given state. Reinforcement learning algorithms rely on the value function for computing the policy of the agent: the best action to take in any state is the one that will result in a state with the highest value. The problem arises when the goal or reward structure of the environment changes. In such a case, the agent needs to relearn the value function from scratch. 
\vspace*{\baselineskip}\\
A more efficient approach is to factor the dynamics of the environment from the goals, and estimate the value function on the fly when given a new goal or task. Specifically, the value function can be decomposed into a product of the successor representation (the dynamics) and the reward function (goals). Thus, even if the reward structure changes, one can use a previously learned successor representation to quickly estimate the new value function and consequently the new policy.
\vspace*{\baselineskip}\\
Let $s_t$ and $r_t$ represent the state and reward obtained by the agent at time $t$. Also let $\gamma$ be a discount factor. Let $\rho$ be a state-to-action mapping (or ``policy'' as defined in the reinforcement learning sense \cite{sutton-barto-book}). Note that $\rho$ contrasts with the sequence-of-actions policy $\pi$ used above for active inference: while reinforcement learning is based on the Bellman optimality principle and aims to optimize a function of states (the optimal policy)
 for maximizing expected future reward, active inference aims to optimize a functional of beliefs about states to evaluate a sequence of actions. Because beliefs about states can change during inference and learning, there is no unique state-to-action policy in active inference. The equivalent (Bellman) recursion in belief space entails a recursive estimation of EFE \parencite{Friston-et-al2021} that we circumvent using the successor representation \parencite{dayan1993improving}.

Given a state-to-action policy $\rho$, the value $v_\rho(s)$ of a state $s$ is defined as:
\begin{equation}
    v_\rho(s) = \E_\rho\left[\sum_{k=0}^{\infty}\gamma^k r_{t+k} \mid s_t = s \right]
\end{equation}
We assume the reward function $r(\cdot)$ only depends on the current state $s$. Also, let $p(s'|s,a)$ be the transition probability from state $s$ to $s'$ using action $a$. Then, the value function can be defined recursively with the following ``Bellman equation'' \parencite{sutton-barto-book}:
\begin{equation}
    v_\rho(s) = r(s) + \sum_a \rho(a|s)\sum_{s'}\gamma p(s'|s,a)v_\rho(s') \label{eq:bellman-sum}
\end{equation}

The successor representation, which represents the expected cumulative discounted future occupancy of state $s'$ from state $s$, is defined as:
\begin{equation}
    M_\rho(s,s') = \E_\rho\left[\sum_{k=0}^{\infty}\gamma^k \delta(s_{t+k},s') \mid s_t = s \right]\label{eq:successor-sum}
\end{equation}
where $\delta$ is the Kronecker delta function equaling 1 if the two arguments are equal and zero otherwise.

Consider the case of a discrete MDP. Let $\*v\in \mathbb{R}^{n_s}$ be a vectorized form of the value function with entries $v_\rho(s)$. Let $T \in \mathbb{R}^{n_s\times n_s}$ be the one-step state transition matrix for policy $\rho$ whose entries $T_\rho(s,s')$ represent the probability $p(s'|s)=\sum_a \rho(a|s)p(s'|s,a)$. Let $\*r\in \mathbb{R}^{n_s}$ be the reward function vector with entries $r(s)$. The Bellman equation (Equation~\ref{eq:bellman-sum}) can be written in vector form as:
\begin{eqnarray}
    \*v &=&\*r +\gamma T\*v\nonumber\\
    &=& \*r +\gamma T[\*r +\gamma T[\*r +\gamma T[\cdots]]]\nonumber\\
    &=& [I +\gamma T +\gamma^2 T^2+\cdots]\*r \label{eq:bellman} 
\end{eqnarray}
where $I$ is the identity matrix. The successor matrix $M \in \mathbb{R}^{n_s\times n_s}$, whose entries are $M_\rho(s,s')$ (see Equation~\ref{eq:successor-sum}), can be written as:
\begin{equation}
    M = I +\gamma T +\gamma^2 T^2+\cdots = \sum_{k=0}^{\infty}\gamma^k T^k = (I - \gamma T)^{-1}
    \label{eq:succ}
\end{equation}
Substituting $M$ for the summation in Equation~\ref{eq:bellman}, we have:
\begin{equation}
    \*v = M\*r
\end{equation}
This equation can be written in non-vector form as: 
\begin{equation}
    v_\rho(s)=\sum_{s'}M_\rho(s,s')r(s')
    \label{eq:val_succ}
\end{equation}

Suppose the discrete states are numbered $1,\ldots,N$ and we are given a trajectory of states obtained by executing a policy $\rho$, for example, a uniformly random exploration policy. If the state at time step $t$ in the trajectory is $s_t = i$, then the successor matrix can be learned using the temporal difference learning rule \parencite{dayan1993improving}:
\begin{equation}
M_\rho(i,j) = M_\rho(i,j) + \alpha(\delta(s_{t+1},j) + \gamma M_\rho(s_{t+1},j) - M_\rho(i,j))
\label{eq:TD-successor-learning}
\end{equation}
where $\alpha$ is the learning rate.
Alternatively in vectorized form, for a state transition from $s$ to $s'$, each row $s$ of $M$ can be updated as follows:
\begin{equation}
M(s,:) = M(s,:) + \alpha(\mathbf{1}_{s'} + \gamma M(s',:) - M(s,:))
\label{eq:TD-successor-learning-vector}
\end{equation}
where $\mathbf{1}_{s'}$ is a one-hot vector with value 1 at the position corresponding to state $s'$.

If the state transition probability matrix $T$ is known, either through learning by interacting with the environment or by prior knowledge, we can use Equation~\ref{eq:succ} to calculate the successor matrix directly as $M = (I - \gamma T)^{-1}$ \parencite{russek2017predictive}.

$T$, $M$ and $\*v$ depend on the policy $\rho$ used to explore the environment. To get a good estimate of the true value function, we need to have a policy that facilitates sufficient exploration of the state space. A simple choice is a policy that selects an action uniformly at random from $\mathcal{A}$. For such a policy, we can obtain a default state transition matrix:
\[\tilde{T} = \frac{1}{n_a}\sum_{a\in\mathcal{A}} B_a\]
where $B_a$ is the transition matrix for action $a$. This yields the following estimate of the successor matrix: 
\begin{equation}
\tilde{M} = (I - \gamma\tilde{T})^{-1}
\end{equation}

In a POMDP, the agent lacks direct access to the hidden states and instead relies on observations obtained through interactions with the environment. Computing the successor matrix directly is therefore not possible. However, as discussed earlier, the agent can use inference to compute belief states, which are posterior probability estimates of the hidden state. While these belief states $b_t$ are distributions over the entire state space, computing the successor representation requires a single state estimate $\hat{s}_t$ to allow us to update the appropriate row of the successor matrix. To obtain $\hat{s}_t$, we can either sample from $b_t$ or select the state with the highest posterior probability. The successor matrix can thus be computed as a function of state estimates derived from belief states.
% In the case of a POMDP, instead of using a state-action value function, we can use an observation value function:
% \[\nu(o)= \E_{q(s|o)}[\nu(s)] = qM\mathcal{G}^\pi\] where $q$ is the categorical variational posterior.
\vspace*{\baselineskip}\\
%Now that we have established how the successor matrix can be computed, let us understand how it can be used in conjunction with free energy computations in order to plan more efficiently.
One of the main drawbacks of computing the EFE for all possible sequences of actions $\pi$ is that there are an exponential number of such sequences. However, following \parencite{millidge2023active}, we can emulate Equation~\ref{eq:val_succ} for computing the value function from the successor representation, and define the expected free energy function $\nu(s)$ as follows:
\begin{equation}
    \nu(s) = \sum_{s'}M(s,s')G(s') \label{eq:nu-equation}
\end{equation}
where $G(s')$ is the one-step EFE evaluated at state $s'$. Note the similarity to Equation~\ref{eq:val_succ}, with the free energy $G(s')$ replacing the reward function $r(s')$. For planning, the agent can use these $\nu$ values for the next states $s'$ to generate an optimal sequence of actions.

In summary, the agent first computes the successor matrix, either directly using random exploration and temporal difference learning, or using the transition matrix. Next, given the matrices $A$, $B$, and $C$, the agent computes at each time step the $\nu$ values for the states that can be reached from a particular state when different actions are executed. Finally the agent can determine the action to take from a given state at each time step by picking the action that leads to the state with the highest $\nu$ value. Chaining together these actions produces the optimum policy $\pi^*$.

\subsection{Learning Hierarchical States and Actions using the Successor Representation}
As discussed in the previous section, the successor representation is useful for computing new value functions on the fly, but as the state space size increases, learning a full successor matrix becomes difficult. We address this problem by learning state and action hierarchies. Hierarchical approaches, such as using higher-level abstract actions (also known as options, \cite{sutton1999between}), have been widely studied in reinforcement learning. A hierarchy that involves abstract states and actions can significantly reduce the number of planning steps required, though potentially at the cost of not having a fully optimal plan with respect to the lowest-level actions. Animals and humans appear to use hierarchical strategies, such as using a sequence of pre-learned habitual behaviors to solve a complex task \parencite{friston2018deep,tomov2020discovery}.

%  \begin{figure}[ht]
%     \centering
%     \includegraphics[width=0.5\textwidth]{images/grid.png}
%     \caption{FrozenLake with $n=8, d=2, m=4$}
%     \label{fig:grid}
% \end{igure}

\begin{figure}[!htbp]
\begin{minipage}[t]{\textwidth} % or '[b]', if desired
\begin{subfigure}{0.45\textwidth}
        \includegraphics[width=\textwidth]{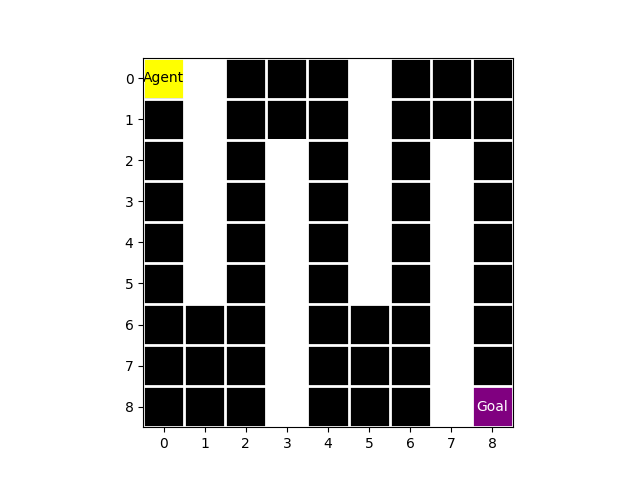} 
        \caption{}
        \label{fig:grid_mdp}
        \end{subfigure}
        \hfill
        \begin{subfigure}{0.45\textwidth}
        \includegraphics[width=\textwidth]{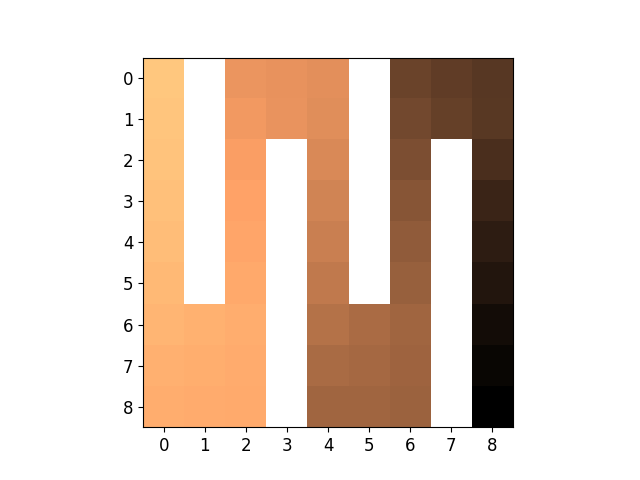} 
        \caption{}
        \label{fig:M_origin_mdp}
        \end{subfigure}
    \begin{subfigure}{0.45\textwidth}
        \includegraphics[width=\textwidth]{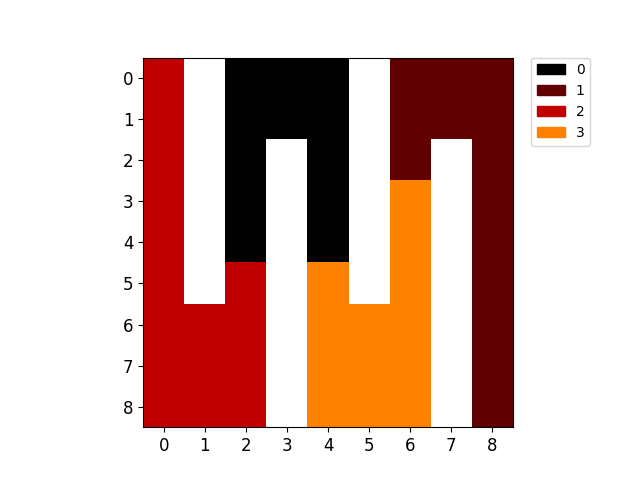}
        \caption{}
        \label{fig:macro_state_mdp}
    \end{subfigure}
    \hfill
    \begin{subfigure}{0.45\textwidth}
    \centering
    \includegraphics[width=\textwidth]{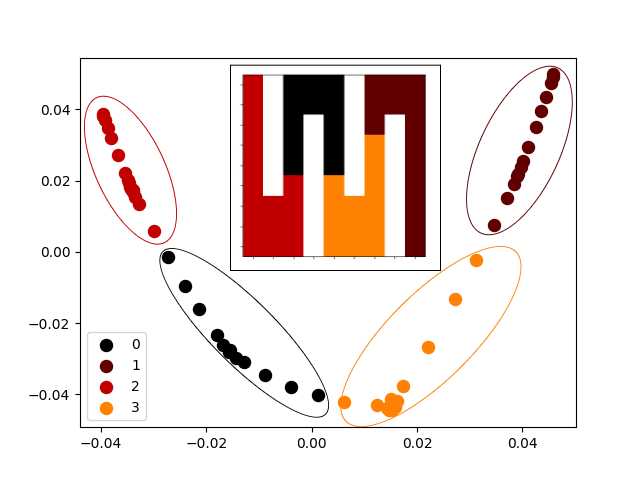}
    \caption{}
    \label{fig:cluster_mdp}
    \end{subfigure}
    \caption{{\bf Task 1: Serpentine MDP Gridworld Environment}\\ (a) A $9\times 9$ Gridworld with a single start state (yellow), a single goal state (purple), and walls (solid white rectangles). (b) Successor representation of states with respect to the start state. The states closer to the start state have larger successor values and are represented by a lighter shade of brown. (c) Macro states learned by the model ($4$ in number). (d) Visualization of the $4$ macro state clusters in terms of projections of the $81$ micro states into an embedding space.}
\label{fig:mdp}
\end{minipage}
\end{figure}

 We illustrate our approach to learning hierarchical states and actions using the classic problem of navigating in a ``Gridworld.'' Figure~\ref{fig:grid_mdp} shows a $9\times 9$ grid with a serpentine path layout, defining a Markov Decision Process (MDP). The MDP is defined as follows: \begin{equation}
     {\cal M}_1 = \{S,A,T,R\}
 \end{equation} where $S$ is the set of states corresponding to discrete locations (black squares in Figure~\ref{fig:grid_mdp}), $A$ is the set of 4 discrete actions (attempt to move left (L), right (R), up (U), and down (D) in the grid), $T$ is the Markov transition function:
 \begin{equation}
     T:S\times A\rightarrow S
 \end{equation} determining the next state given the current state and action, and $R:S\rightarrow \mathbb{R}$ is the reward function specifying a scalar reward for each state. We assume for simplicity that both $T$ and $R$ are deterministic. Note that $T$ is determined by the structure of a particular Gridworld, i.e., where the walls are, and executing an action that moves the agent against a wall results in the agent remaining in the same state/location (e.g., action U, L or R in the starting state in Figure~\ref{fig:grid_mdp}).
 
A common task for the agent in this Gridworld (``Task 1'') is to navigate to a specially designated state called the goal state. The goal state could potentially change from one scenario to the next. In our experiments, the reward function is defined as follows for any state $s \in S$:
\begin{equation}
    R(s) = \begin{cases}
    +100 & \text{if } s \text{ is the goal state} \\
    -0.1 & \text{otherwise} 
    \end{cases}
\end{equation}
We use the negative reward (or ``punishment'') of -0.1 to encourage short paths to the goal.

Consider the problem of learning a 2-level state-action hierarchy for Task 1. We assume that the agent uses the following uniformly random action selection policy to explore the environment.
\begin{equation}
    \text{For any } s \in S, \pi(s) = a \text{ with probability }\frac{1}{4} \text{ for the 4 possible } a \in A
\end{equation}
The agent uses this exploration of the Gridworld and the resulting transitions (from a state $s$ to state $s'$ in each time step) to learn a (potentially incomplete) successor matrix $M$ using the temporal difference rule (Equation~\ref{eq:TD-successor-learning-vector}):
\begin{equation}
M(s,:) = M(s,:) + \alpha(\mathbf{1}_{s'} + \gamma M(s',:) - M(s,:))
\end{equation}
where $\mathbf{1}_{s'}$ is a one-hot vector with value 1 at the position corresponding to state $s'$. Figure~\ref{fig:M_origin_mdp} shows the successor representation with respect to the start state $s_0$, i.e., the row $M(s_0,:)$, for the Gridworld in Figure~\ref{fig:grid_mdp}. 

In order to learn abstract states at the higher (``macro'') level, we cluster the lower-level (``micro'') states using the successor representation as a {\em similarity metric between states}. We tested the following clustering methods:
\begin{itemize}
    \item k-Medoids: This method is similar to k-Means, except the cluster centers are actual data points \parencite{Kaufman1990}.
    \item Agglomerative Clustering: This is a bottom-up ``hierarchical clustering'' algorithm where each data point starts as its own cluster and the method iteratively merges the two most similar clusters until a specified condition is met \parencite{Murtagh2012}. 
    \item Spectral Clustering: This method involves a spectral decomposition of the successor matrix $M$ and uses the eigenvectors of a Laplacian matrix of $M$ to reveal underlying clusters \parencite{ng2002spectral} (see \cite{Friston-et-al-2025a} for an alternate approach to creating macro states based on renormalising generative models). 
\end{itemize}
In our experiments, we used spectral clustering because it yielded superior performance over the other two methods, in terms of requiring less training episodes (or equivalently less complete successor matrices) to get accurate clusters. To visualize the clusters in our figures, we used spectral embedding on the successor matrix \parencite{scikit-learn}. 

Figure~\ref{fig:macro_state_mdp} shows the four macro states for Task 1 learned from spectral clustering of lower-level states (the number of macro states $k = 4$ was set a priori). Figure~\ref{fig:cluster_mdp} shows a 2D visualization of the clusters of lower-level states. 

Given the learned macro states (abstract states), the agent learns ``macro actions'' (abstract actions) for transitions from one macro state to another as follows. While exploring the environment, if the agent notices a transition from a micro state $s_1$ in macro state $S_1$ to a micro state $s_2$ that resides in a new macro state $S_2$, then $s_2$ is designated as a ``bottleneck'' state. The agent learns a macro action $A_{1\rightarrow 2}$ for the transition $S_1\rightarrow S_2$ using active inference in the lower level to plan paths from any lower level state in $S_1$ to the bottleneck state $s_2$. The macro action $A_{1\rightarrow 2}$ is learned as a lower-level state-action policy $\rho_{1\rightarrow 2}$ using these paths computed using active inference. We repeat this learning process for pairs of adjacent macro states. 

Given a set of macro states $S_1,\ldots,S_4$ and macro actions $A_{i\rightarrow j}$ for navigating from $S_i$ to $S_j$, the agent can learn a successor matrix at the higher level, either by leveraging past trajectories (used for learning macro states and actions) or by obtaining new trajectories directly from the environment, or both. Given these trajectories containing transitions from macro state $S$ to macro state $S'$, the higher level successor matrix is learned using the same temporal difference rule as Equation~\ref{eq:TD-successor-learning-vector}:
\begin{equation}
M(S,:) = M(S,:) + \alpha(\mathbf{1}_{S'} + \gamma M(S',:) - M(S,:))
\end{equation}
where $\mathbf{1}_{S'}$ is a one-hot vector with value 1 at the position corresponding to state $S'$.

For active inference at the higher level, the agent requires EFE values for macro states. For each macro state $S$, the agent aggregates the one-step EFE values from all the micro states that the macro state encompasses:
\begin{equation}
    G(S)=\sum_i G(s_i) \text{ for all }s_i\in S
\end{equation}
In the case of a simple MDP, these EFE values are equivalent to the rewards of the states. This follows because there is no expected information gain (since the states are fully observable) and EFE reduces to expected reward (see Equation~\ref{eq:efe-tradeoff}). Given the macro level successor matrix and the macro state EFE values, the agent can perform active inference with macro actions by first computing the EFE function $\nu(S)$ (Equation~\ref{eq:nu-equation}) for all macro states $S$. Then, given any current macro state $S_i$, the agent selects the macro action $A^*_{i\rightarrow j}$ that will result in the macro state $S_j$ with the highest $\nu$ value:
\begin{equation}
    A^*_{i\rightarrow j} = A_{i\rightarrow j} \text{ such that }\nu(S_j) \geq \nu(S_k) \text{ for all } k 
\end{equation}
Finally, the agent executes the lower-level policy $\rho_{i\rightarrow j}$ corresponding to the macro action $A^*_{i\rightarrow j}$. This process is repeated until the agent reaches the goal state. 

\section{Results}
\label{sec:results}
\subsection{Performance Comparison between Hierarchical and\\Non-Hierarchical Active Inference in Task 1}
\label{sec:task1}
We compared the hierarchical active inference model described in the previous section with a model that uses active inference and the successor representation only at the lower-level \parencite{millidge2023active} on Task 1. 
While the results presented here are for a particular Gridworld model for Task 1 (the ``Serpentine'' layout), we have obtained similar results for other wall configurations in Gridworld, and different locations of start and goal states. In particular, the classic ``Four Rooms'' environment is used for the goal revaluation experiment later in this section (see also Appendix~\ref{sec:four_rooms} for the full clustering and planning pipeline on Four Rooms).

\begin{figure}[!htbp]
\begin{minipage}[t]{\textwidth}
    \begin{subfigure}{0.329\textwidth}
        \centering
        \includegraphics[width=\textwidth]{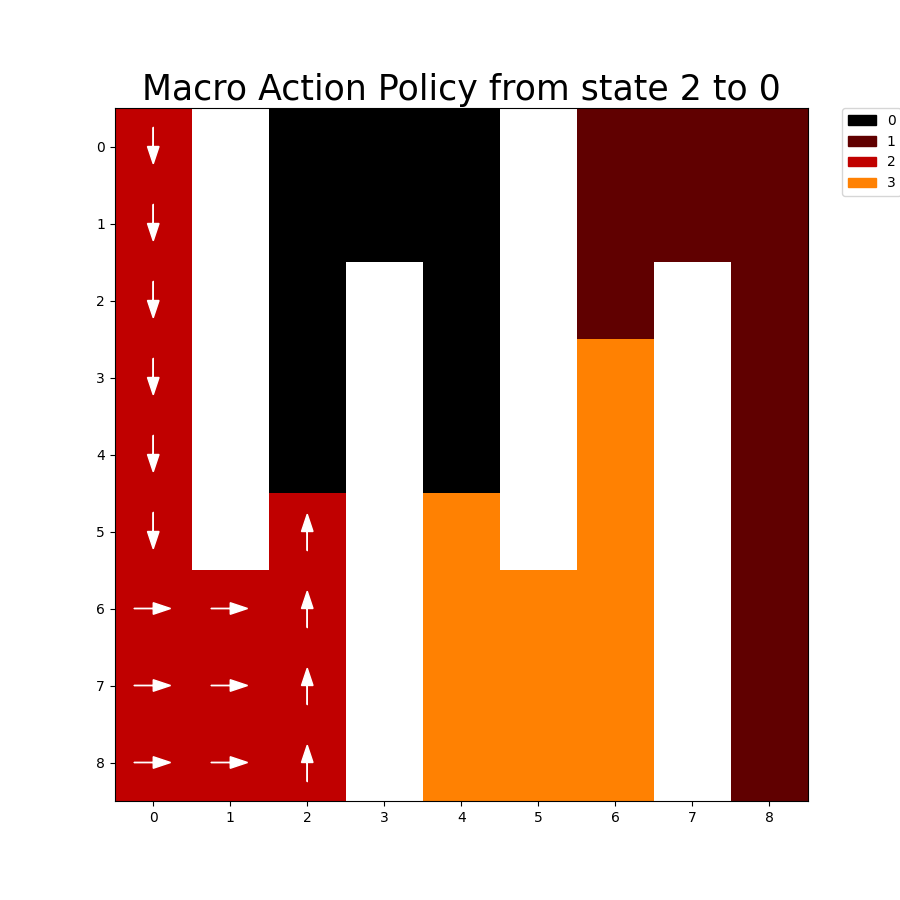} 
        \caption{}
        \label{fig:MA_2_0}
    \end{subfigure}
    \begin{subfigure}{0.329\textwidth}
        \centering
        \includegraphics[width=\textwidth]{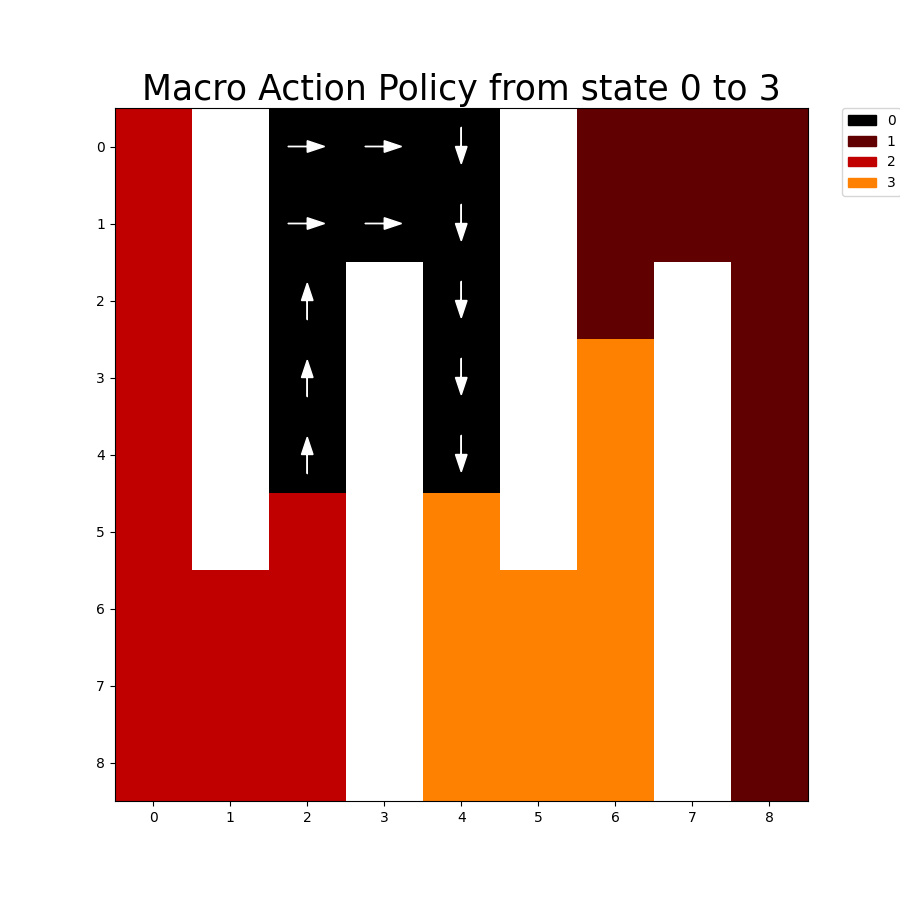}
        \caption{}
        \label{fig:MA_0_3}
    \end{subfigure}
    \begin{subfigure}{0.329\textwidth}
        \centering
        \includegraphics[width=\textwidth]{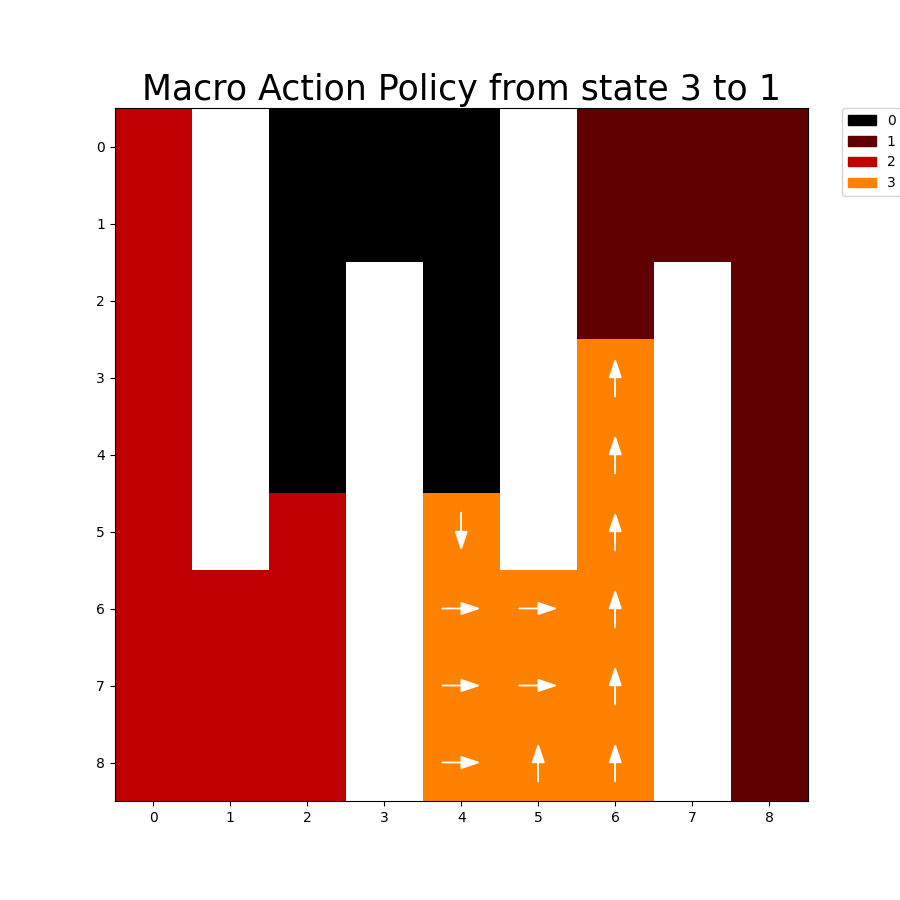}
        \caption{}
        \label{fig:MA_3_1}
    \end{subfigure}
    \caption{{\bf Macro Actions for Task 1.} (a)-(c) represent macro actions that execute a policy at the lower level for transitioning from one macro state to another (here, macro actions for $S_2\rightarrow S_0$, $S_0\rightarrow S_3$, and $S_3\rightarrow S_1$ are shown). The white arrows indicate the most probable action from each micro state in the first macro state for reaching the bottleneck state of the second macro state. }
\label{fig:MA}
\end{minipage}
\end{figure}

Figure~\ref{fig:MA} shows three examples of macro actions learned by the model for the learned macro states in Figure~\ref{fig:mdp}. These macro actions are deterministic state-action policies that help navigate the agent from one macro state to an adjoining macro state. As described in the previous section, the agent learns these macro action policies through exploration and active inference. Each learned macro action guides the agent to a bottleneck micro state in the second macro state, from any micro state in the first macro state (for example, macro state 2 to macro state 0 in Figure~\ref{fig:MA_2_0}). 

\begin{figure}[!htbp]
% \begin{minipage}[t]{\textwidth}
    \begin{subfigure}{\textwidth}
     \centering
        \includegraphics[width=0.55\textwidth]{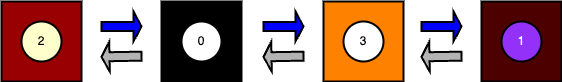}
        \caption{}
        \label{fig:macro_transitions_mdp}
    \end{subfigure}
    \hfill
    \begin{subfigure}{\textwidth}
        \centering
        \includegraphics[width=0.40\textwidth]{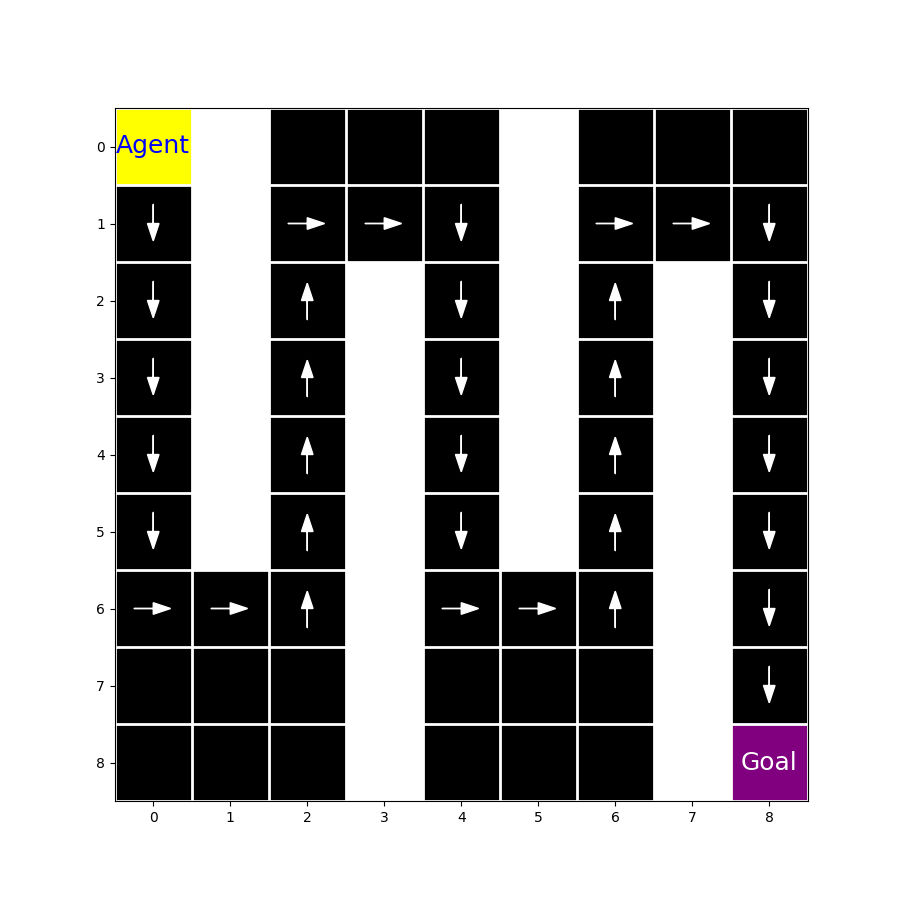} 
        \caption{}
        \label{fig:micro_plan_mdp}
    \end{subfigure}
    \caption{{\bf Planning using Hierarchical Active Inference for Task 1.} \\(a) Higher-level transition function and higher-level plan. The arrows represent the possible transitions between the learned macro states using the learned macro actions. The blue arrows represent the optimal sequence of macro actions (obtained using active inference at the macro level) to navigate from macro state $2$ (which contains the agent's starting micro state) to macro state $1$ (which contains the goal micro state).\\
    (b) Lower-level action sequence executed by the agent as a result of higher-level plan. The arrows represent the micro actions executed by the agent based on the optimal macro action plan depicted by the blue arrows in (a), along with the micro actions used within macro state 1 to reach the goal. Compare the relatively large number of micro actions needed to reach the goal micro state to the much smaller number of macro actions (here, $3$) that needed to be inferred with hierarchical active inference, illustrating the efficacy of planning using hierarchical states and actions.}
\label{fig:plan_mdp}
% \end{minipage}
\end{figure}

Figure~\ref{fig:plan_mdp} illustrates the mechanics of hierarchical planning using the proposed model. The agent first learns the transition dynamics of higher level macro states and macro actions (Figure~\ref{fig:macro_transitions_mdp}) as part of the process of learning macro actions. This higher-level transition matrix can be used for higher-level planning using active inference: the blue arrows in Figure~\ref{fig:macro_transitions_mdp} show the optimal sequence of macro actions found by the model for navigating from macro state~2 (containing the start state) to the macro state containing the goal (macro state~1). The lower-level policy for each macro action in this sequence is then executed to navigate from one macro state to the next, ultimately reaching the last macro state in the sequence (macro state~1). Within macro state~1, the agent uses lower-level planning based on active inference to reach the goal micro state. Figure~\ref{fig:micro_plan_mdp} depicts the sequence of micro actions resulting from this process, showing that the agent successfully navigates from the start micro state to the goal micro state.

\begin{figure}[!htbp]
\centering
\begin{minipage}[t]{\textwidth}
\begin{subfigure}{0.475\textwidth}

    \centering
    \includegraphics[width=\textwidth]{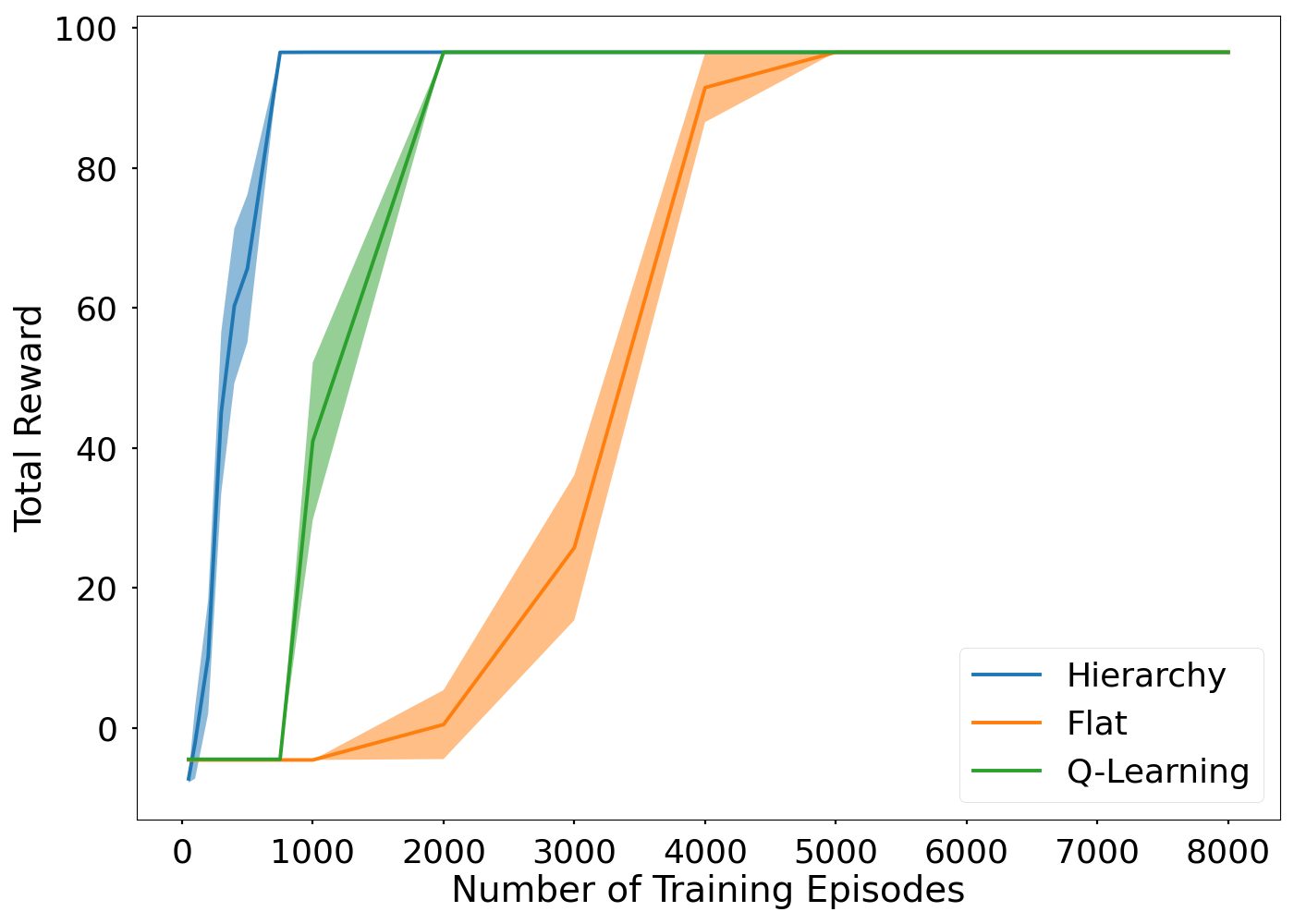}
    \caption{}
    \label{fig:rewards_full}
\end{subfigure}
\hfill
\begin{subfigure}{0.475\textwidth}
    \centering
    \includegraphics[width=\textwidth]{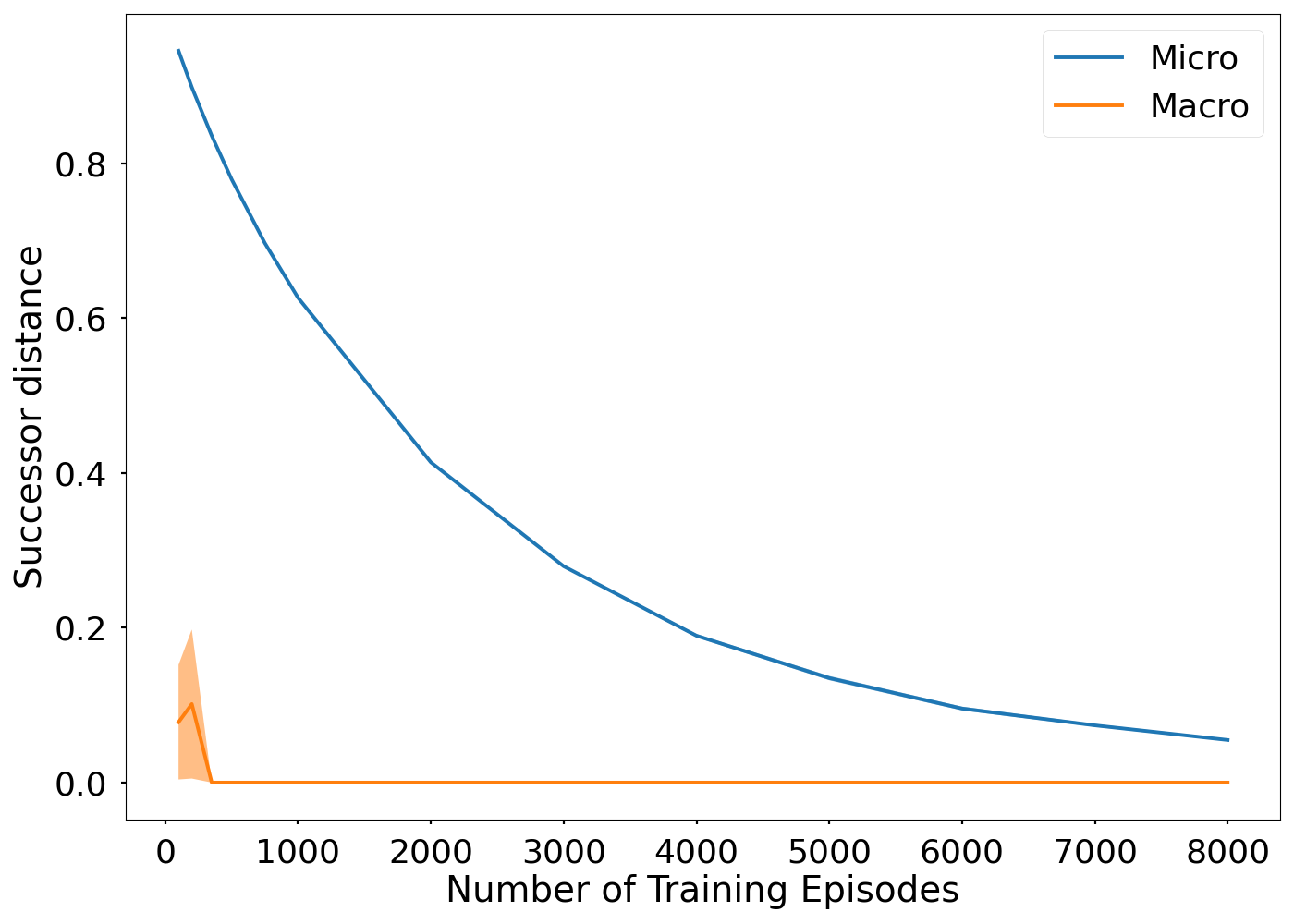}
    \caption{}
    \label{fig:succ_dist_full}
\end{subfigure}

\vspace{0.5em}

\begin{subfigure}{0.475\textwidth}
    \centering
    \includegraphics[width=\textwidth]{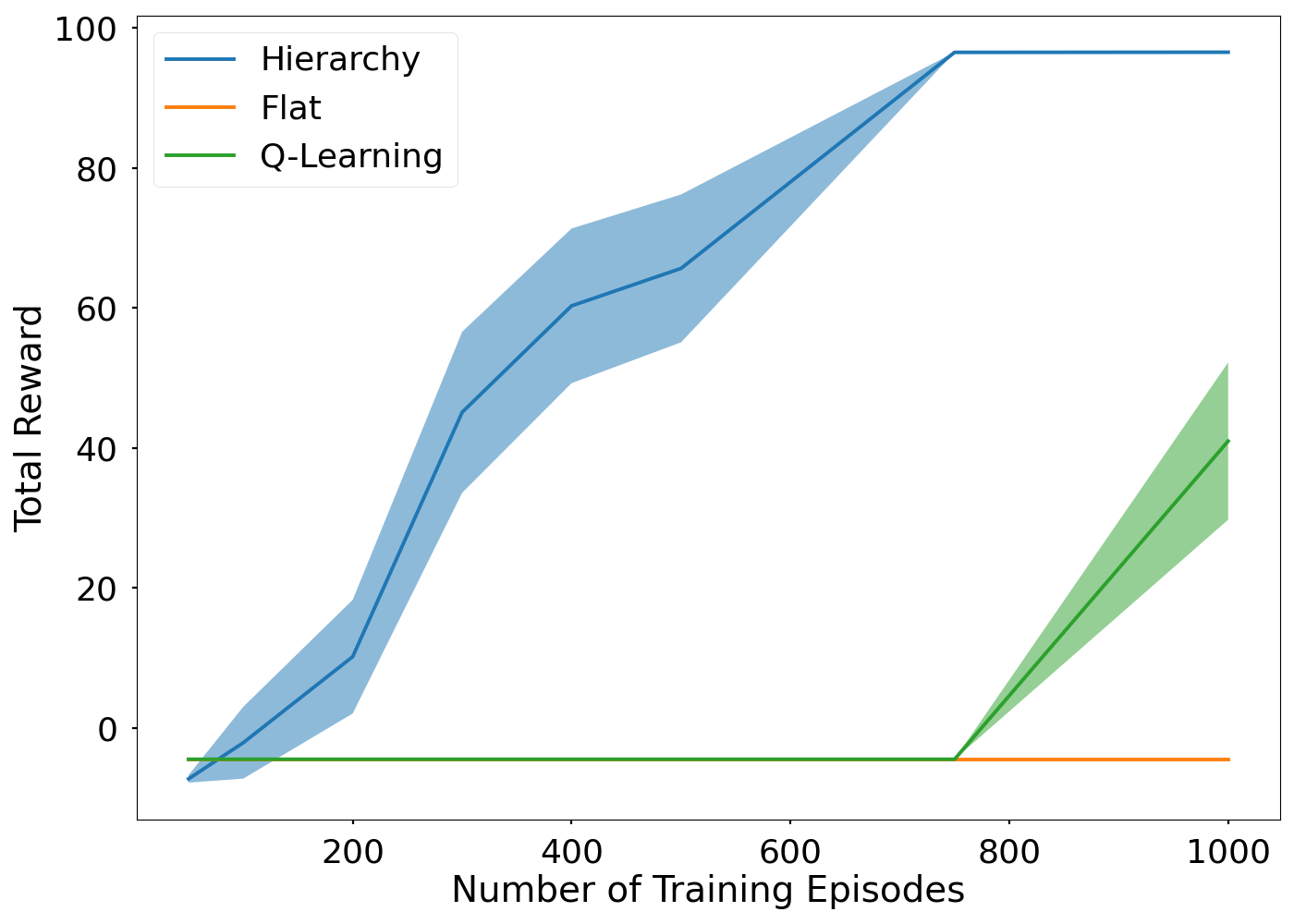}
    \caption{}
    \label{fig:rewards_initial}
\end{subfigure}
\hfill
\begin{subfigure}{0.475\textwidth}
    \centering
    \includegraphics[width=\textwidth]{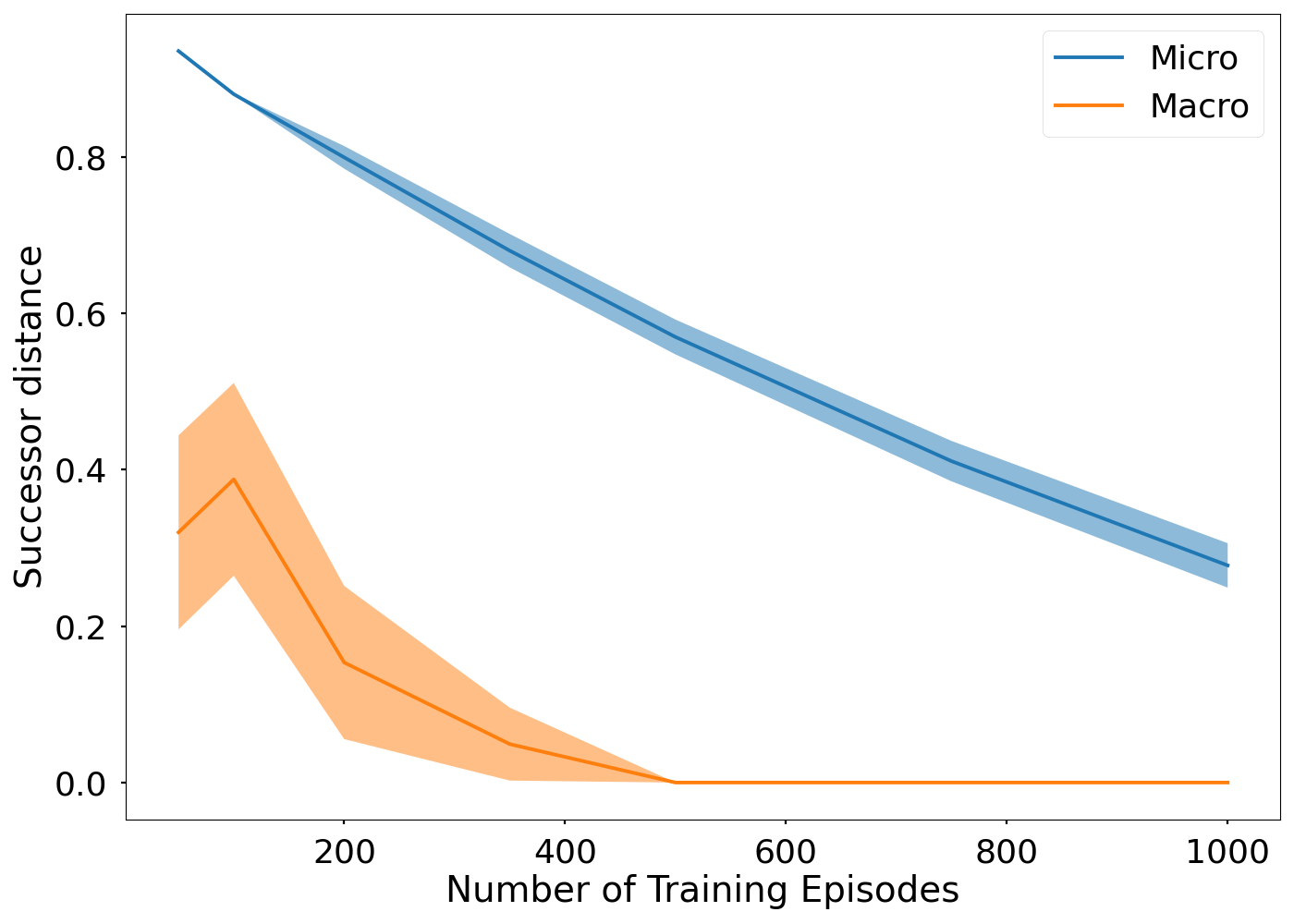}
    \caption{}
    \label{fig:succ_dist_initial}
\end{subfigure}

\vspace{0.5em}

\begin{subfigure}{0.475\textwidth}
    \centering
    \includegraphics[width=\textwidth]{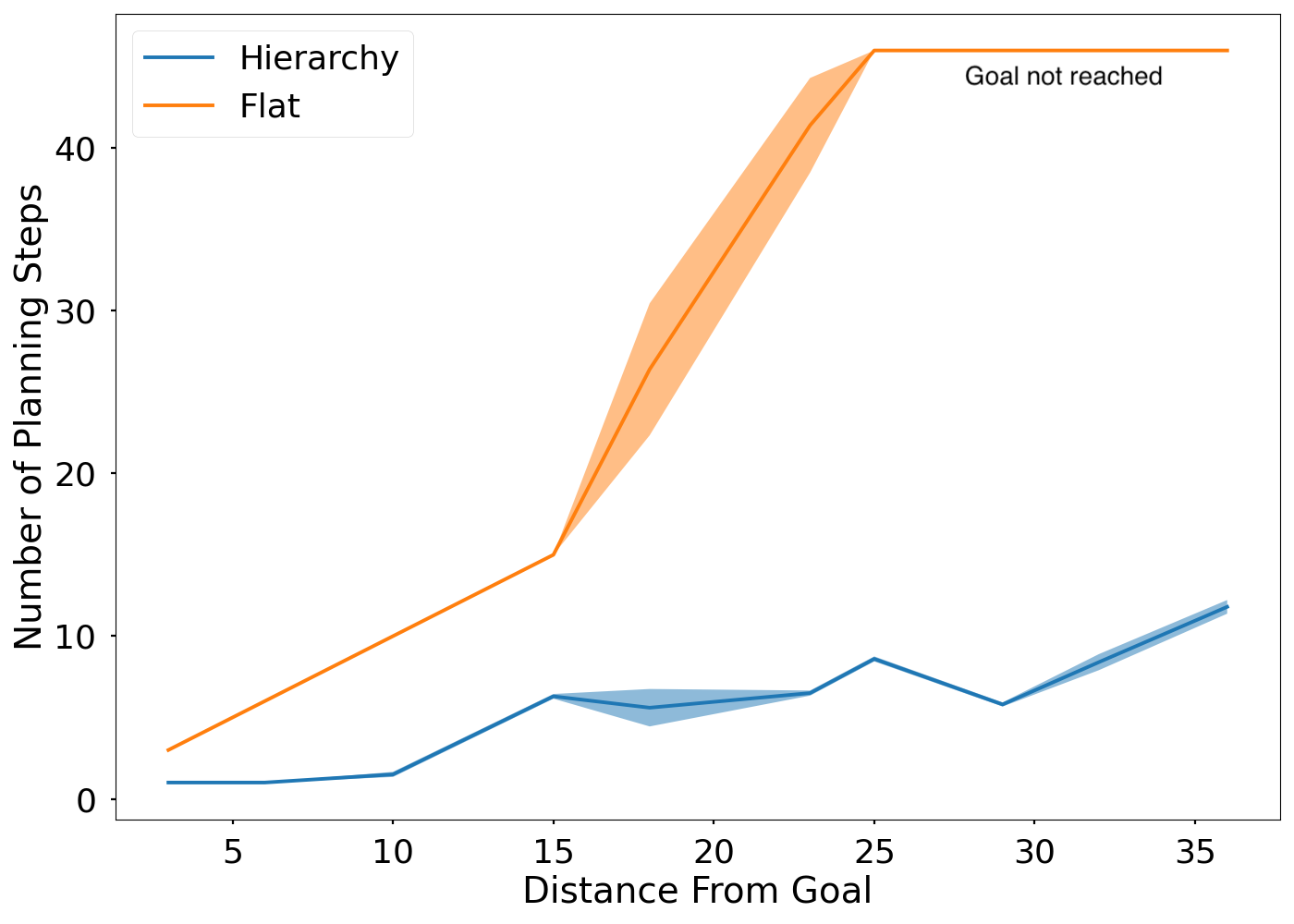}
    \caption{}
    \label{fig:distances_goal}
\end{subfigure}
\hfill
\begin{subfigure}{0.475\textwidth}
    \centering
    \includegraphics[width=\textwidth]{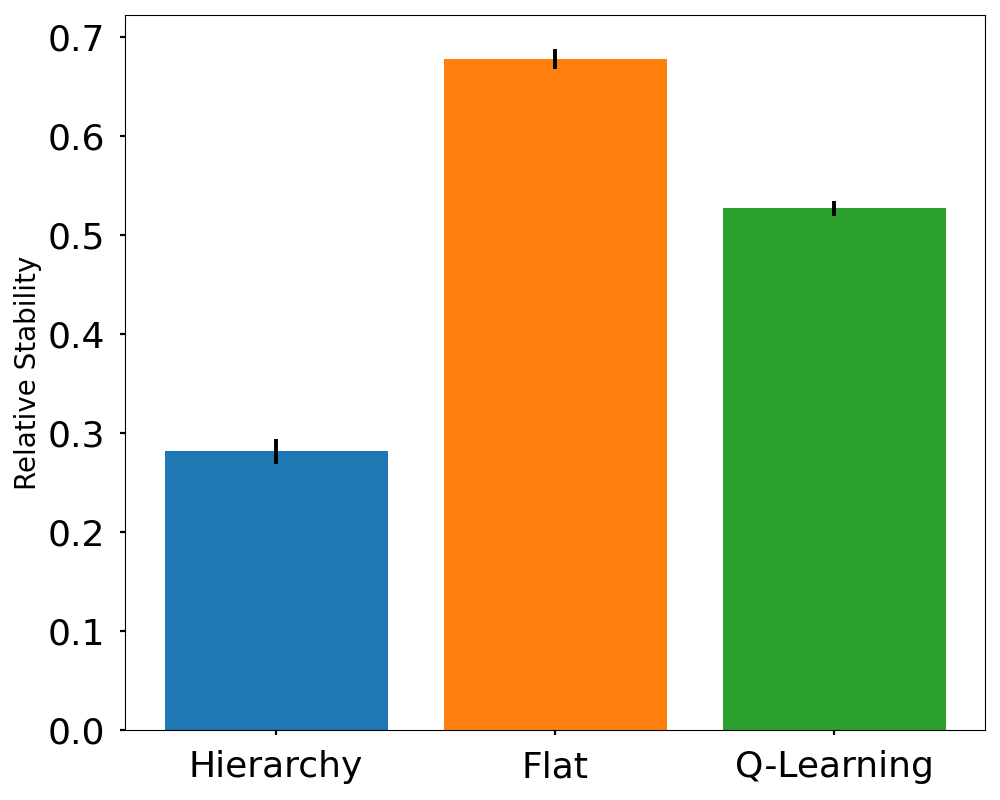}
    \caption{}
    \label{fig:rel_stability}
\end{subfigure}

\caption{{\bf Comparison of Planning Performance using Hierarchical, Flat Active Inference and RL baseline Models.}\\(continued next page)}
\end{minipage}
\end{figure}
\addtocounter{figure}{-1}
\begin{figure}[!htbp]
    \caption{
    (a) Total reward obtained by the agent as a function of number of training episodes. The flat model takes $10\times$ the number of training episodes compared to the hierarchical model in order to reach the goal. The Q-Learning model performs better than the flat active inference but hierarchical model works much better.
    (b) Macro versus micro level successor matrix distances. The distance between the true successor matrix and the learned successor matrix at different stages of the learning process for a flat versus hierarchical agent. 
    (c)-(d) Zoomed-in versions of (a)-(b). The plots show (a) and (b) for the first $1000$ training episodes to highlight the initial behavior of each agent more clearly.
    (e) Number of planning steps needed to find the goal as a function of distance of the start micro state from the goal. For each distance, planning performance was tested given a fixed number of training episodes ($1200$). Note that the flat model fails to reach the goal for distances 25 steps or more from the goal (the agent was required to find the goal within a certain upper bound of planning steps).
    (f) The relative stability (RS) of both the hierarchical and flat methods (the lower the better). The hierarchical method is more stable ($RS = 0.281\pm 0.013$) and maintains performance closer to its best achieved return than flat control ($RS = 0.677\pm 0.010$), while the RL method is in between ($RS = 0.526\pm 0.008$).}
\label{fig:graphs_mdp}
\end{figure}

%We find that with the aid of the action and state hierarchy, the model is able to plan to the goal with much less training as compared to the flat model. At inference time there are much fewer planning steps that need to be inferred, thus allowing the agent to do inference more efficiently and accurately. 
We compared the performance of our hierarchical model to a flat active inference model in learning a successor matrix and using the current learned matrix to navigate to the goal micro state. Also, as a baseline, we used a simple Q-Learning model \parencite{sutton-barto-book}. Both types of models were tested for multiple random initializations. The average total reward obtained by each type of model is plotted in Figure~\ref{fig:rewards_full} (the shaded regions correspond to +/- 1 standard deviation). Since the rewards are sparse, with only the goal giving a positive reward, the average total reward also behaves as a proxy for the average success rate of the model, so if the average reward is close to $100$, this means that the agent always reached the goal, while if it is negative, the agent never reached the goal.
As a sanity check we also evaluated a random policy that selects actions
uniformly at random, without learning a value function. 

As shown in
Table~\ref{tab:random_baseline}, the agent with a uniformly random action selection policy almost never reaches the goal
in the serpentine layout. Even with a generous budget of 1000 steps, the
success rate is only 12.1\%. Moreover, the average reward decreases with
the step budget: although more steps provide additional opportunities to
stumble upon the goal, the majority of episodes exhaust the full budget of steps in trying to reach the goal, accumulating a step cost of $-0.1$ per time step that far outweighs the occasional $+100$ reward when an episode ends in success.

\begin{table}[h]
\centering
\caption{Random Policy: Success Rate and Average Reward vs.\ Maximum Steps Allowed}
\label{tab:random_baseline}
\begin{tabular}{ccc}
\toprule
\textbf{Max Steps Allowed} & \textbf{Success Rate (\%)} & \textbf{Avg Reward} \\
\midrule
45 & 0.0 & $-4.50$ \\
100 & 0.0 & $-10.00$ \\
250 & 0.0 & $-25.00$ \\
500 & 1.5 & $-48.38$ \\
1000 & 12.1 & $-84.34$ \\
\bottomrule
\end{tabular}
\end{table}

As seen in Figure~\ref{fig:rewards_full}, as the number of training episodes increases, the hierarchical model quickly leverages its current learned macro states and macro actions to achieve increasingly larger total reward, eventually reaching the goal state consistently after only about 750 training episodes (see Figure~\ref{fig:rewards_initial}). In contrast, the flat model fails to reach the goal even after several thousand training episodes due to the large number of planning steps needed at the micro action level to find a path to the goal location. The Q-learning model fared somewhat better compared to the flat model, but hierarchical active inference model was much better at successfully navigating to the goal.

How is the hierarchical model able to learn to navigate to the goal state so quickly compared to the flat model? The answer is suggested by Figures~\ref{fig:succ_dist_full} and \ref{fig:succ_dist_initial}, which plot the distance between the true successor matrix and the current learned successor matrix as a function of the number of training episodes. By comparing these plots to Figures~\ref{fig:rewards_full} and \ref{fig:rewards_initial}, we can see that the superior planning performance of the hierarchical model can be attributed to the much faster learning of the successor matrix at the macro level. Figure~\ref{fig:succ_dist_initial} illustrates how the macro level successor matrix can converge to the true matrix, even if the micro level successor matrix is not yet very accurate.

An additional test of the two models for Task 1 involved moving the goal to progressively larger distances from the start micro state. As shown in Figure~\ref{fig:distances_goal}, the hierarchical model scales much better, with modest increases in the number of planning steps as distance to the goal increases, whereas the flat model scales poorly and fails to reach the goal when distances are 25 steps or more.

Next, we calculated the learning stability of both the hierarchical and flat methods using a metric called R-stability, introduced in \parencite{Ororbia2023}. This metric measures the average normalized deviation of performance from the best achieved reward/return over the final phase of training. Lower values indicate that the agent after learning the policy maintains near-optimal behavior, i.e., it avoids any form of catastrophic forgetting. Higher values would indicate that model may be unstable or even collapse after transiently learning a good policy. As observed in Figure~\ref{fig:rel_stability}, the hierarchical model achieves a substantially lower value than the flat model and Q-learning baseline, indicating that once effective behavior is learned by the hierarchical model, it is consistently maintained. This suggests that hierarchical structure not only accelerates learning and planning, but also stabilizes behavior by reducing sensitivity to exploration noise and local deviations in policy execution. 

\begin{figure}[!htbp]
\begin{minipage}[t]{\textwidth}

    \begin{subfigure}{\textwidth}
        \centering
        \includegraphics[width=\textwidth]{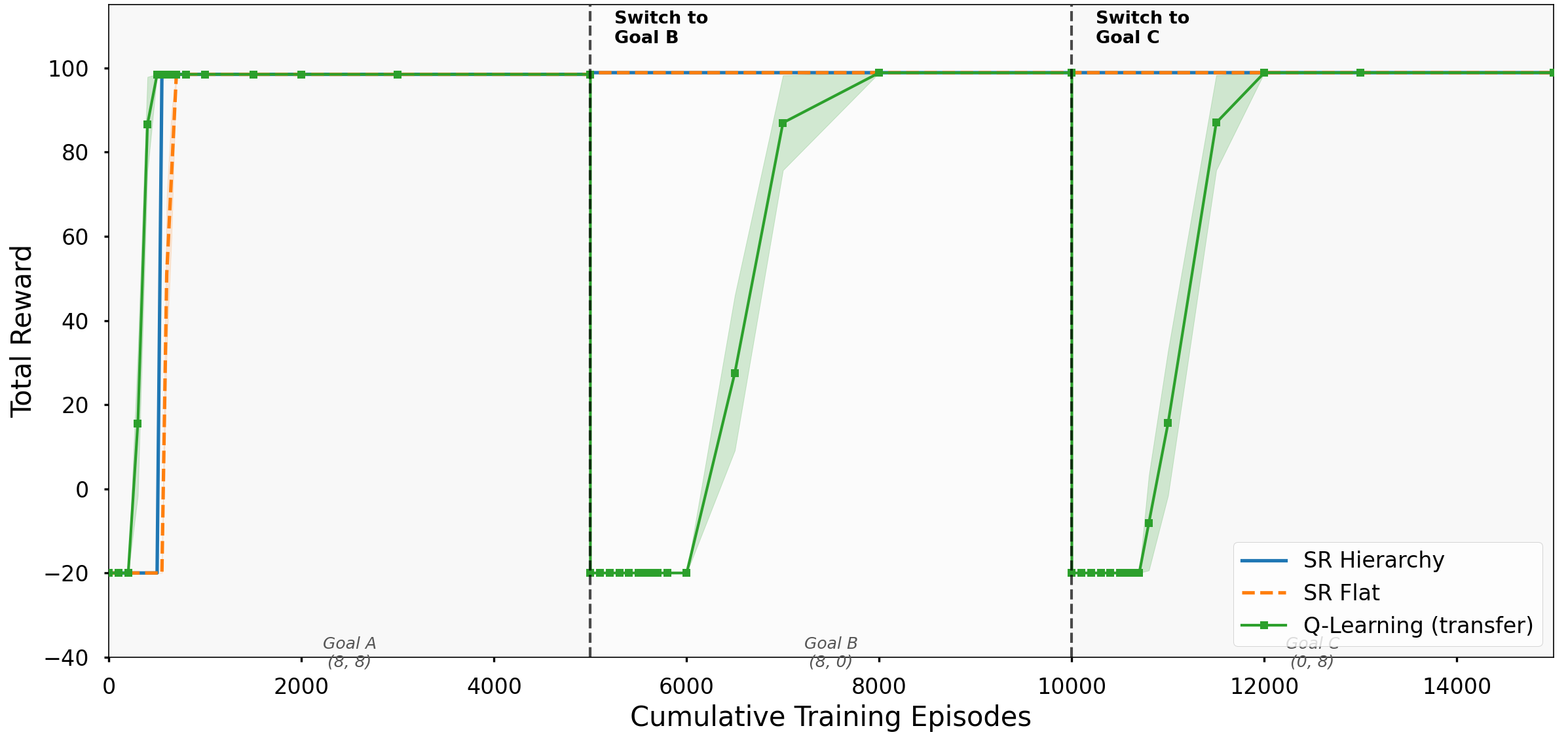}
        \caption{}
        \label{fig:goal_reval}
    \end{subfigure}
    \hfill
    \begin{subfigure}{\textwidth}
        \centering
        \includegraphics[width=0.5\textwidth]{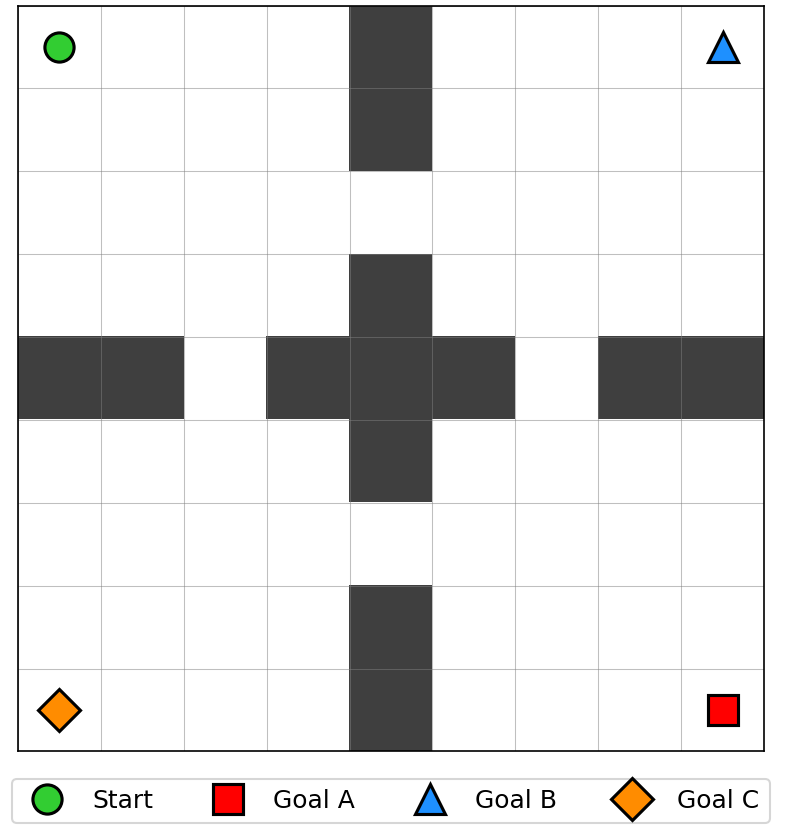}
        \caption{}
        \label{fig:layout}
    \end{subfigure}

    \caption{{\bf Goal Change Results.} \\
    (a) Performance when the goal is changed for the Q-learning agent and the hierarchical and flat agents based on successor representation (SR). The SR agents quickly re-plan, whereas Q-learning requires substantial retraining. 
    (b) Four rooms environment illustrating the sequence of goals used during the experiment.}

    \label{fig:goal_reval_combined}

\end{minipage}
\end{figure}

Our final experiment compared the performance of the three models in a re-planning task where the goal is periodically changed, as seen in Figure~\ref{fig:goal_reval}. 
The Q-learning method in this task does not start from scratch when the goal is changed; it keeps its learned Q values and continues its training process. The multiple goal locations are placed at different rooms in the four rooms environment, the layout of which is shown in Figure~\ref{fig:layout}. Clearly, the hierarchical and flat methods, both based on the successor representation (SR), are instantly able to re-plan and navigate to a new goal since they have learned a successor matrix that is independent of the reward structure of the environment and can therefore quickly compute a new value function. On the other hand, Q-learning has to effectively unlearn its existing Q-value function while trying to learn a new function to navigate to the new goal location. 

%Further, using bottleneck states and a hierarchical model allows the agent to reach the goal even when the underlying successor matrix isn't as accurate. Figure~ \ref{fig:succ_dist_initial} when viewed in conjunction with Figure~\ref{fig:rewards_initial} clearly demonstrates how the hierarchical agent is able to receive a high total reward soon after the macro level successor converges to the true value, even if the micro level successor matrix is still not very accurate. And it takes a large number of epochs for the micro level successor to converge and lower-level model to reach the goal, as can be viewed in figures~\ref{fig:rewards_full} and~\ref{fig:succ_dist_full}.

\subsection{Task 2: Gridworld with Key}
\label{sec:task2}
\begin{figure}[!htbp]
\begin{minipage}[t]{\textwidth}
    \begin{subfigure}{0.45\textwidth}
        \includegraphics[width=\textwidth]{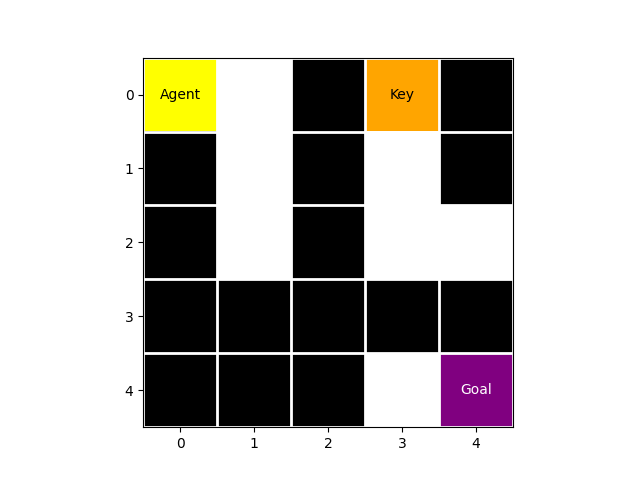} 
        \caption{}
        \label{fig:grid_key}
    \end{subfigure}
    \hfill
     \begin{subfigure}{0.44\textwidth}
        \includegraphics[width=\textwidth]{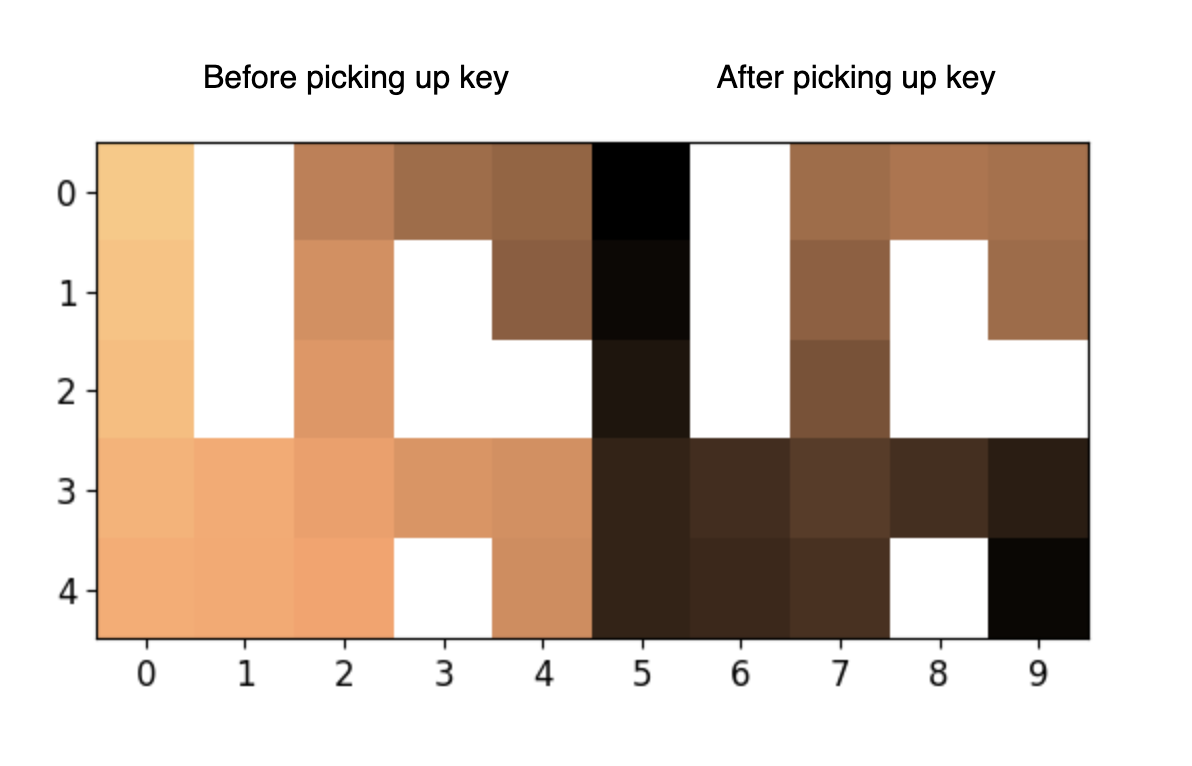} 
        \caption{}
        \label{fig:origin_key}
    \end{subfigure}
    \begin{subfigure}{0.45\textwidth}
        \includegraphics[width=\textwidth]{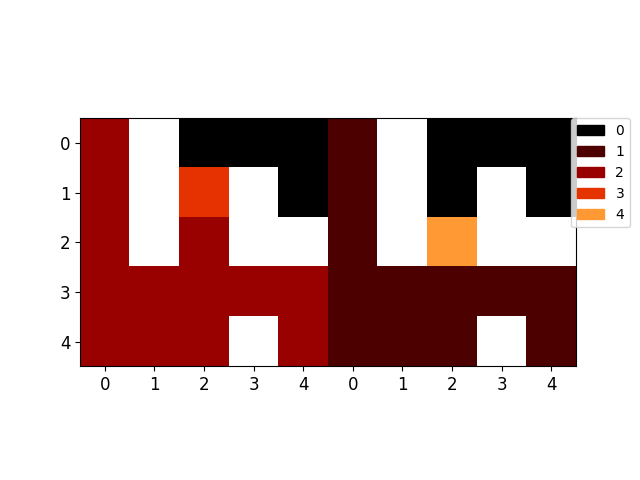}
        \caption{}
        \label{fig:macro_state_key}
    \end{subfigure}
    \hfill
    \begin{subfigure}{0.5\textwidth}
        \centering
        \includegraphics[width=\textwidth]{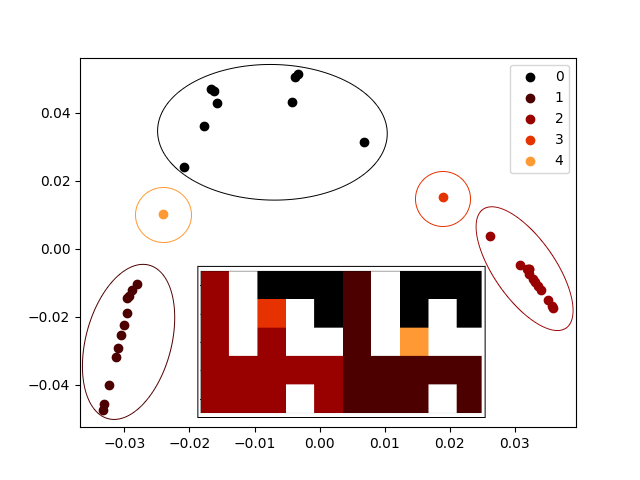}
        \caption{}
        \label{fig:cluster_key}
    \end{subfigure}
    \caption{{\bf Task 2: Gridworld with Key.}\\
    (a) A $5\times 5$ Gridworld with the agent's start location (yellow), goal (purple), key location (orange), and walls (white).
    (b) Successor representation with respect to start micro state (brighter is higher successor value). The left half corresponds to the micro states of the agent prior to picking up the key, while the right half corresponds to after.
    (c) Macro states learned by the model (number of macro states was set to $5$).
    (d) Visualization of the $5$ macro state clusters showing the micro states they contain.}
    \label{fig:key}
\end{minipage}
\end{figure}

We next tested the hierarchical model on a Gridworld task with the added complexity of requiring the agent to have a key to be able to ``unlock'' the goal when it is reached and get the reward. Figure~\ref{fig:grid_key} illustrates this new task (``Task 2''): the agent must navigate to the location containing the key (which may be at a distance from the goal) and execute a new ``pickup" action to get the key, before making its way to the goal. The action space for this task therefore consists of $5$ actions: the four actions to move in each of the four directions and the pickup action.

Thus, if the number of locations in the Gridworld is $n$, then the number of micro states in the Gridworld with Key task is $2n$, assuming each micro state represents a location as well as whether or not the agent has the key. While a factored representation would be more efficient, for simplicity, we use here a one-hot vector with length $2n$ to represent each state at the micro level.

\begin{figure}[!htbp]
\begin{minipage}[t]{\textwidth} 
    \begin{subfigure}{0.45\textwidth}
        \includegraphics[width=\textwidth]{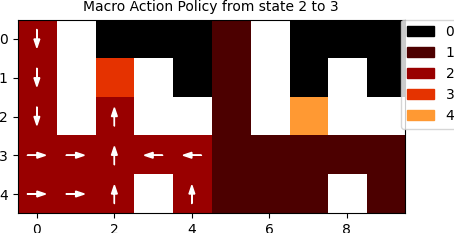}
        \caption{}
        \label{fig:MA_2_3_key}
    \end{subfigure}
    \hfill
    \begin{subfigure}{0.45\textwidth}
        \centering
        \includegraphics[width=\textwidth]{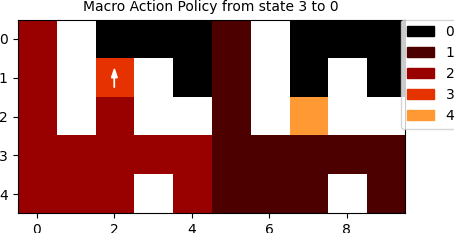}
        \caption{}
        \label{fig:MA_3_0_key}
    \end{subfigure}
    \begin{subfigure}{0.45\textwidth}
        \includegraphics[width=\textwidth]{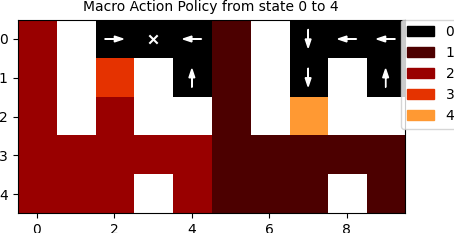}
        \caption{}
        \label{fig:MA_0_4_key}
    \end{subfigure}
    \hfill
    \begin{subfigure}{0.45\textwidth}
        \centering
        \includegraphics[width=\textwidth]{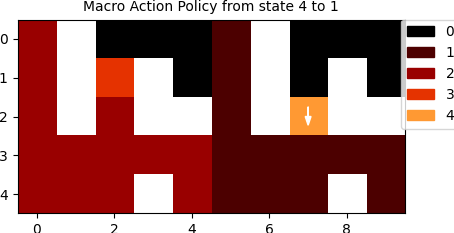}
        \caption{}
        \label{fig:MA_4_1_key}
    \end{subfigure}
    \caption{ {\bf Macro Actions for Task 2.}\\ (a)-(d) show four of the macro actions, in terms of their corresponding lower level policy, for transitioning from one macro state to another. The macro actions shown are for the transitions $S_2\rightarrow S_3$, $S_3\rightarrow S_0$, $S_0\rightarrow S_4$ and $S_4\rightarrow S_1$ ($S_i$ represents macro state $i$ in the plots). For each $S_i\rightarrow S_j$ macro action, the white arrows indicate the most probable action for each micro state within macro state $S_i$ that enable the agent to reach the bottleneck state of $S_j$. `x' represents the pick-up action. }
\label{fig:macro_action_key}
\end{minipage}
\end{figure}

Figures~\ref{fig:origin_key}-\ref{fig:cluster_key} illustrate how the agent learns the macro states for Task 2. The learned successor representation (for the start state) is shown in Figure~\ref{fig:origin_key}: for visual depiction, the states corresponding to locations where the agent has not yet picked up the key are shown on the left, and the states corresponding to the same locations but after the key has been picked up are shown on the right. The results of clustering the successor representation to obtain macro states are shown in Figures~\ref{fig:macro_state_key}-\ref{fig:cluster_key} (other macro states are also possible depending on choice of clustering method and parameters). Figure~\ref{fig:macro_action_key} shows four of the macro actions learned by the model to transition between the macro states in Task 2, based on the macro states learned in Figure~\ref{fig:key}. Note that given the increase in the size of the state space compared to Task 1, the agent requires more training episodes than in Task 1 for learning the macro states, macro actions, and successor matrix for active inference.

\begin{figure}[!htbp]
\begin{minipage}[t]{\textwidth}
\begin{subfigure}{\textwidth}
\includegraphics[width=\textwidth]{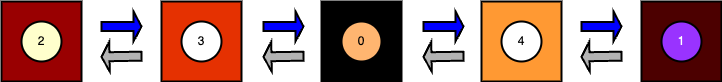}
\caption{}
\label{fig:macro_plan_key}
\end{subfigure}
\begin{subfigure}{0.45\textwidth}
        \includegraphics[width=\textwidth]{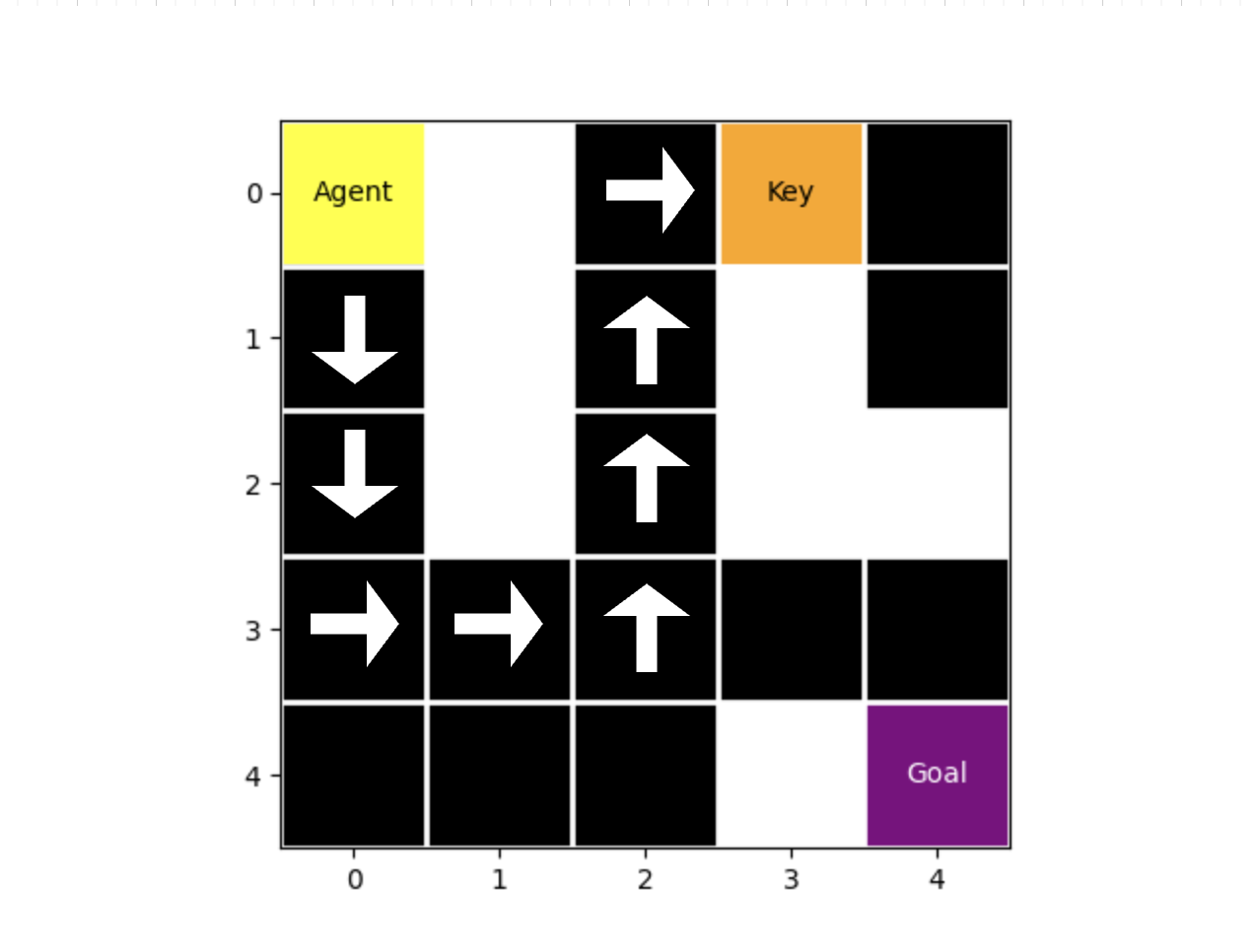} 
        \caption{}
        \label{fig:path_key_1}
    \end{subfigure}
    \hfill
    \begin{subfigure}{0.45\textwidth}
        \includegraphics[width=\textwidth]{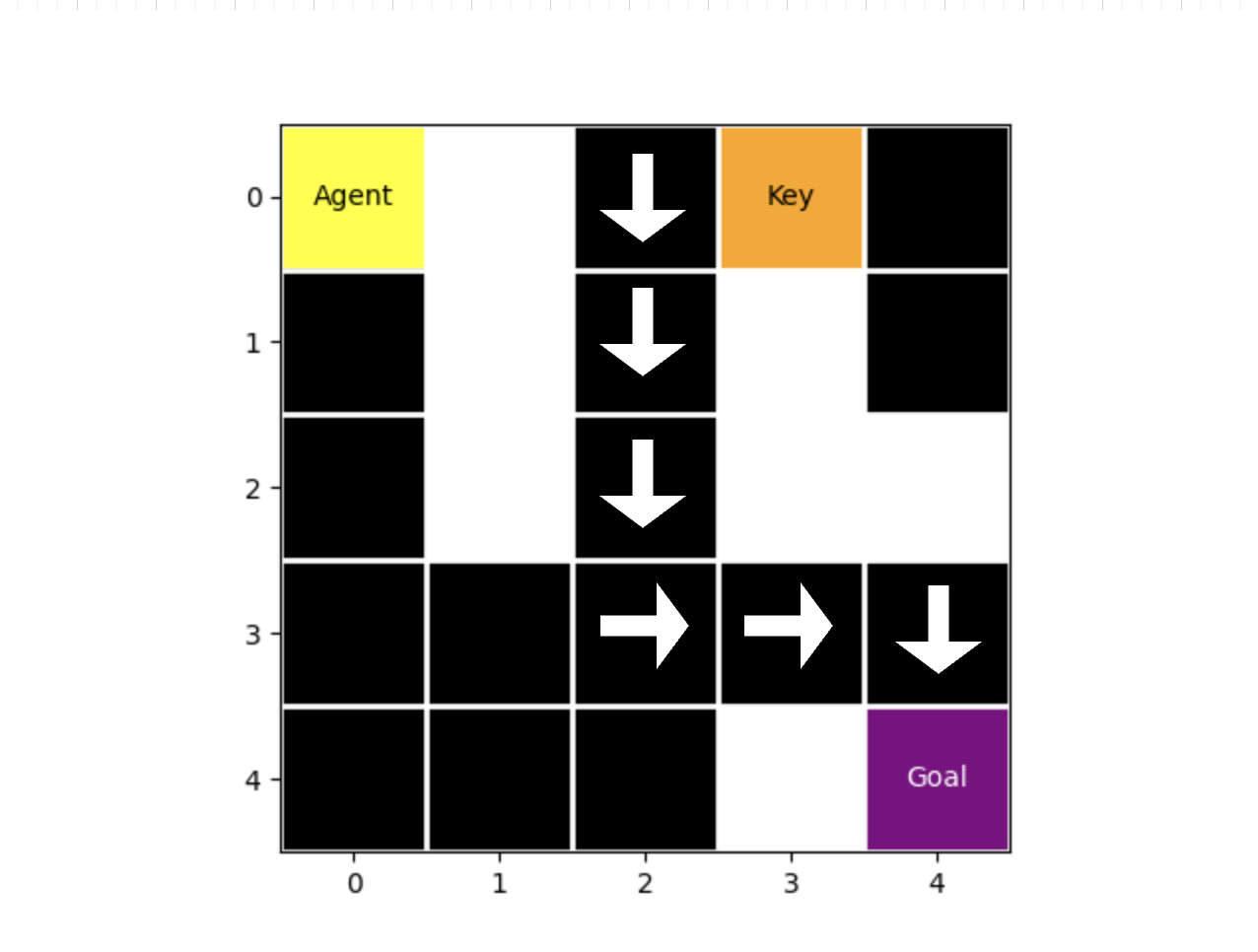} 
        \caption{}
        \label{fig:path_key_2}
    \end{subfigure}
    \caption{{\bf Planning using Hierarchical Active Inference for Task 2.}\\
    (a) Higher Level Planning: The arrows represent macro actions between pairs of adjacent macro states. The blue arrows represent the macro action plan inferred by the agent through active inference at the macro level. This sequence of macro actions takes the agent from macro state $2$ (containing the start location, no key) to $1$ (containing the goal location, with key) through the macro state $0$ which contains the location with the key.
    (b)-(c) show the execution of the plan in (a) in terms of the micro actions (white arrows) executed by the agent, allowing it to navigate from the start location to the goal, with a detour to pick up the key before going to the goal.}
\label{fig:path_key}
\end{minipage}
\end{figure}

The macro level plan, obtained via active inference using the learned macro states and actions, is depicted by the blue arrows in Figure~\ref{fig:macro_plan_key} (gray arrows represent other macro actions not used in this plan). The plan includes the macro action for transitioning from macro state $S_0$ to macro state $S_4$, which includes picking up the key (see Figure~\ref{fig:MA_0_4_key}). The execution of the macro level plan and the corresponding micro actions are depicted in Figures~\ref{fig:path_key_1} and \ref{fig:path_key_2}, which show that the agent is able to successfully navigate to the key, pick it up, and reach the goal to unlock the reward.

\subsection{Task 3: POMDP Gridworld}
\label{sec:task3}
Our results thus far illustrate the efficacy of the successor representation in the MDP case. We now investigate hierarchical planning using active inference in the case where we have a partially observable environment. We modify Task 1 by introducing noise in the Gridworld environment.

In a standard MDP, the agent observes the hidden state exactly, meaning its observation space is identical to its state space. Now consider introducing some localized noise into the observation of the true state of the environment. Under this condition, for a given true state or grid location, the agent receives an observation that suggests it is located somewhere in the local neighborhood of the agent. To be more specific, in the likelihood matrix $A$ that encodes $P(o|s)$, say $o_k$ is the observation corresponding to the true state $s_k$ and $o_{k'}$ corresponds to an observation in the local neighborhood of $s_k$. We define the local neighborhood as any location one step to the left, right, up, or down from $s_k$ (excluding walls). The noise is evenly distributed in this neighborhood. Then, given a noise parameter $\eta$, the likelihood is defined as $P(o_k|s_k) = 1-\eta$ and $\sum_{k'}P(o_{k'}|s_k) = \eta$ where all the $P(o_{k'}|s_k)$ have equal values. The agent is assumed to have prior knowledge of this noise (e.g., through learning).

While active inference generally considers a partially observed environment (i.e., a POMDP), learning the successor matrix is not as straightforward in POMDPs as in the case of MDPs since the agent doesn't have access to the actual states. Instead the agent estimates a belief vector via state inference and uses that to update the successor matrix with the TD rule. Similar to the MDP case, the agent aggregates the EFE values of the micro states in order to compute the EFE for the macro states. In this case, the EFE values also depend on the entropy of the observational noise and not just the rewards of the state.

If we run our hierarchical model on a modification of Task 1 with added localized observational noise, the agent is able to plan to navigate to the goal, with clusters similar to the MDP case (for results, see Appendix~\ref{sec:pomdp}).

We focus on a new dimension introduced to the task when we convert the Gridworld MDP to a POMDP. A major advantage of using active inference for planning is that it naturally balances exploration versus exploitation. Here we investigate how our hierarchical agent balances these two aspects when navigating to a goal in our noisy POMDP environment.
\begin{figure}[!htbp]
\begin{minipage}[t]{\textwidth} % or '[b]', if desired
    \begin{subfigure}{0.45\textwidth}
        \includegraphics[width=\textwidth]{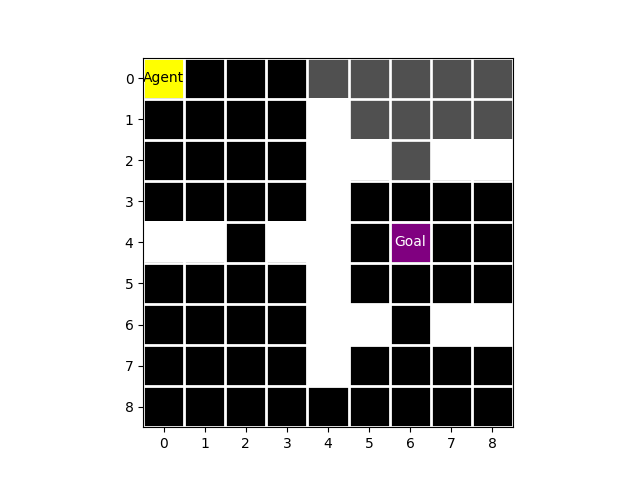} 
        \caption{}
        \label{fig:grid_pomdp}
    \end{subfigure}
    \hfill
    \begin{subfigure}{0.44\textwidth}
        \includegraphics[width=\textwidth]{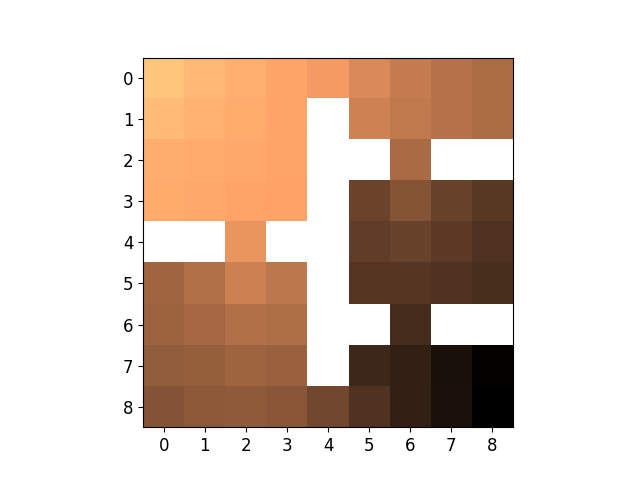} 
        \caption{}
        \label{fig:origin_pomdp}
    \end{subfigure}
    \begin{subfigure}{0.45\textwidth}
        \includegraphics[width=\textwidth]{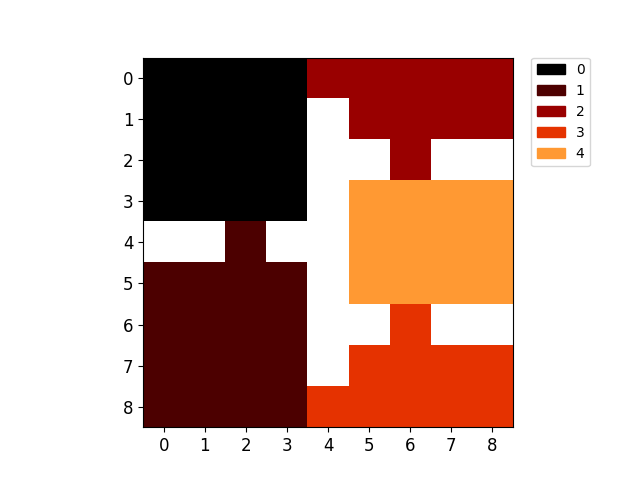}
        \caption{}
        \label{fig:macro_pomdp}
    \end{subfigure}
    \hfill
    \begin{subfigure}{0.5\textwidth}
        \centering
        \includegraphics[width=\textwidth]{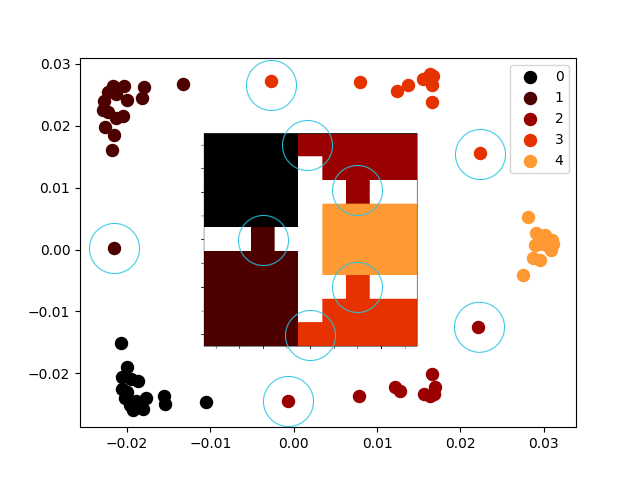}
        \caption{}
        \label{fig:cluster_pomdp}
    \end{subfigure}
    \caption{{\bf Task 3: POMDP Gridworld Environment with five rooms.}
    (a) A $9\times 9$ Gridworld with a single agent (yellow), a single goal (purple), walls (white) and high noise room (gray) with a reward of $100$ associated with the goal and overall observational noise parameter $\eta = 0.1$.
    (b) Successor representation of states with respect to initial state (brighter is closer).
    (c) Macro states learned by the model, which coincide with the $5$ rooms in the model.
    (d) Visualization of the $5$ macro state clusters in terms of projections of the micro states into an embedding space. Note that the corridors between pairs of the $5$ rooms have some separation compared to the rest of the clusters in the embedding space, as indicated by the blue circles. }
\label{fig:pomdp_five}
\end{minipage}
\end{figure}

\begin{figure}[!t]
\begin{minipage}[]{\textwidth}
    \begin{subfigure}{0.329\textwidth}
        \centering
        \includegraphics[width=\textwidth]{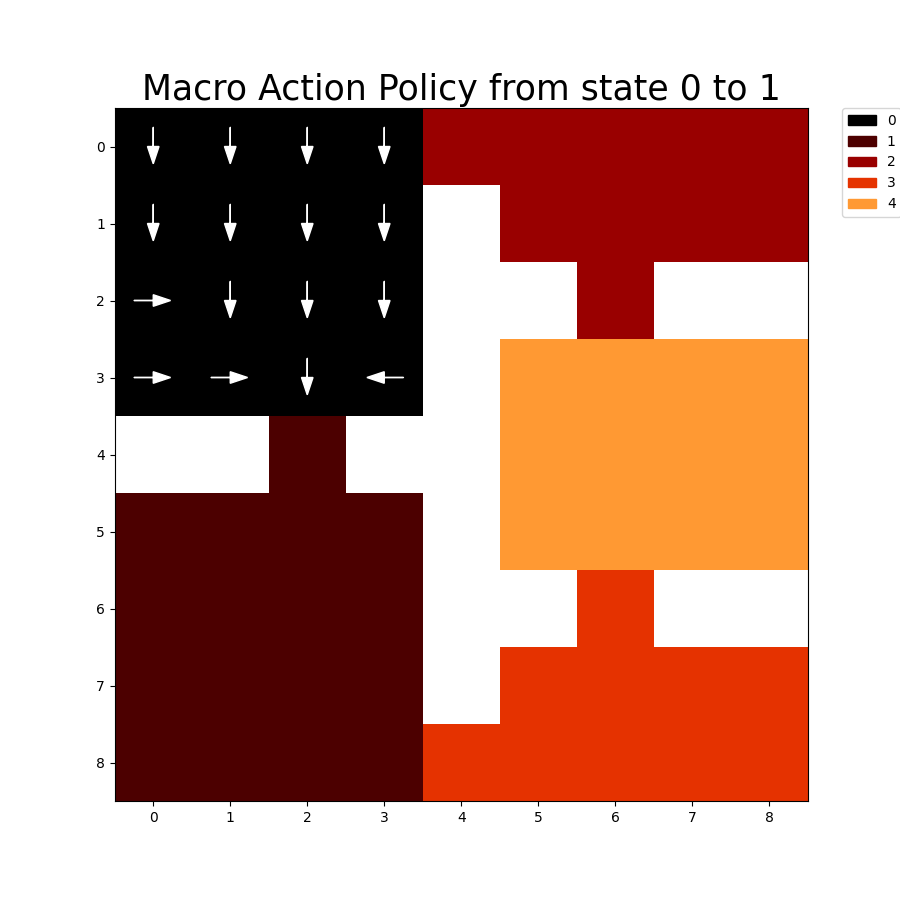}
        \caption{}
        \label{fig:MA_0_1_pomdp}
    \end{subfigure}
    \begin{subfigure}{0.329\textwidth}
        \centering
        \includegraphics[width=\textwidth]{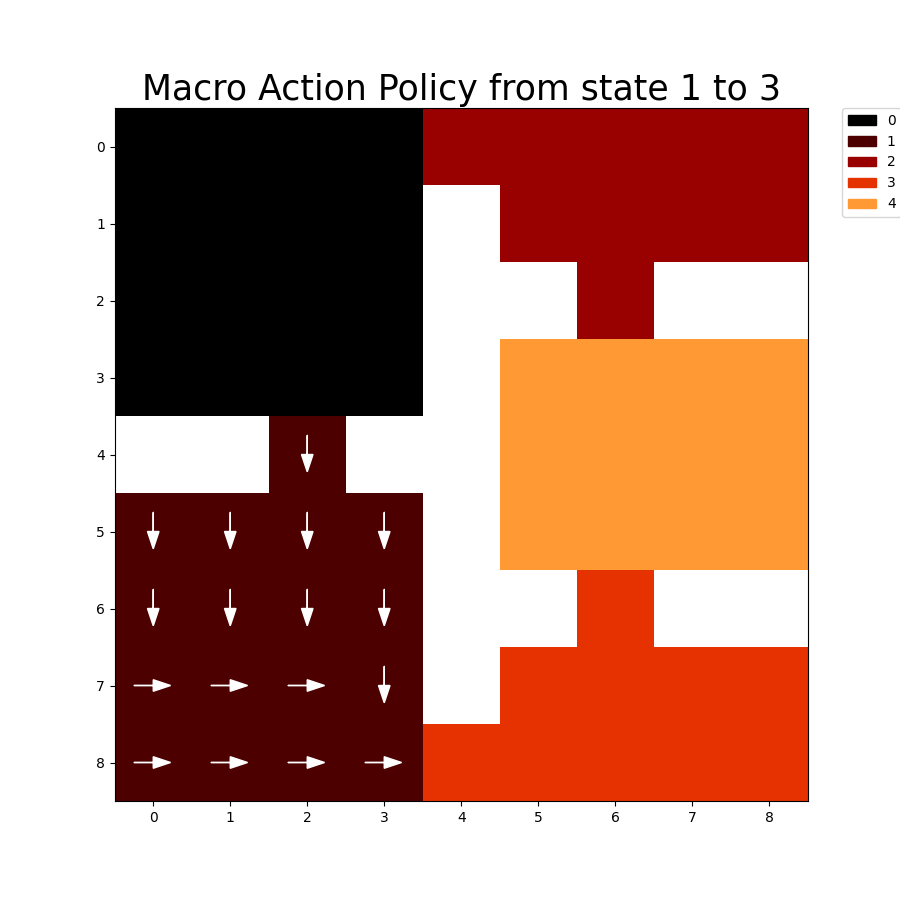}
        \caption{}
        \label{fig:MA_1_3_pomdp}
    \end{subfigure}
    \begin{subfigure}{0.329\textwidth}
        \centering
        \includegraphics[width=\textwidth]{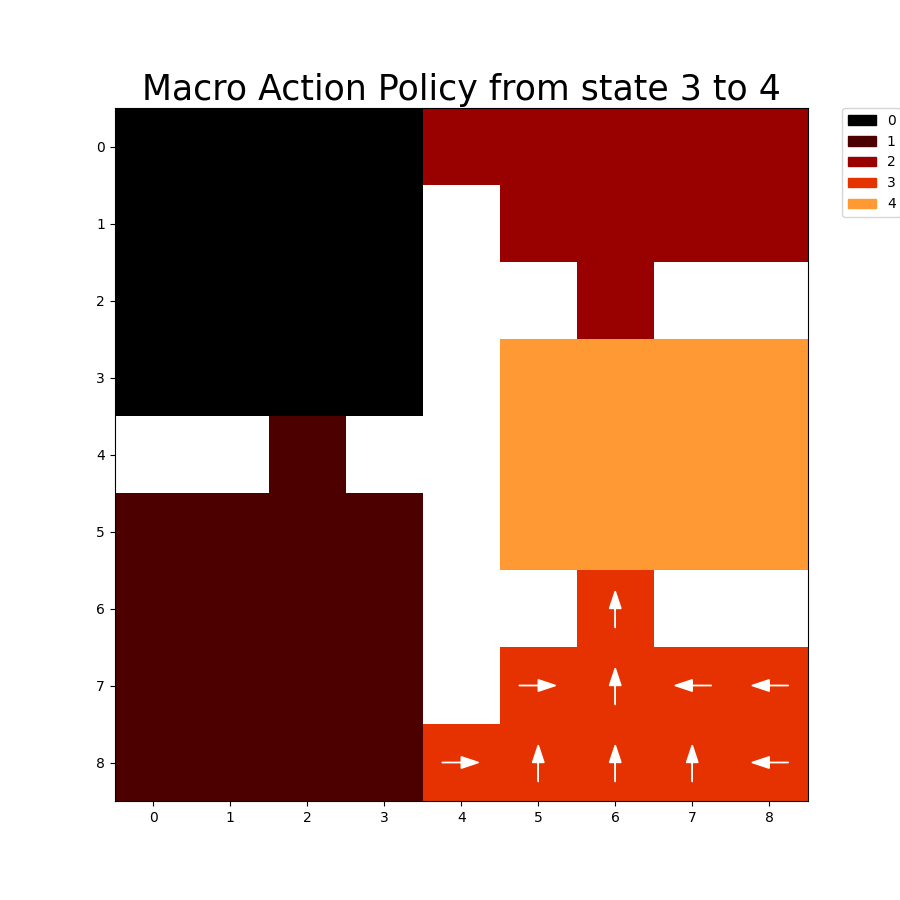}
        \caption{}
        \label{fig:MA_3_4_pomdp}
    \end{subfigure}
    \caption{{\bf Macro Actions for Task 3.}\\ (a)-(c) represent macro actions that execute a policy at the lower level for transitioning from one macro state to another (here, macro actions for $S_0\rightarrow S_1$, $S_1\rightarrow S_3$, and $S_3\rightarrow S_4$ are shown). The white arrows indicate the most probable action from each micro state in the first macro state in order to eventually reach the bottleneck state of the second macro state.}
\label{fig:MA_pomdp}
\end{minipage}
\end{figure}

Consider a Gridworld environment with five rooms, as illustrated in Figure~\ref{fig:schema} and shown again in Figure~\ref{fig:grid_pomdp} for convenience. Assume there is some small general observational noise ($\eta=0.1$) throughout the environment similar to what was described above. Additionally, we introduce comparatively larger noise distributed throughout the top right room in the environment (Figure~\ref{fig:grid_pomdp}). Figures~\ref{fig:origin_pomdp}-\ref{fig:cluster_pomdp} show the successor representation and clusters learned when we apply the hierarchical method. These clusters or macro states coincide with the $5$ rooms in the environment when we set the number of clusters to $5$. An interesting observation can be made in the embedding space of the micro states as seen in Figure~\ref{fig:cluster_pomdp}. Even though the states corresponding to the corridors between pairs of rooms are assigned to one of the room clusters, they are located separately in the embedding space, indicating that they serve as a bottleneck state to which the agent can navigate when executing a macro action. Figure~\ref{fig:MA_pomdp} depicts some sample macro actions learned by the agent to navigate from one macro state to another.

\begin{figure}[!t]
\begin{minipage}[]{\textwidth}\centering
\includegraphics[width=0.7\textwidth]{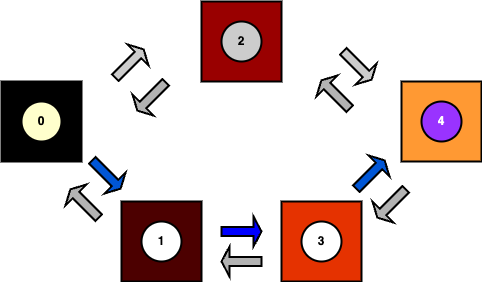}
\label{fig:Five_Rooms_Macro_Action_Planning}
\caption{{\bf Higher Level Planning for Task 3.} The arrows represent the potential macro actions between all pairs of adjacent macro states. The blue arrows represent the sequence of macro actions selected via higher-level active inference to get from macro state $0$ (containing the start state) to macro state $4$ (containing the goal state). Note that with active inference, the agent avoids the shorter path containing macro state $2$ due to its larger observational noise (higher uncertainty) and opts for the longer path instead.}
\label{fig:plan_pomdp}
\end{minipage}
\end{figure}

\begin{figure}[!t]
\begin{minipage}[]{\textwidth}
    \begin{subfigure}{0.45\textwidth}
        \includegraphics[width=\textwidth]{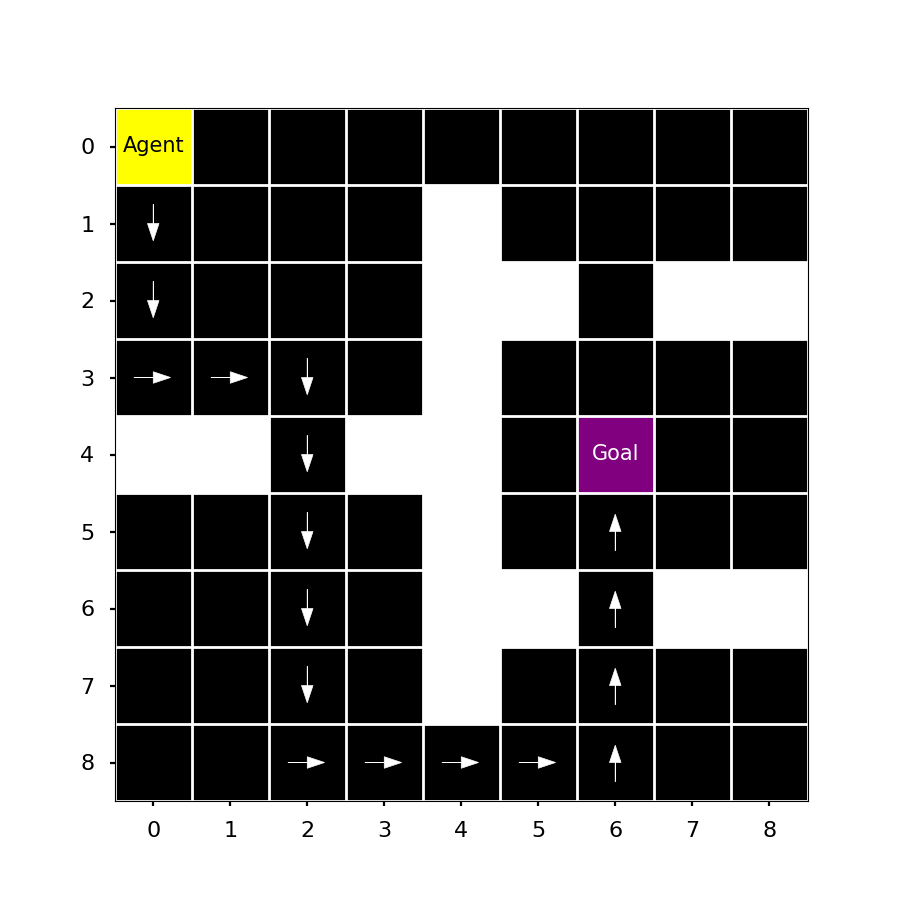} 
        \caption{}
        \label{fig:action_taken_2}
    \end{subfigure} 
        \hfill
    \begin{subfigure}{0.45\textwidth}
        \includegraphics[width=\textwidth]{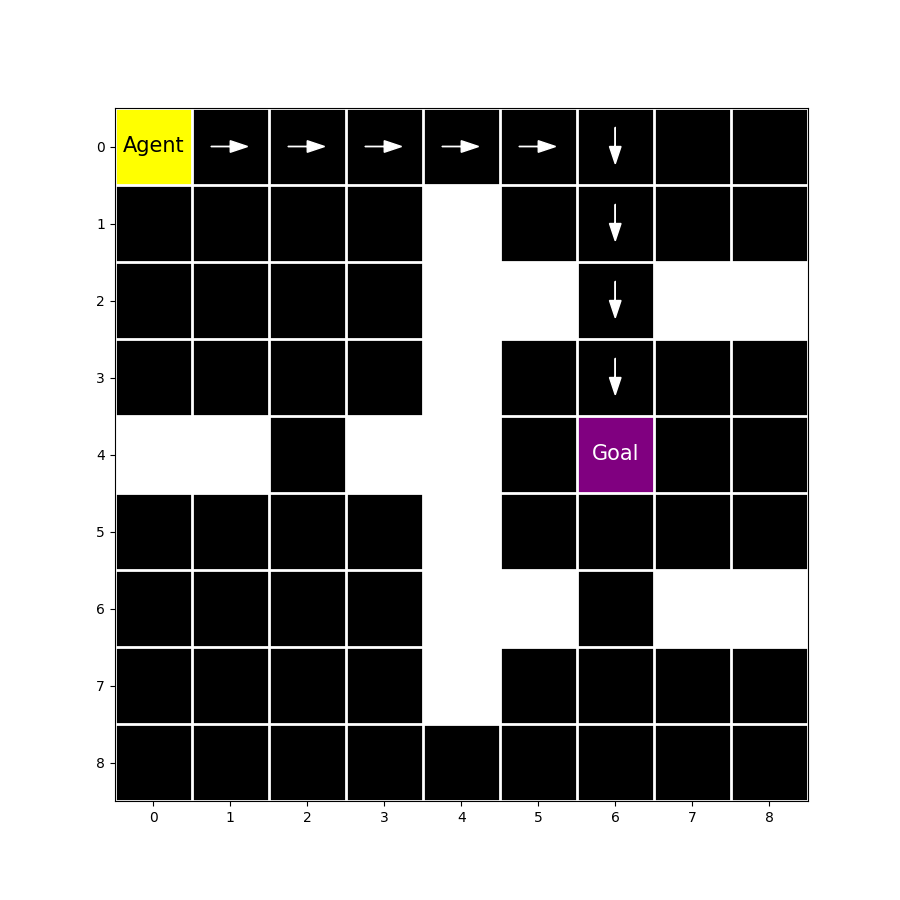}
        \caption{}
        \label{fig:action_taken_1}
    \end{subfigure}

    \caption{{\bf Active Inference optimizes the Tradeoff between Uncertainty and Reward in Task 3.} (a) Minimizing EFE using hierarchical active inference results in the agent naturally avoiding macro states with high noise (top right room), reaching the goal via the slightly longer path through two other rooms. (b) When the noise level in the top right room is decreased to the same noise level as the rest of the environment, the agent picks the shortest path to reach the goal.
    }
\label{fig:action_taken_noise}
\end{minipage}
\end{figure}

If the amount of noise introduced in the top right room is sufficiently high compared to the rest of the environment, we expect active inference to select a path that avoids the noisy room due to the uncertainty associated with it, even if it means accruing some negative reward in traversing a longer path to reach the goal. Figures~\ref{fig:plan_pomdp} and \ref{fig:action_taken_2} demonstrate that this is indeed the case: the agent opts for the longer path to the goal because it offers a lower EFE value, optimizing the tradeoff between the entropy of the noise in the top right room versus the decrease in reward due to the longer path. When the noise in the top right room is decreased to the overall noise level across the environment, active inference prefers the shorter path (Figure~\ref{fig:action_taken_1}).

\begin{figure}[!htbp]
    \centering
    \includegraphics[width=0.5\linewidth]{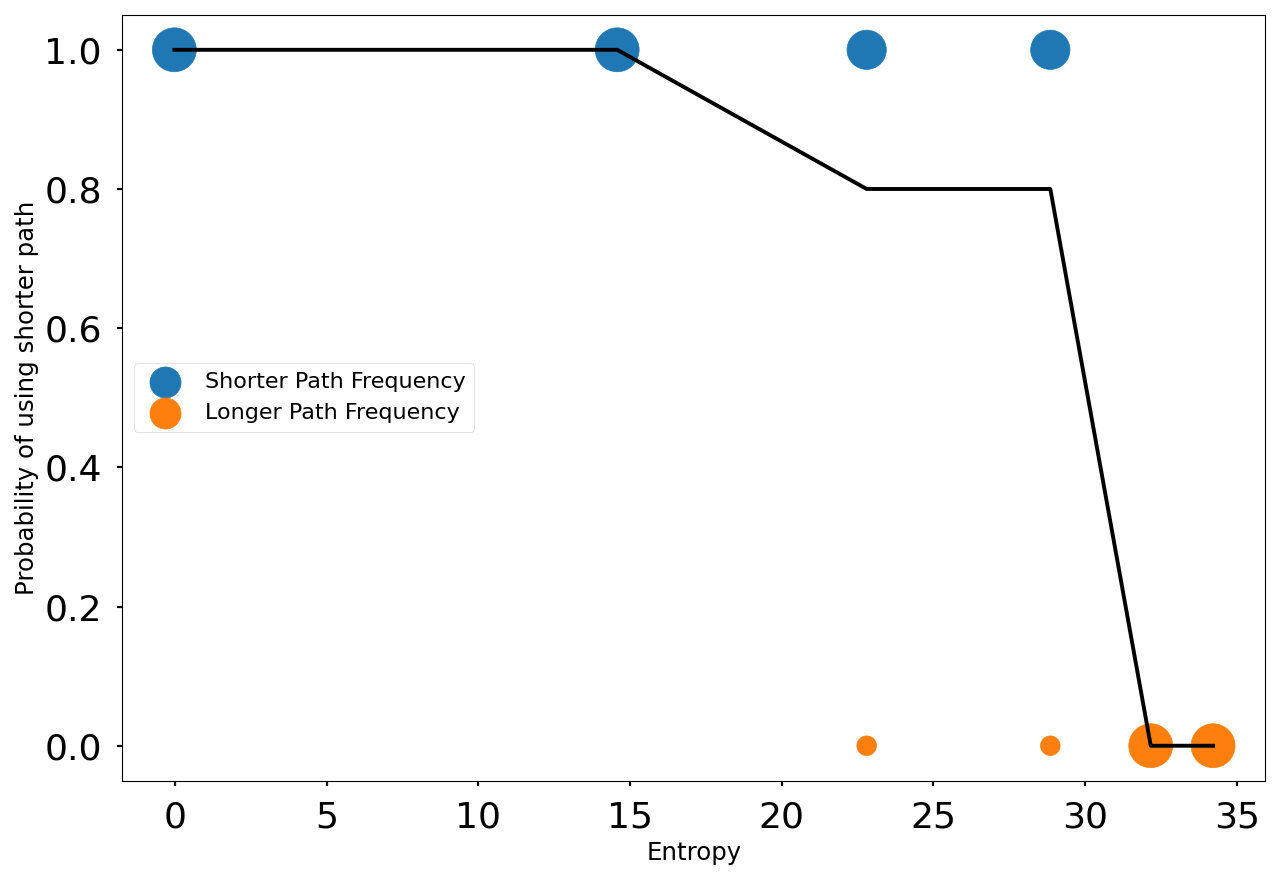}
    \caption{{\bf Tradeoff between Uncertainty and Reward as a Function of Noise in Task 3.} The plot shows the probability that the agent, using active inference, chooses the shorter path to the goal in Task 3 as the noise in the top right room is increased by increasing the entropy of the observation model for that room. As the entropy increases, the agent increasingly prefers the longer path over the shorter one.}
    \label{fig:graph_entropy}
\end{figure}

To explore the tradeoff optimized by active inference between avoiding high uncertainty versus maximizing rewards, we varied the level of noise in the top right room by progressively increasing the entropy of the observation model. Figure~\ref{fig:graph_entropy} shows that for lower levels of noise, the agent picks the shorter path to the goal but once the entropy crosses a particular threshold, the agent opts for the longer path to the goal. The threshold itself depends on many factors, such as the value of the rewards/penalty for an action at each step, the level of noise across the environment compared to the noise in the top right room, and the discount rate used to learn the successor matrix.

\subsection{Task 4: Mountain Car}
\label{sec:task4}
We now consider a more complex classic control problem from the OpenAI Gym suite, a benchmark collection of problems and environments widely used in reinforcement learning \parencite{brockman2016openaigym}. As in the Gridworld navigation problems above, this environment requires long horizon planning under delayed or sparse reward signals. In contrast to Gridworld however, the agent operates in a continuous state space, thus requiring additional considerations before our model can be used. Note that there is no observational noise and the task is therefore an MDP problem.

\begin{figure}[!htbp]
\begin{minipage}[t]{\textwidth}
    \centering

    \begin{subfigure}{\textwidth}
        \centering
        \includegraphics[width=0.65\textwidth]{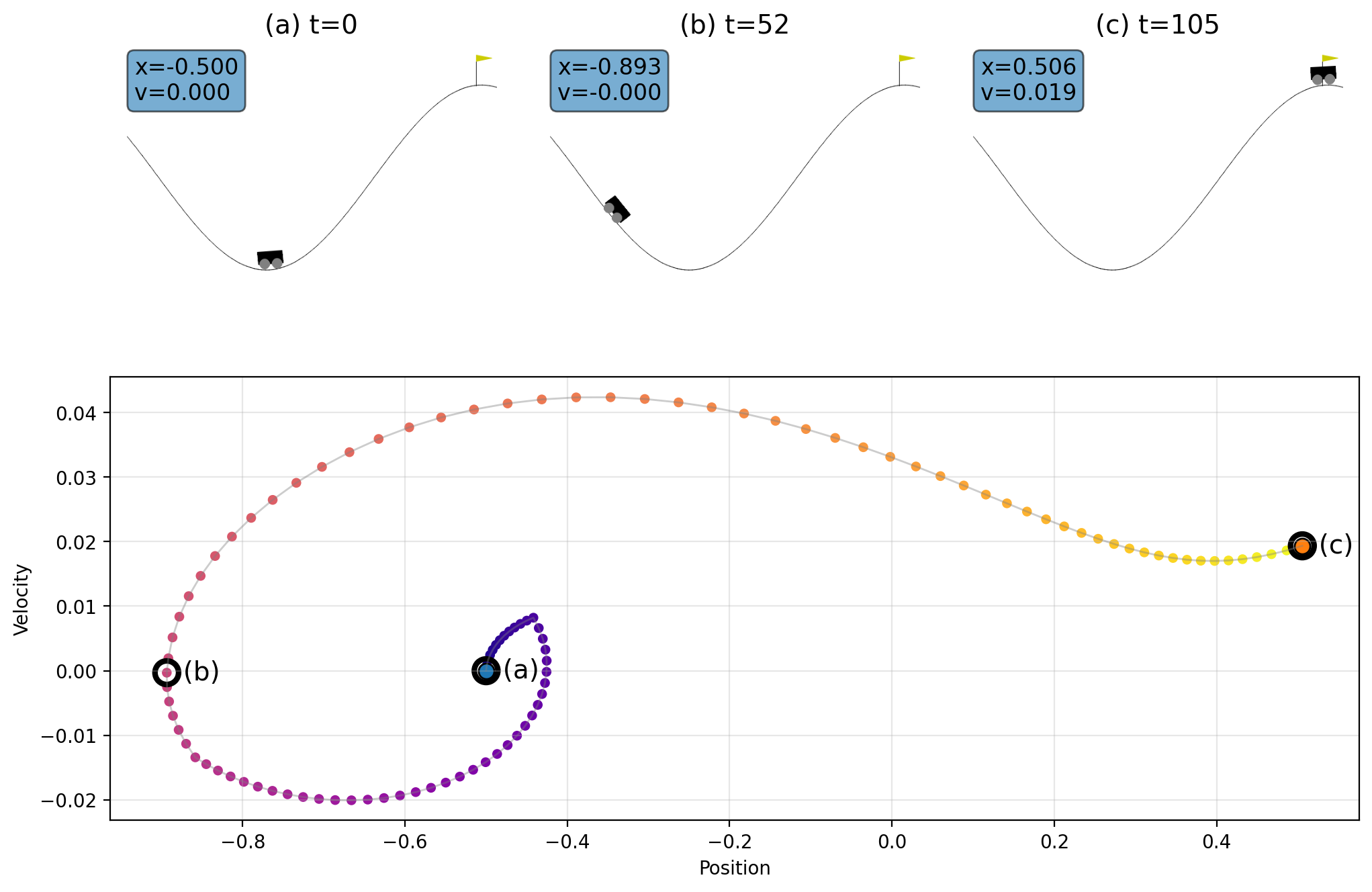}
        \caption{}
        \label{fig:MC_traj}
    \end{subfigure}

    %\vspace{0.1em}

    \begin{subfigure}{\textwidth}
        \centering
        \includegraphics[width=0.50\textwidth]{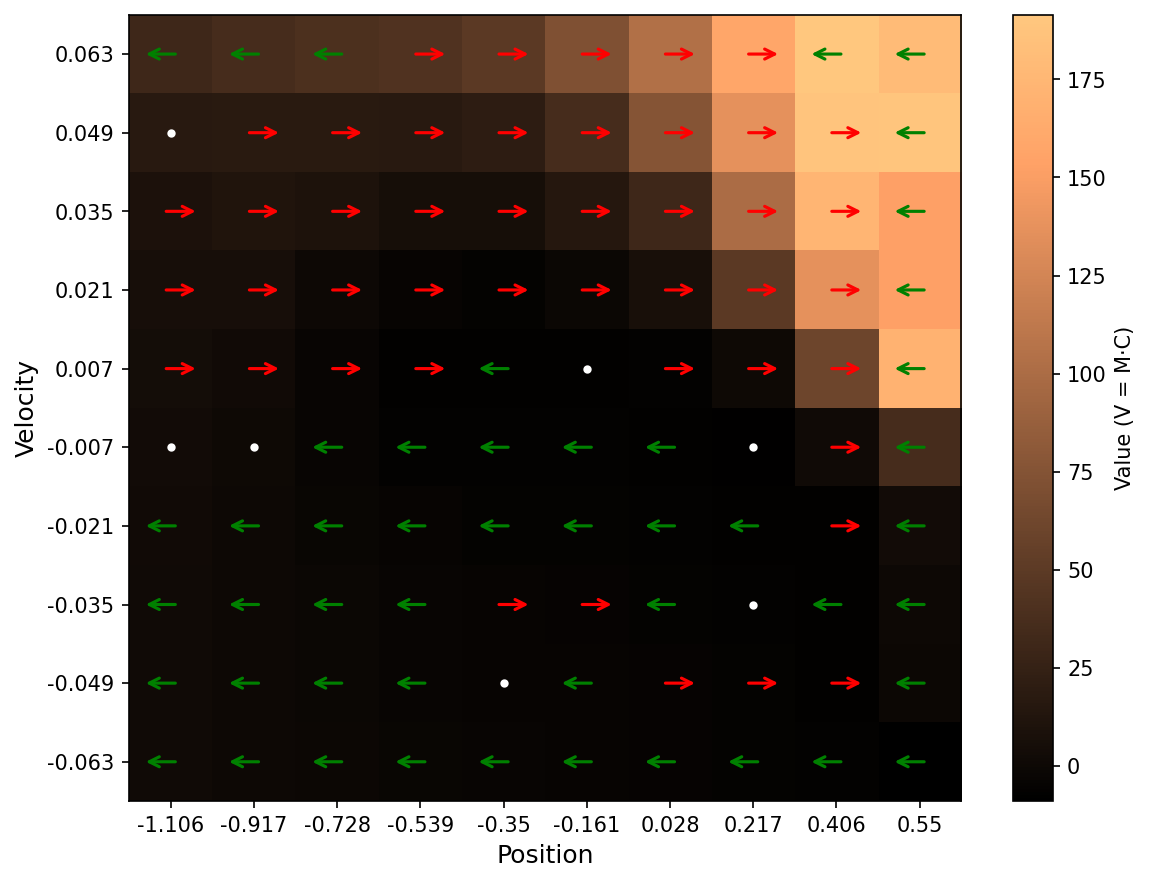}
        \caption{}
        \label{fig:MC_value}
    \end{subfigure}

    \caption{{\bf Mountain Car Problem and Value Function.}\\
    (a) (Top panels) Three key stages depicted as snapshots in the MountainCar problem. We show the agent at its start state, the leftmost turning point where it begins accelerating to the right, and the final goal location at the top right hill. (Bottom panel). Trajectory of a successful run plotted in position-velocity state space, with the three stages shown in the top panels marked on the trajectory.
    (b) Learned value function over the state space (brighter colors indicate higher value), together with the optimal action at each micro state. Left and right arrows denote acceleration directions, while a dot represents no acceleration. States closest to the goal exhibit the highest values.}

    \label{fig:MC_combined}
\end{minipage}
\end{figure}

In this task, a cart is at the bottom of a valley between two hills (Figure~\ref{fig:MC_traj} top left panel, $t=0$). The goal is to reach the top of the hill on the right side but the agent lacks the acceleration required in order to drive directly to the goal. Instead, the agent must exploit the environment’s passive dynamics by first moving away from the goal (towards the left) to build up sufficient kinetic energy before ascending the hill on the right, as depicted in Figure~\ref{fig:MC_traj} (top three panels).

The state space is continuous and 2-dimensional, with position $x\in [-1.2,0.6]$ and velocity $v\in [-0.07, 0.07]$. If the cart reaches $x=-1.2$, i.e., the leftmost point in Figure~\ref{fig:MC_traj} (top plots), the velocity is set to~$0$. The action space is discrete and can be one of three options: accelerate a fixed amount to the left, accelerate the same fixed amount to the right, or no acceleration. The environment dynamics are deterministic, with each action (fixed acceleration to the left or right) resulting in smooth but nonlinear state transitions due to gravity. An episode terminates when the cart position exceeds a threshold of $x\geq 0.5$, which corresponds to going past the top of the right hill, or when a maximum time horizon ($200$ steps) is reached. The reward signal is sparse, with the agent receiving a reward of $0$ when the cart reaches the top of the right hill ($x\geq 0.5$) and a reward of~$-1$ at each time step until termination, encouraging solutions that reach the goal in as few steps as possible.

Note that this environment is well suited for evaluating hierarchical and long-horizon planning methods, as the optimal policy requires non-greedy behavior and temporally extended action sequences.

Although the Mountain Car environment is defined over a continuous state space, our successor-based method requires a discrete representation. This discretization enables the method to construct the hierarchical representations via clustering. Hence, we partition the continuous position and velocity dimensions into a fixed number of bins, yielding a discrete grid of micro-states that approximate the underlying dynamics. In the experiments, the number of position and velocity bins were both set to $10$. As a result, the discretized state space consists of $100$ micro-states. 

\begin{figure}[!htbp]
\centering

\setlength{\tabcolsep}{2pt} % (optional, if you want tighter spacing)

\begin{subfigure}{0.45\textwidth}
    \centering
    \includegraphics[width=\textwidth]{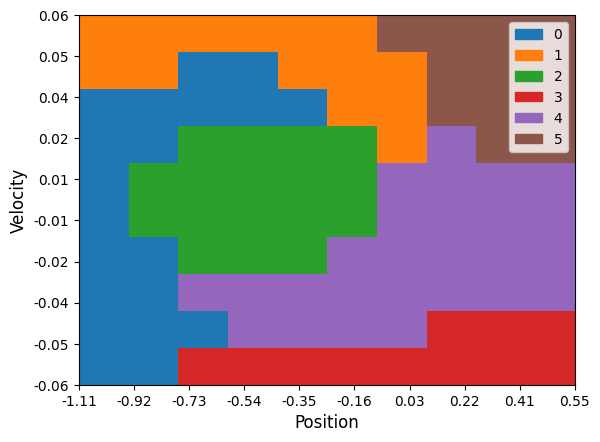}
    \caption{}
    \label{fig:MC_clust}
\end{subfigure}
\hfill
\begin{subfigure}{0.45\textwidth}
    \centering
    \includegraphics[width=\textwidth]{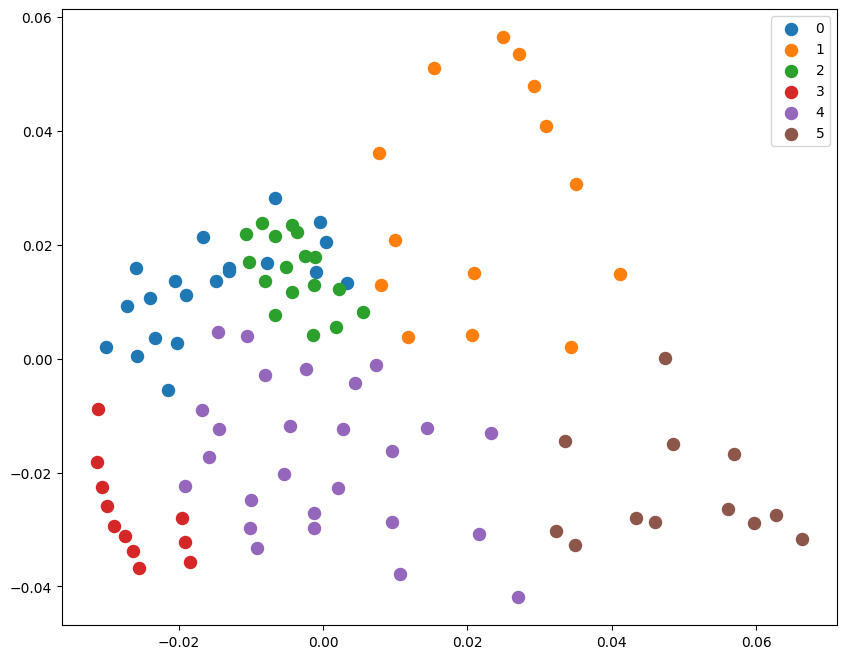}
    \caption{}
    \label{fig:MC_clust_viz}
\end{subfigure}

\vspace{4pt} % reduce / adjust vertical gap
\begin{subfigure}{0.45\textwidth}
    \centering
    \includegraphics[width=\textwidth]{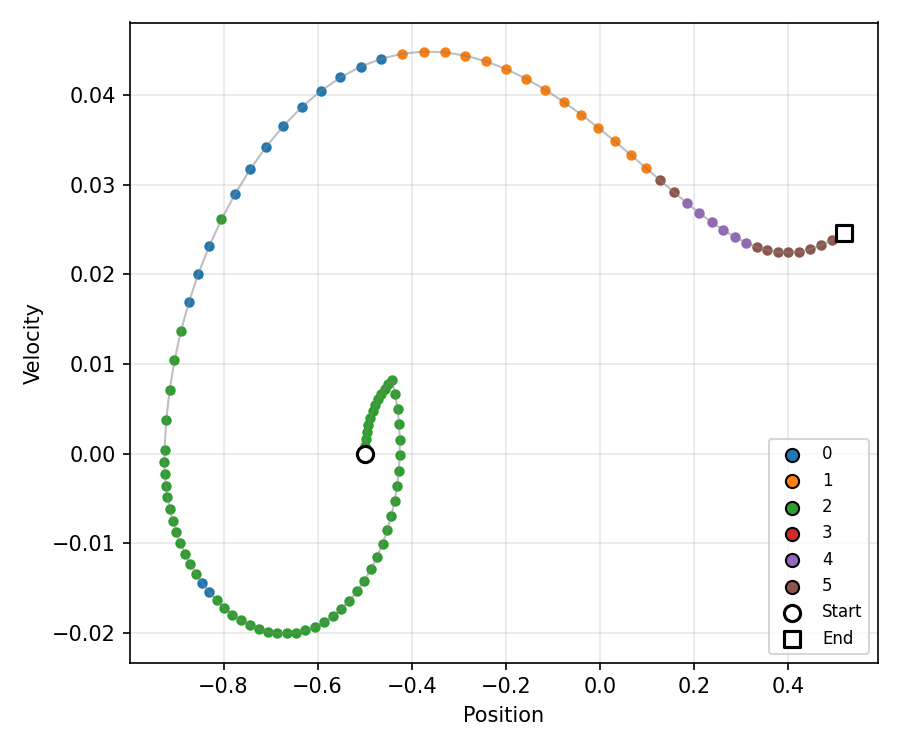}
    \caption{}
    \label{fig:MC_traj_macro}
\end{subfigure}
\hfill
\begin{subfigure}{0.45\textwidth}
    \centering
    \includegraphics[width=\textwidth]{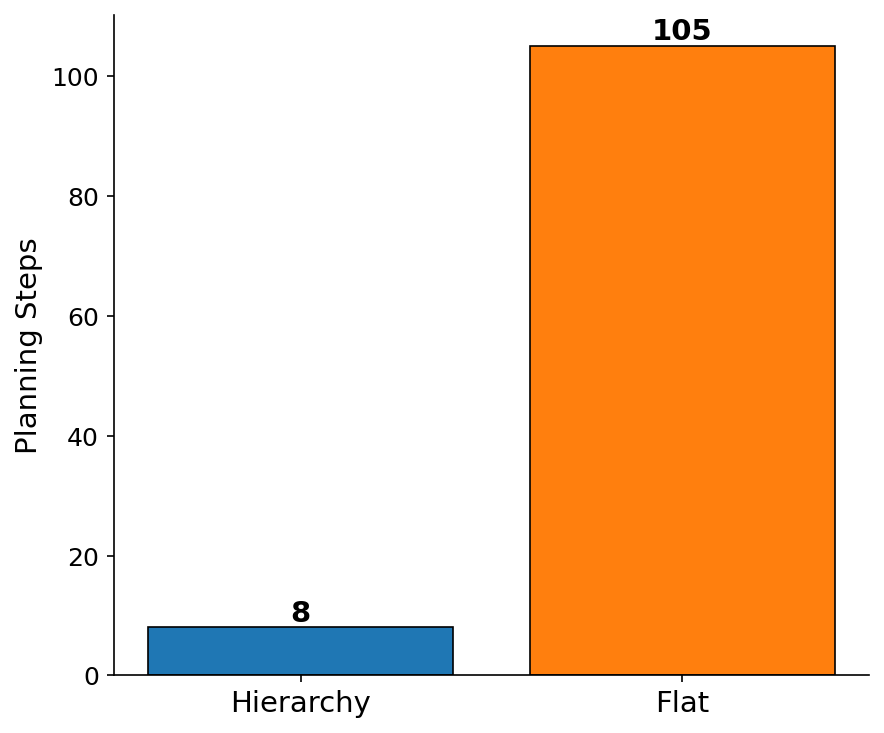}
    \caption{}
    \label{fig:MC_hier_flat}
\end{subfigure}

\caption{{\bf Hierarchical Active Inference for Mountain Car.}\\
(a) \textbf{State Space Clustering}. The $100$ micro states generated from a discretization of the continuous Mountain Car state space were clustered into $6$ macro states using spectral embedding. 
(b) \textbf{Clustering Embeddings}. Visualization of the macro states in the spectral embedding space (arbitrary units).
(c) \textbf{Successful Trajectory obtained using Hierarchical Active Inference}. Trajectory shown in position-velocity state space with each micro state data point traversed by the mountain car colored according to its macro state cluster. Active inference rocks the car back and forth, allowing it to reach the top of the right hill (endpoint of trajectory).
(d) \textbf{Hierarchical vs.\ Flat Active Inference}. Hierarchical active inference requires substantially fewer action selection steps.
}
\label{fig:MC_macro}
\end{figure}

In our hierarchical active inference method, the successor representation serves a dual purpose: it is used to both construct a free energy-based value function (Figure~\ref{fig:MC_value}) as well as extract hierarchical structure through state clustering, as shown in Figure~\ref{fig:MC_clust}. In the discrete Gridworld setting used in previous sections, we were able to employ the successor representation over the complete state space to achieve these two goals. However, for the continuous-valued Mountain Car problem, we found that there were often outliers formed in the clusters and the clusters are not as stable as in the Gridworld case. We therefore used a radial basis function (RBF) kernel as a regularizer (see Appendix~\ref{sec:smooth_M} for details).

% However, in the case of the MountainCar environment, we discovered that constructing clusters using the full state space led to increased instability. A key reason is that the velocity component of the state changes rapidly and introduces high short-term variability, which adversely affects the stability of the clustering process. 
% To address this, we simplify the clustering stage by using only the position dimension of the state to cluster the model by averaging out the successor representation for different velocities and the same position. More concretely, all states that share the same position but differ in velocity get assigned to the same cluster, as can be observed in Figure~\ref{fig:MC_clust} This produces more stable and interpretable higher-level abstractions. Also, in practice, the position component is more useful in terms of clustering, with even the goal state being defined only in terms of position. During lower level planning, active inference still is performed over the full state space, and the value function derived is using the successor representation that includes both position and velocity. Thus, the hierarchy has a more robust smaller state representation built over a fine-grained lower state representation. 

To promote stability and robust exploration and learning, we used an exploration strategy where the agent randomly selected a micro action (fixed acceleration to the left/right or no acceleration) and executed it $k$ times in a row (we used $k = 5$); this effectively implemented a simple type of macro action at a level above the basic micro action. We then learned a set of higher level macro actions composed of these intermediate fixed macro actions, i.e., the learned macro actions from one macro state to another were sequences of these intermediate macro actions. To learn the macro states, we applied spectral clustering (as in previous sections) to the learned successor representation over the micro states; the results are shown in Figure~\ref{fig:MC_clust} and \ref{fig:MC_clust_viz}. 

We used the learned macro states and macro actions to perform hierarchical active inference to solve the Mountain Car control problem. Figure~\ref{fig:MC_traj_macro} shows a successful trajectory in the 2D position-velocity state space of the mountain car, with the color of the data points indicating how the method was able to infer the correct sequence of macro actions to drive the mountain car from one macro state to another, ultimately reaching the top of the right hill. Figure~\ref{fig:MC_hier_flat} visually depicts the difference in the number of planning steps used by the flat versus hierarchical active inference method: the hierarchical method requires only 8 planning steps compared to the 105 required by the flat model.
%%%====================================================================
%%% Task 5: PointMaze
%%% ====================================================================
\subsection{Task 5: PointMaze}
\label{sec:pointmaze}
To further explore the applicability of our hierarchical active inference method to problems with both a continuous state space and a continuous action space, we tested performance on PointMaze, a 2D maze navigation task from the Gymnasium-Robotics library \parencite{gymnasium_robotics} that is built on the MuJoCo physics engine. PointMaze can be regarded as a continuous version of the Gridworld task, requiring planning through spatially complex environments with walls, corridors, and rooms.

The agent controls a point mass confined to a planar maze with rigid walls. The state space consists of a 4D vector $(x, y, v_x, v_y)$ denoting the agent's position and velocity, plus a desired goal position $(x_g, y_g)$. The action space is a continuous 2D force vector $\mathbf{u} \in [-1, 1]^2$, applied at each physics simulation step, with the point mass moving approximately $0.0024$ units per step, a movement that is extremely fine-grained relative to the maze's scale.

\begin{figure}[!t]
    \centering
    \begin{subfigure}{0.4\textwidth}
        \centering
        \includegraphics[width=\textwidth]{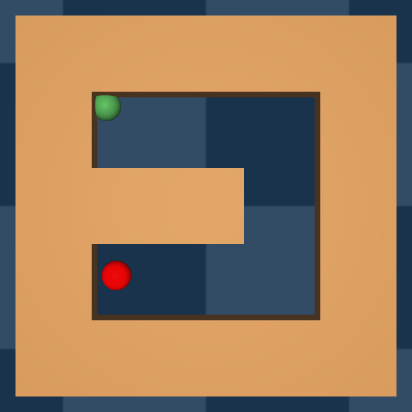}
        \caption{}
    \end{subfigure}
    \hfill
    \begin{subfigure}{0.4\textwidth}
        \centering
        \includegraphics[width=\textwidth]{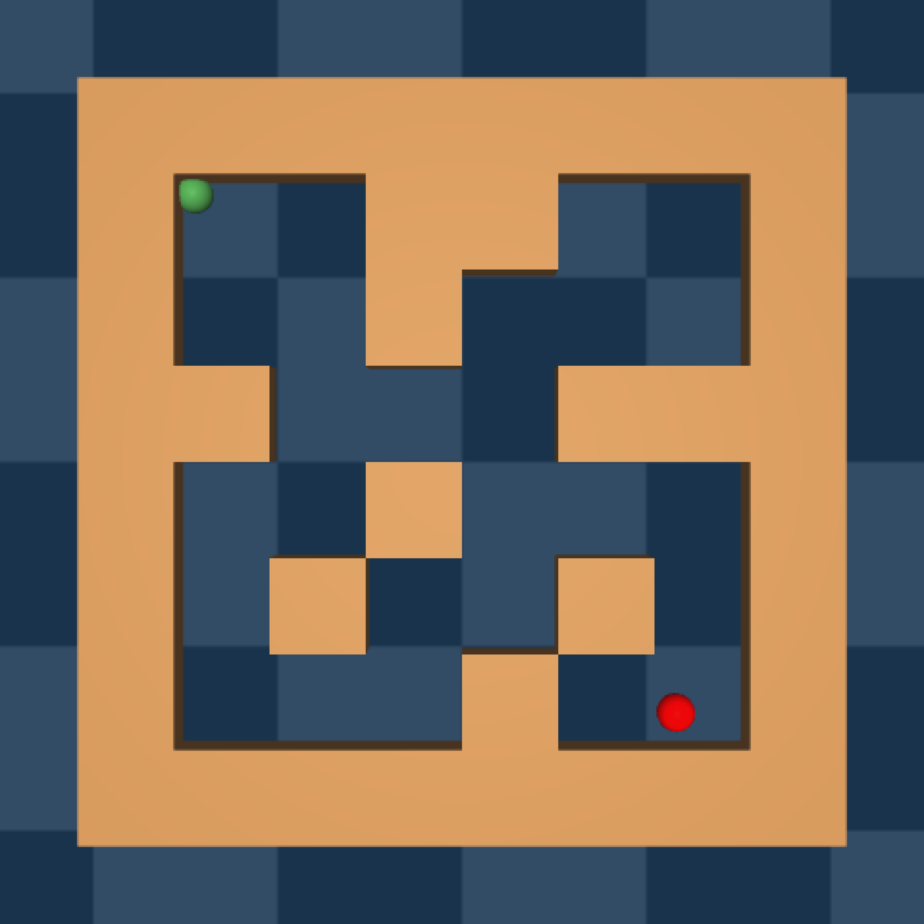}
        \caption{}
    \end{subfigure}

    \vspace{0.5em}

    \begin{subfigure}{0.45\textwidth}
        \centering
        \includegraphics[width=\textwidth]{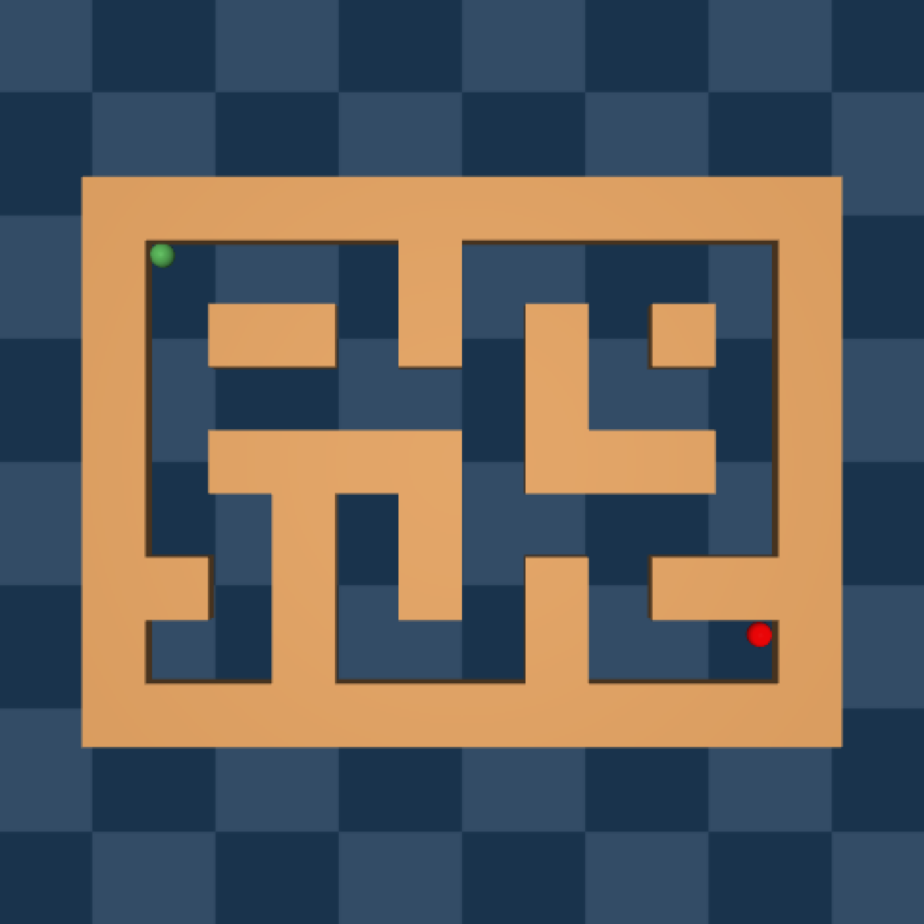}
        \caption{}
    \end{subfigure}
    \caption{{\bf PointMaze Variants.} (a) UMaze ($5 \times 5$), (b) Medium ($8 \times 8$), and (c) Large ($9 \times 12$). The green ball is the agent (point mass) and the red ball marks the goal location. Walls (brown) partition the continuous space into rooms connected by narrow corridors. Actions are continuous as well (see text for details).}
    \label{fig:PM_environments}
\end{figure}

We evaluated three maze layouts, summarized in Table~\ref{tab:PM_maze_variants} and illustrated in Figure~\ref{fig:PM_environments}. Each maze variant is configured with multiple goal locations for our re-planning experiments: 5 goals for UMaze and Medium, and 6 goals for Large, spanning all rooms and corridors. A goal is considered reached when the agent's position is within $0.35$ units of the goal's center.

\begin{table}[ht]
\centering
\caption{PointMaze: Variant Maze Specifications. \emph{Grid} refers to the maze cell layout; \emph{Bins} to the discretization resolution; \emph{Navigable} to the number of non-wall bins (approximate).}
\label{tab:PM_maze_variants}
\begin{tabular}{lcccc}
\toprule
\textbf{Variant} & \textbf{Grid} & \textbf{Bins} & \textbf{Total States} & \textbf{Navigable States} \\
\midrule
UMaze & $5 \times 5$ & $20 \times 20$ & 400 & $\sim$220 \\
Medium & $8 \times 8$ & $32 \times 32$ & 1024 & $\sim$416 \\
Large & $9 \times 12$ & $48 \times 36$ & 1728 & $\sim$736 \\
\bottomrule
\end{tabular}
\end{table}

PointMaze is well suited for evaluating our hierarchical framework for several reasons. First, like Gridworld, the maze walls partition the navigable space into rooms connected by narrow passages, the kind of bottleneck structure that spectral clustering of the successor representation is well suited to discover. 
% Unlike MountainCar (where clusters correspond to energy-level regions in phase space), PointMaze clusters correspond to \emph{physical rooms}, providing an intuitive validation of the hierarchical decomposition. 
Second, the three maze variants provide a natural difficulty gradient: UMaze ($\sim$220 navigable states, 4 clusters) allows rapid iteration and can potentially be solved by both flat and hierarchical approaches, while Large ($\sim$736 states, 12 clusters) may especially benefit from a hierarchical approach. Third, PointMaze provides an easy way to test re-planning capability (by selecting new goal locations), affording an opportunity to demonstrate zero-cost re-planning to new goals given the successor representation's reward-independence.

As in our Mountain Car task, we discretize the continuous state space into spatial bins, with the continuous 2D force action space mapped to 8 discrete directions. We also employed a \emph{smooth stepping} mechanism (similar to Mountain Car) that allows the agent to repeat each action for multiple physics steps until a bin transition occurs. This helps bridge the gap between the fine-grained physics ($\sim$0.0024 units/step) and our coarse spatial bin resolution ($\sim$0.25 units/bin). Complete details of the discretization, smooth stepping, and hyperparameters are provided in Appendix~\ref{sec:pointmaze_details}.

\noindent{\bf Single Goal Planning}
%\label{sec:PM_planning}

\begin{figure}[!t]
    \centering
    \begin{subfigure}{0.45\textwidth}
        \centering
        \includegraphics[width=\textwidth]{figures/PointMaze/umaze_environment.png}
        \caption{}
    \end{subfigure}
    \hfill
    \begin{subfigure}{0.45\textwidth}
        \centering
        \includegraphics[width=\textwidth]{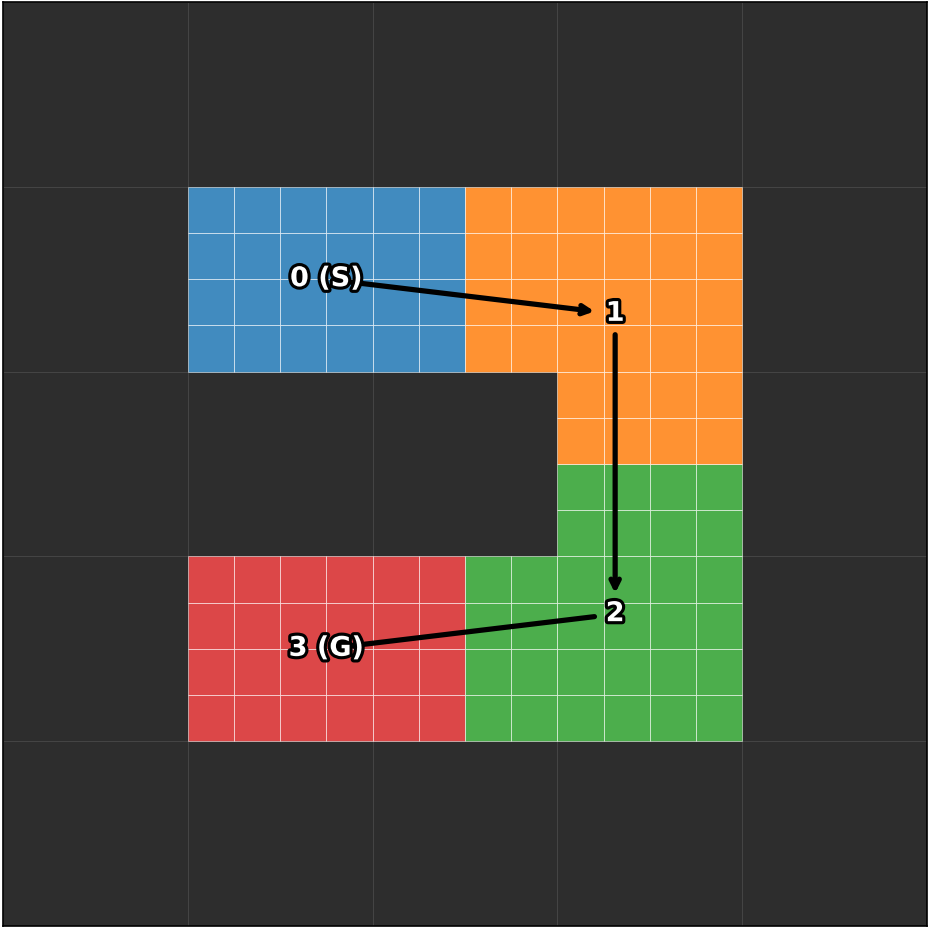}
        \caption{}
    \end{subfigure}
    \caption{{\bf UMaze and Macro State Clusters.} (a) UMaze environment. (b) Four clusters (colored regions) discovered by spectral clustering of the successor representation, overlaid on the discretized maze layout (dark areas denote walls). Arrows show the macro level policy directing the agent from each macro state toward the goal. The start (S) and goal (G) macro states are also marked on the figure.}
    \label{fig:PM_clusters}
\end{figure}

\noindent Figure~\ref{fig:PM_clusters} shows the macro state clusters discovered by spectral clustering of the learned successor representation on UMaze. The four clusters correspond to the distinct spatial regions of the U-shaped maze and very naturally arise from the maze's spatial geometry. 

\begin{figure}[!htbp]
    \centering
    \includegraphics[width=0.5\textwidth]{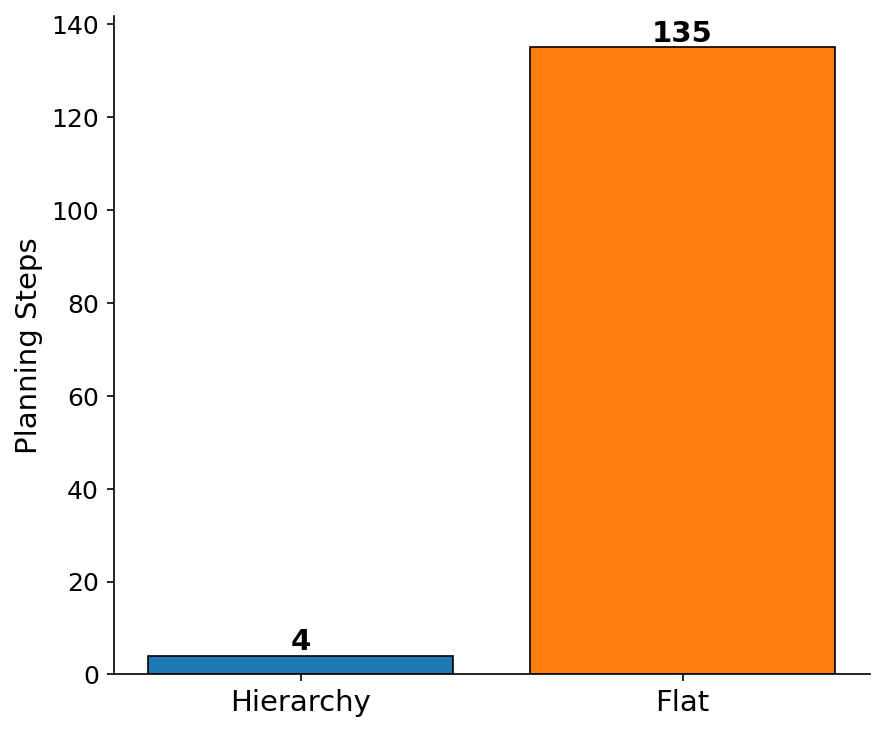}
    \caption{{\bf UMaze: Hierarchical vs.\ Flat Active Inference Performance Comparison.}}
    \label{fig:PM_planning}
    % \begin{subfigure}{0.45\textwidth}
    %     \centering
    %     \includegraphics[width=\textwidth]{figures/PointMaze/umaze_planning_cost.png}
    %     \caption{Re-planning cost (MACs)}
    %     \label{fig:PM_plan_cost}
    % \end{subfigure}
\end{figure}

Figure~\ref{fig:PM_planning} quantifies the active inference planning efficiency gap: both hierarchical and flat agents can reach the goal on UMaze, but the hierarchical agent does so in just 4 planning steps while the flat agent requires 135 steps ($\sim$34$\times$ more). This efficiency gap widens dramatically on larger mazes: the flat agent fails outright on the Large maze (Appendix~\ref{sec:PM_single_goal_extended}), where value gradients in the flat representation become too attenuated for the greedy policy to follow. Single-goal planning results for the Medium and Large variants, where the flat agent's failure is more pronounced, are provided in Appendix~\ref{sec:PM_single_goal_extended}.

\begin{figure}[!t]
    \centering
    \includegraphics[width=0.7\textwidth]{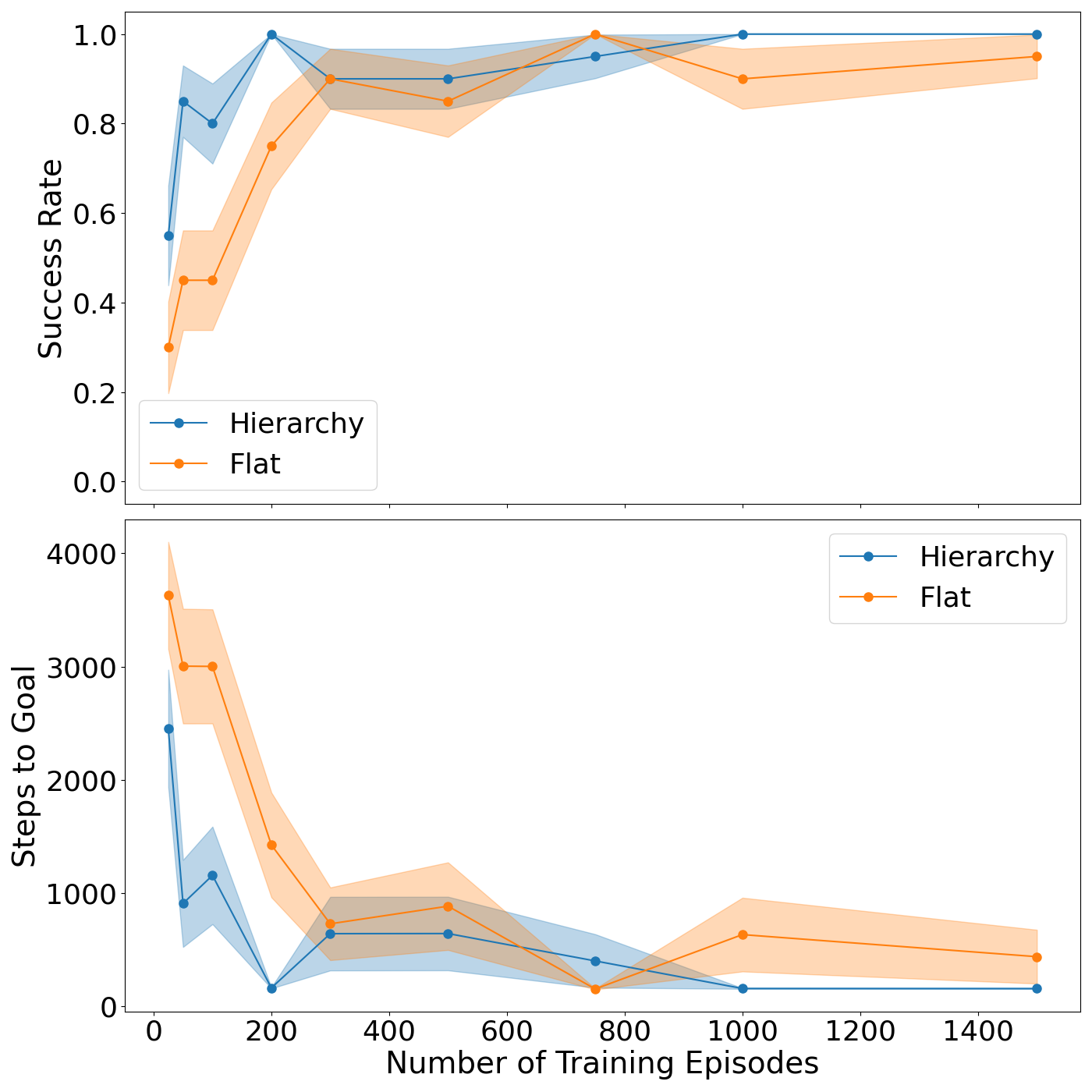}
    \caption{{\bf UMaze Learning Curves.} The plots show mean $\pm$ SEM over 20 seeds. Top: Success rate. Bottom: Micro level steps taken by the policy (capped at 5000). The hierarchical agent (blue) reaches 100\% success by 200 episodes and reaches the goal in ${\sim}$150 steps; the flat agent (orange) lags behind in both metrics and settles with higher variance.}
    \label{fig:PM_eval_curves}
\end{figure}

Figure~\ref{fig:PM_eval_curves} presents learning curves for UMaze, averaged over 20 random seeds. The hierarchical agent reaches 100\% success rate by 200 episodes and is able to reach the goal using ${\sim}$150 steps, largely because the macro level abstraction is robust to errors in the partially-learned successor representation. The flat agent eventually succeeds on UMaze given sufficient training episodes, but converges slower and with higher variance in both success rate and steps to goal.

\noindent{\bf Multi-Goal Re-Planning}
%\label{sec:PM_replan}

\begin{figure}[!t]
    \centering
    \begin{subfigure}{0.48\textwidth}
        \centering
        \includegraphics[width=\textwidth]{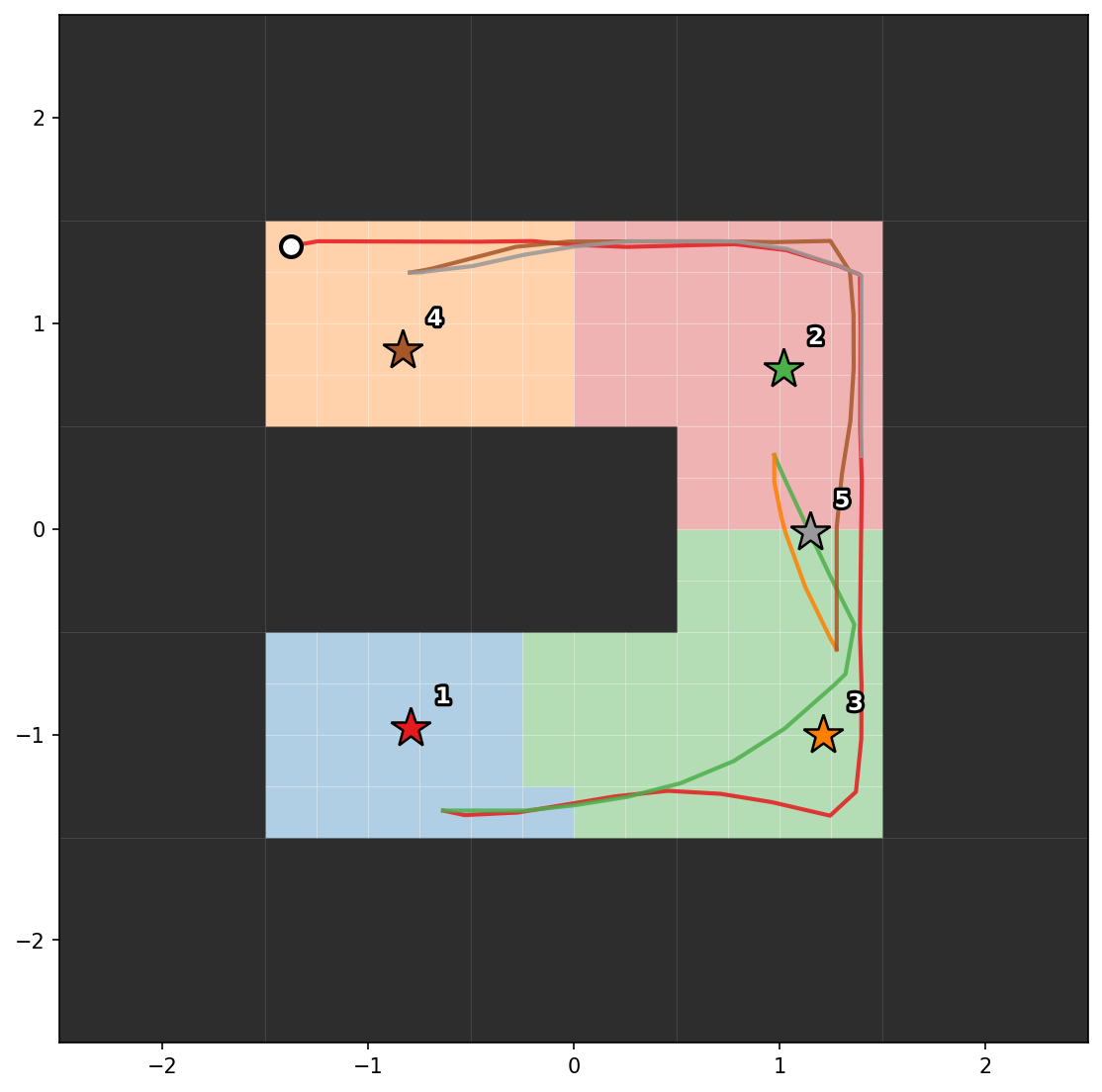}
        \caption{}
    \end{subfigure}
    \hfill
    \begin{subfigure}{0.48\textwidth}
        \centering
        \includegraphics[width=\textwidth]{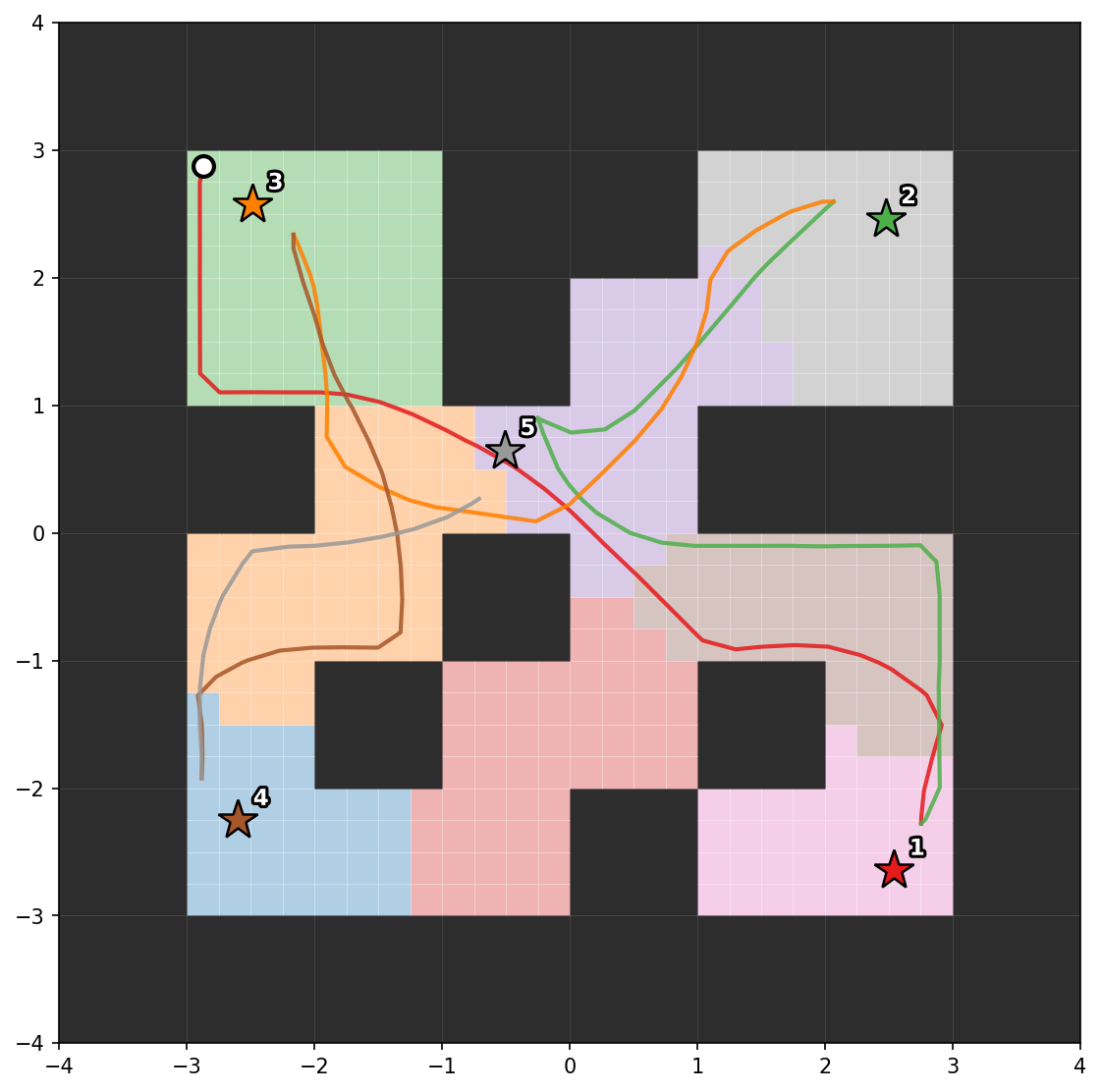}
        \caption{}
    \end{subfigure}

    \vspace{0.5em}

    \begin{subfigure}{0.60\textwidth}
        \centering
        \includegraphics[width=\textwidth]{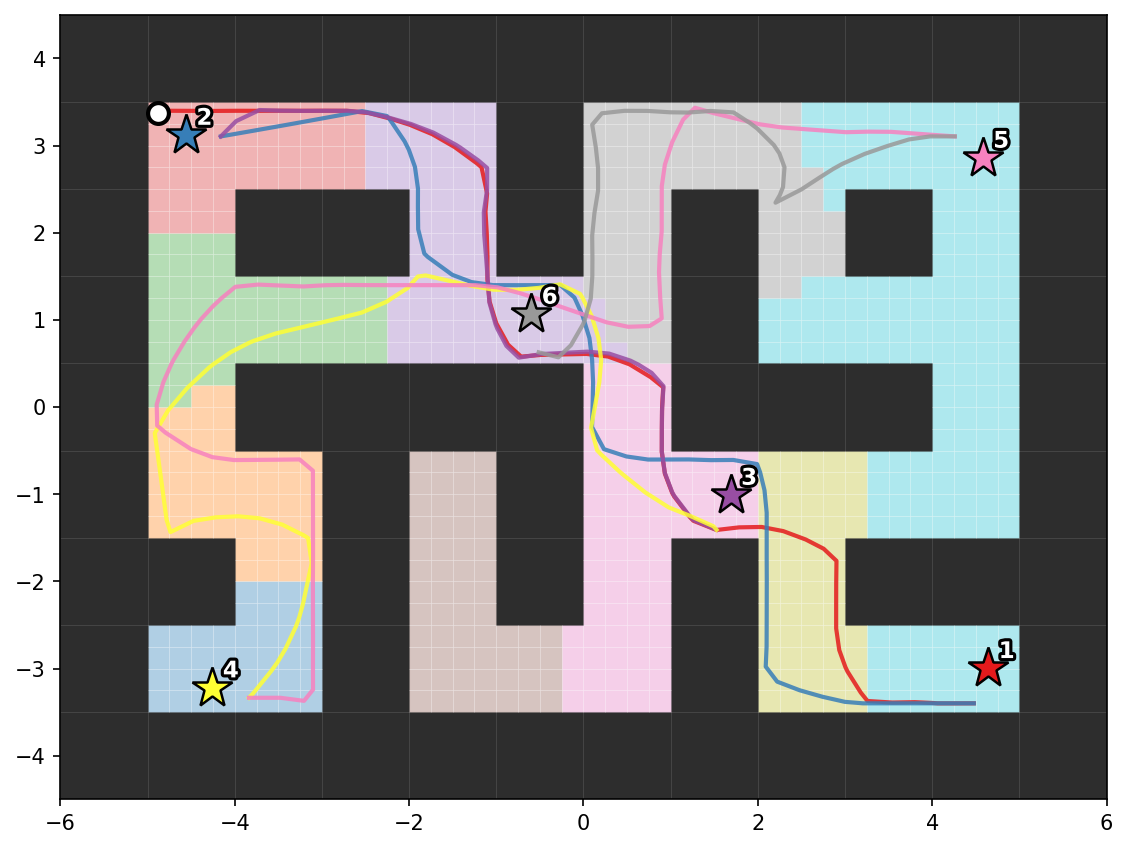}
        \caption{}
    \end{subfigure}
    \caption{{\bf PointMaze: Multi-Goal Re-Planning for Maze Variants.} The agent learns the environment once (i.e., learns the successor representation), then navigates to a sequence of goals without re-learning. (a) UMaze with 5 goals. (b) Medium with 5 goals. (c) Large with 6 goals.}
    \label{fig:PM_replan_all}
\end{figure}

\noindent The reward-independence of the successor representation enables zero-cost re-planning: when the goal changes, only the priors $C$ and $C_{\text{macro}}$ (Figure~\ref{fig:graph_schema}) are recomputed while all other learned structure is reused. Figure~\ref{fig:PM_replan_all} shows re-planning results across all three maze variants: the agent learns the environment once and then navigates to a sequence of goals without re-learning.

\begin{figure}[p]
    \centering
    \begin{subfigure}{0.65\textwidth}
        \centering
        \includegraphics[width=\textwidth]{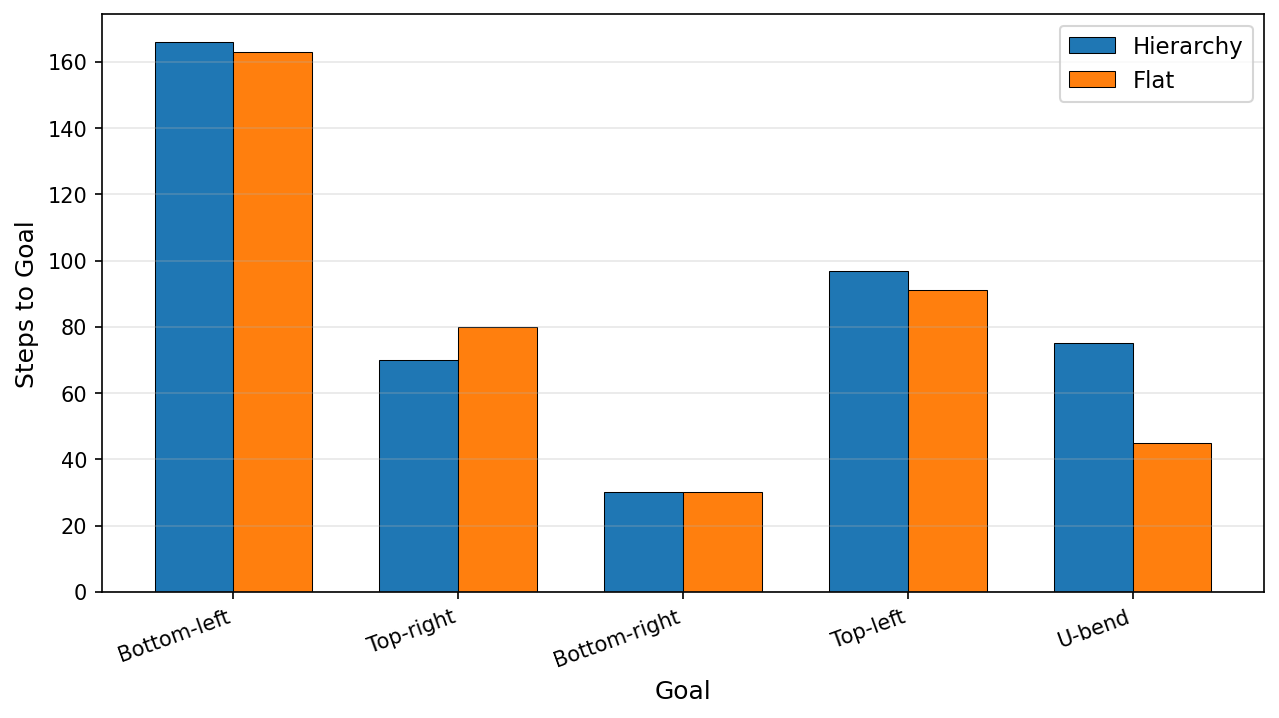}
        \caption{}
    \end{subfigure}
    \\[2pt]
    \begin{subfigure}{0.65\textwidth}
        \centering
        \includegraphics[width=\textwidth]{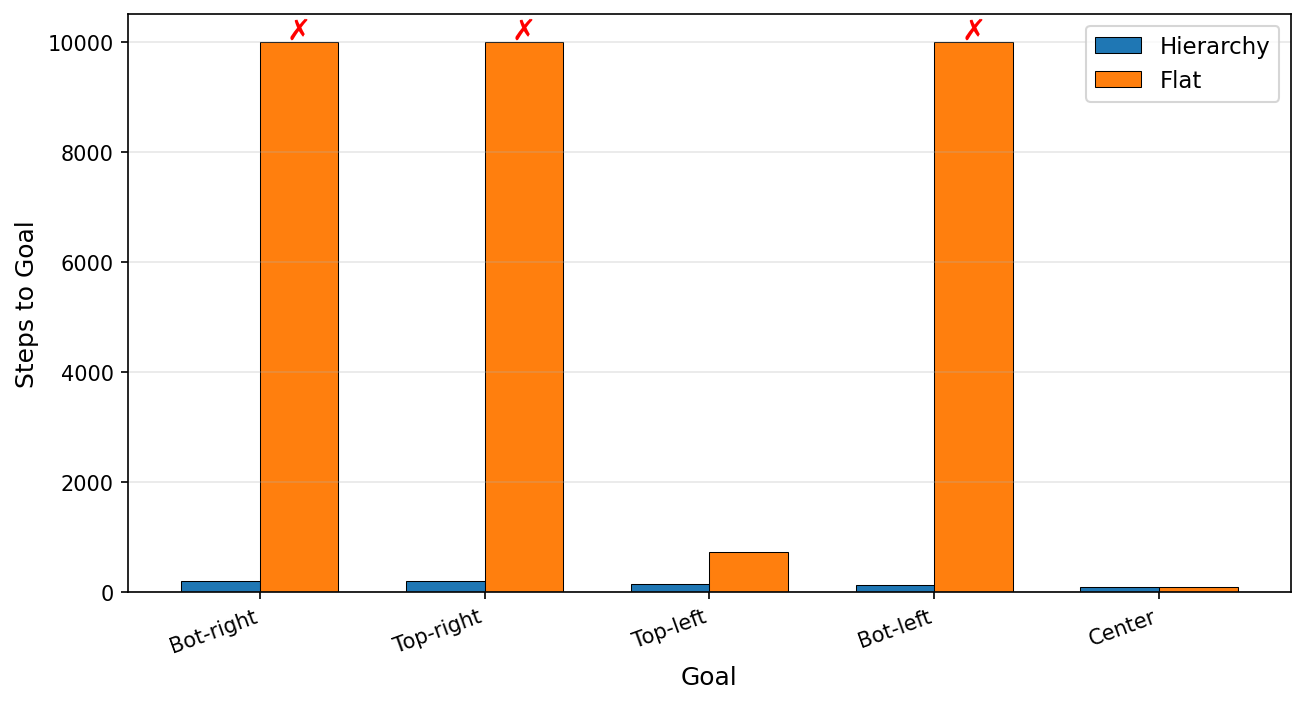}
        \caption{}
    \end{subfigure}
    \\[2pt]
    \begin{subfigure}{0.65\textwidth}
        \centering
        \includegraphics[width=\textwidth]{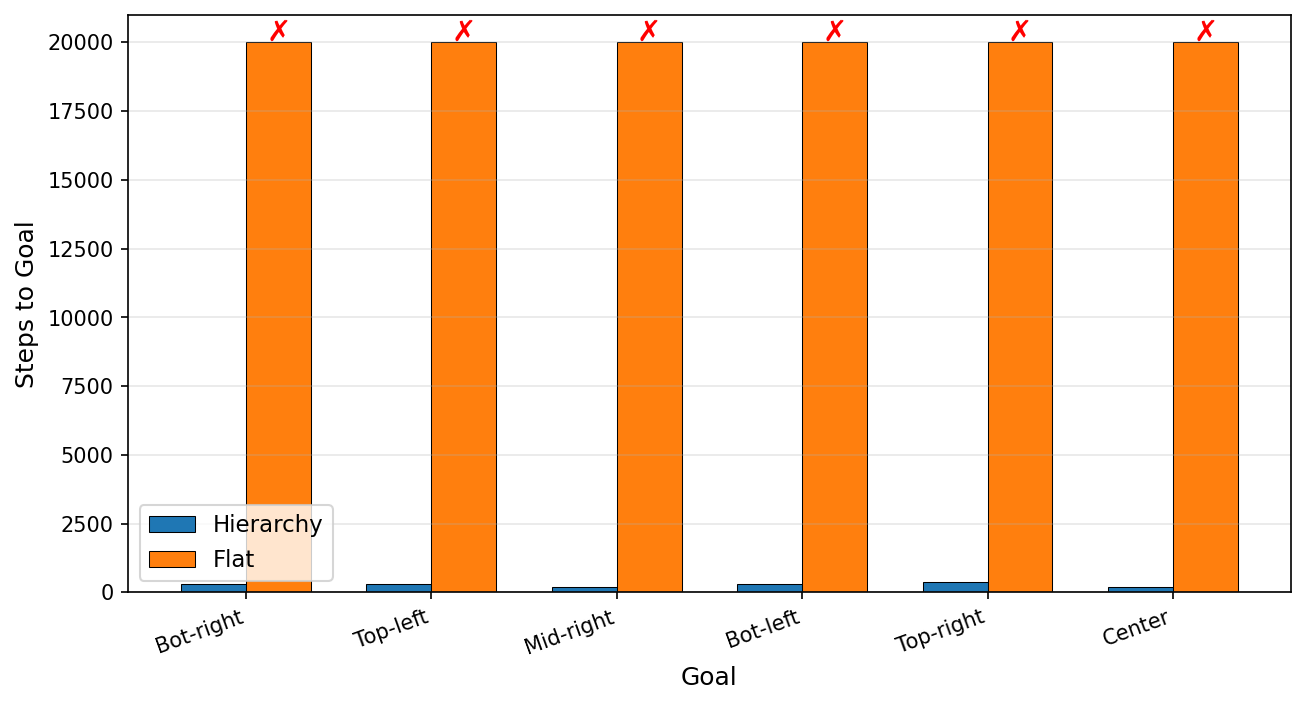}
        \caption{}
    \end{subfigure}
    \caption{{\bf PointMaze: Hierarchy vs.\ Flat Performance Comparison.} (a) UMaze. (b) Medium. (c) Large. Red crosses indicate failure to reach the goal. The hierarchical agent (blue) reaches all goals; the flat agent (orange) fails increasingly as the path to the goal gets longer.}
    \label{fig:PM_replan_steps}
\end{figure}

Figure~\ref{fig:PM_replan_steps} compares the performance, measured in terms of per-goal step counts to the goal, for hierarchical versus flat active inference. The hierarchical agent reaches all goals across maze variants (5/5 UMaze, 5/5 Medium, 6/6 Large) while the flat agent's performance degrades with maze complexity (5/5 UMaze, 2/5 Medium, 0/6 Large). On UMaze, both agents succeed because corridors are short enough for the learned successor representation to be informative. This is not the case for the Medium and Large maze variants: the flat agent fails on goals requiring long cross-room navigation.
Additional trajectory visualizations and clustering details are provided in Appendix~\ref{sec:pointmaze_details}.

\section{Conclusions and Future Work}
Active inference has emerged as a promising biologically-plausible approach to embodied intelligence \parencite{friston2017active, isomura2018vitro} that balances aspects of exploration and exploitation in a natural manner, and also brings action, learning, and perception together in a single unifying framework. However, implementations of active inference have been limited due to the difficulties in scaling traditional active inference methods to large-scale environments. 

In this paper, we explored combining active inference with the successor representations to achieve a hierarchical and tractable framework for active inference that is more amenable to scaling. Specifically, the proposed approach learns both hierarchical states and hierarchical actions, enabling more efficient planning in large-scale environments through hierarchical active inference. 

The hierarchical model was able to plan to multiple goals in navigation tasks using significantly less training as compared to a flat version of the model. Also, the model was able to learn a state hierarchy from the successor matrix even when the successor matrix was not complete enough for the agent to reach the goal using traditional low-level active inference. 

We also presented results from a task that required picking up a key at one location and using it to unlock the reward at a different goal location. These results illustrate how state and action hierarchies can be learned by the model for tasks more complex than simple navigation to a goal location.

Using the POMDP five-rooms task, we showed how the agent naturally optimizes the tradeoff between uncertainty due to observational noise and expected reward by virtue of minimizing expected free energy. Hierarchical planning by active inference allowed the agent to choose between a shorter path to the goal and a longer path in the face of greater uncertainty in the observations. Hierarchical active inference was previously applied to a different multi-rooms task \parencite{Neacsu-et-al2022}, with a focus on concept formation and learning where there is a limited number of rooms with a particular topology, rather than focusing on successor representations for efficient hierarchical planning as we do here.

Our Mountain Car experiments demonstrate that the framework extends to continuous state spaces, where the successor representation is learned over a discretized position-velocity state space and spectral clustering discovers dynamically meaningful macro states. The hierarchical agent solves the task with substantially fewer planning steps than the flat agent by reasoning over temporally extended macro actions.

The PointMaze experiments further highlight how the framework can be used for physics-based maze navigation tasks with continuous states and continuous actions across maze variants of increasing size (UMaze, Medium, Large). The hierarchical agent reliably reaches all goals across all variants while flat active inference degrades systematically with path length as the value gradient becomes too attenuated over long paths. The successor representation's reward-independence enables zero-cost re-planning to new goals by re-using learned structure about the environment. Together, the Mountain Car and PointMaze results demonstrate how hierarchical active inference, by operating at multiple spatial and temporal scales, enables efficient planning in continuous environments.

While our results are encouraging, the hierarchical model needs to be tested on more complex POMDP problems with large state and action spaces. For such scenarios, we intend to explore deep neural networks \parencite{ueltzhoffer2018deep}, focusing on their use for learning feature-based successor representations \parencite{barreto2017successor} and learning hierarchical states and actions. We also intend to test the efficacy of hypernetworks in generating hierarchical state and action networks, paralleling their use in active predictive coding networks \parencite{Rao-APC-2024,Rao-NN-2024}. 

Although we only used a two-level model in this work, the model could be generalized to using three or more levels in the state-action hierarchy. Adding more levels could potentially make planning to achieve goals even more efficient and accurate, opening the door to applying the model to larger-scale POMDP problems.

\section*{Acknowledgments}
This research was supported by the National Science Foundation (NSF) EFRI Grant no.\ 2223495, AFOSR award no.\ FA9550-24-1-0313, a Weill Neurohub Investigator grant, and a ``Frameworks'' grant from the Templeton World Charity Foundation. 

\section*{Appendix}
\setcounter{section}{0}
\setcounter{subsection}{0}
\setcounter{figure}{0}
\section{Gridworld}
\subsection{Four Rooms MDP Problem}
\label{sec:four_rooms}
\begin{figure}[!htbp]
\begin{minipage}[t]{\textwidth} % or '[b]', if desired
    \begin{subfigure}{0.45\textwidth}
        \centering
        \includegraphics[width=0.725\textwidth]{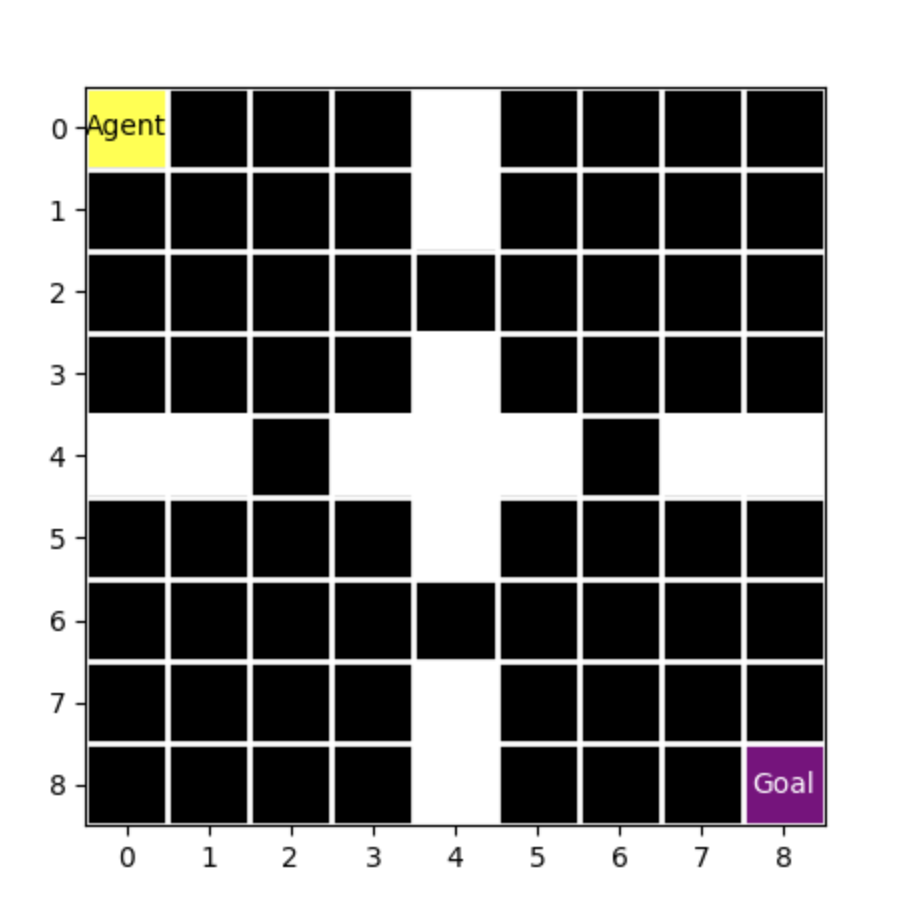} 
        \caption{}
        \label{fig:grid_four}
    \end{subfigure}
    \hfill
    \begin{subfigure}{0.45\textwidth}
        \includegraphics[width=\textwidth]{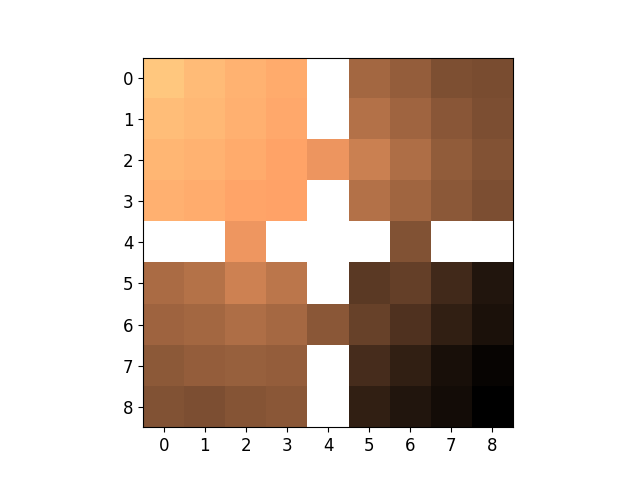} 
        \caption{}
        \label{fig:origin_four}
    \end{subfigure}
    \begin{subfigure}{0.45\textwidth}
        \includegraphics[width=\textwidth]{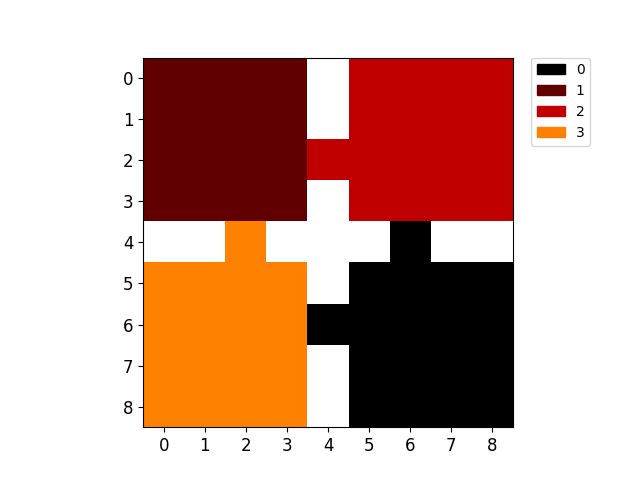}
        \caption{}
        \label{fig:macro_four}
    \end{subfigure}
    \hfill
    \begin{subfigure}{0.45\textwidth}
    \centering
    \includegraphics[width=\textwidth]{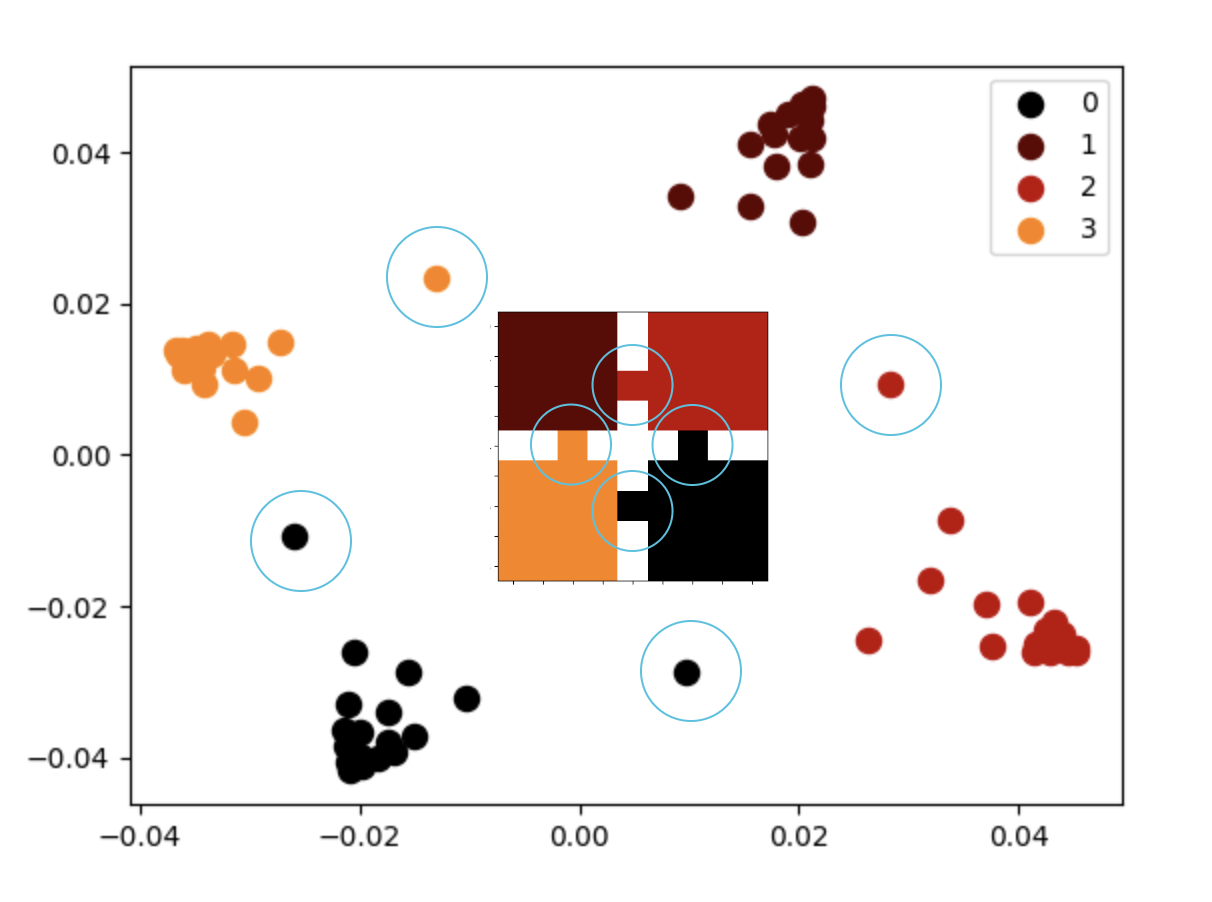}
    \caption{}
    \label{fig:cluster_four}
    \end{subfigure}
    \begin{subfigure}{0.45\textwidth}
    \centering
    \includegraphics[width=\textwidth]{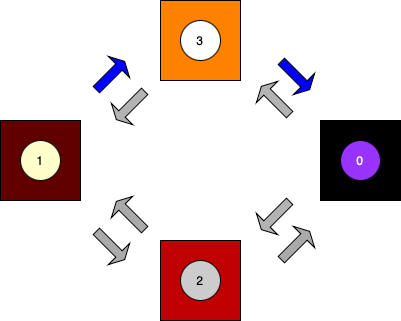}
    \caption{}
    \label{fig:macro_plan_four}
    \end{subfigure}
    \hfill
    \begin{subfigure}{0.45\textwidth}
        \includegraphics[width=\textwidth]{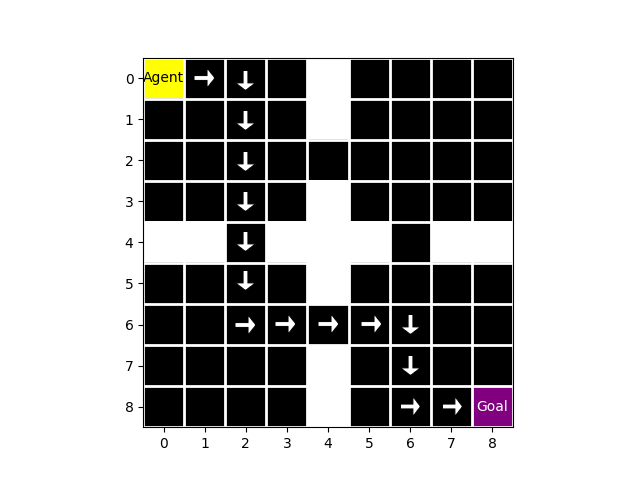}
        \caption{}
        \label{fig:path_four}
    \end{subfigure}
    \caption{{\bf Results for Four Rooms MDP Problem.} See text for details.}
\label{fig:mdp_four}
\end{minipage}
\end{figure}

In Section~\ref{sec:task1}, we examined a specific example of a Gridworld environment (``Serpentine'') with a unique optimal path to the goal. Additionally, due to the placement of walls, this path is highly winding, requiring a large number of steps to traverse from the start state to the goal state. However, our method applies to any wall configuration as long as a valid path exists. Take, for example, the classic Four Rooms problem (Figure~\ref{fig:mdp_four}). Figure~\ref{fig:grid_four} depicts the $9\times 9$ Four Rooms Gridworld containing an agent (yellow) and a single goal (purple), with the walls colored white. Figure~\ref{fig:origin_four} shows the learned successor representation of states, where states in closer proximity appear brighter due to their larger entries in the successor matrix. Figure~\ref{fig:macro_four} presents the macro states that are learned by our model by clustering based on the successor matrix. Since there are four rooms, we picked the same number of clusters, leading to the agent learning a separate cluster for each room. Figure~\ref{fig:cluster_four} visualizes these four macro state clusters by projecting the micro states into an embedding space. Note that in Figure~\ref{fig:cluster_four}, the corridors connecting rooms have been separately marked by blue circles. While these corridors have been arbitrarily assigned to one of the four room clusters, one can observe how these states occupy a distinct region in the embedding space, effectively functioning as bottleneck states. The agent’s hierarchical active inference involving planning at the macro level is shown in Figure~\ref{fig:macro_plan_four} while execution of the plan at the micro level is shown in Figure~\ref{fig:path_four}.

\subsection{POMDP Version of Task 1}
\label{sec:pomdp}
\begin{figure}[!htbp]
\begin{minipage}[t]{\textwidth}
    \begin{subfigure}{0.45\textwidth}
        \includegraphics[width=\textwidth]{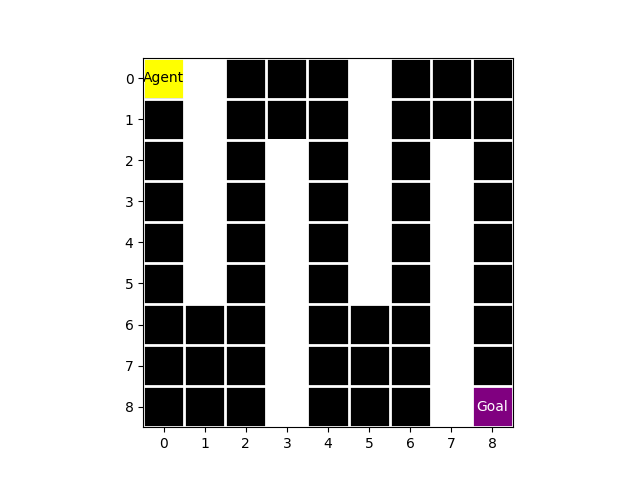} 
        \caption{}
        \label{fig:grid_noisy}
        \end{subfigure}
        \hfill
    \begin{subfigure}{0.44\textwidth}
        \includegraphics[width=\textwidth]{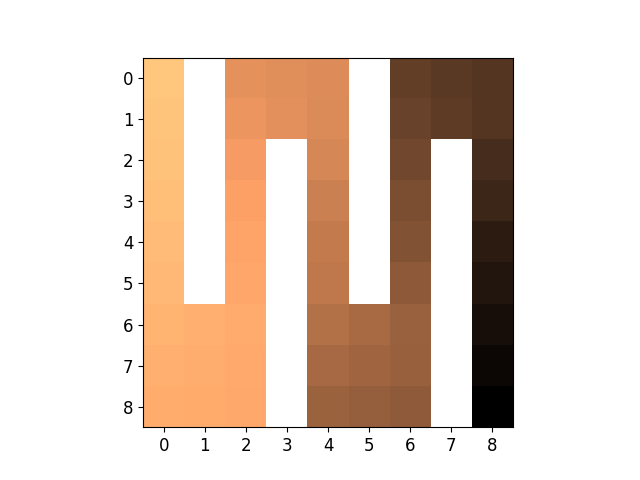} 
        \caption{}
        \label{fig:origin_noisy}
    \end{subfigure}
    \begin{subfigure}{0.45\textwidth}
        \includegraphics[width=\textwidth]{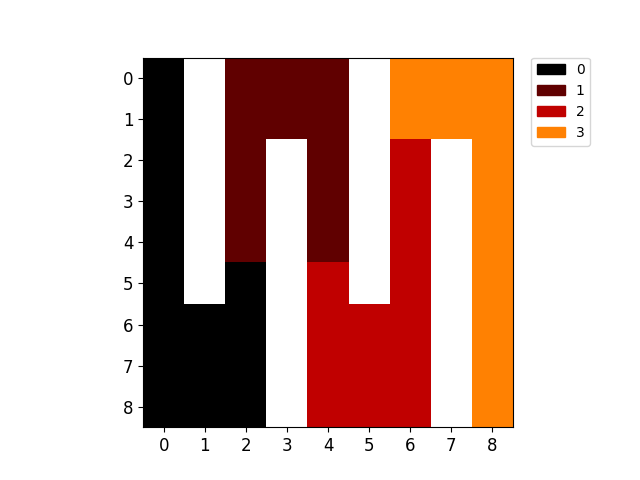}
        \caption{}
        \label{fig:macro_noisy}
    \end{subfigure}
    \hfill
    \begin{subfigure}{0.5\textwidth}
    \centering
    \includegraphics[width=\textwidth]{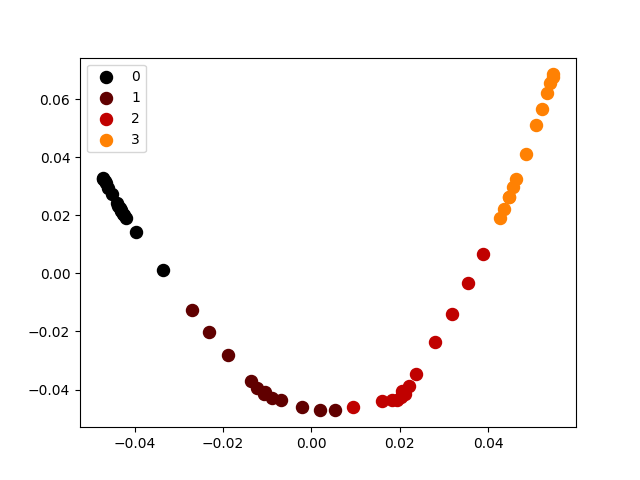}
    \caption{}
    \label{fig:cluster_noisy}
    \end{subfigure}
    \begin{subfigure}{0.45\textwidth}
        \includegraphics[width=\textwidth]{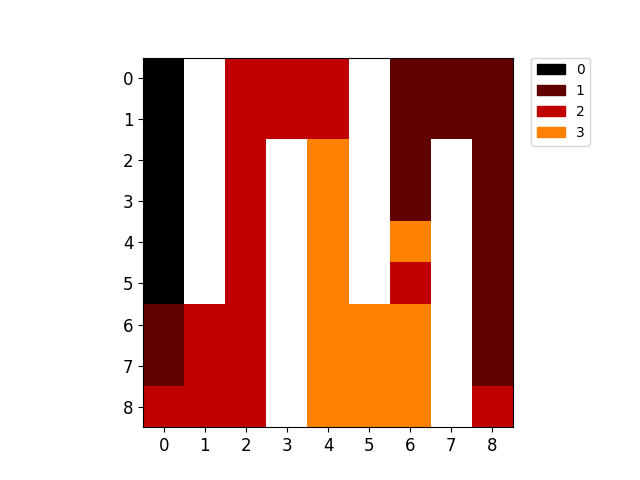}
        \caption{}
        \label{fig:macro_bad}
    \end{subfigure}
    \hfill
    \begin{subfigure}{0.5\textwidth}
    \centering
    \includegraphics[width=\textwidth]{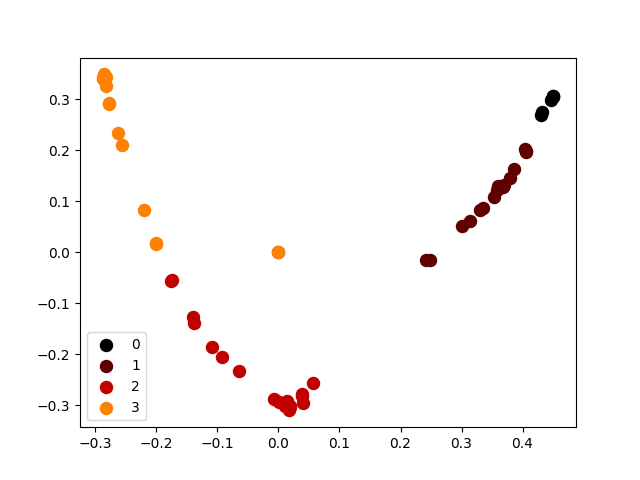}
    \caption{}
    \label{fig:cluster_bad}
    \end{subfigure}
    \caption{{\bf Results for POMDP Version of Task 1.} See text for details.}
\label{fig:pomdp}
\end{minipage}
\end{figure}

We tested the model on a POMDP version of the Gridworld in Task $1$, obtained by adding observational noise. The noise was localized, i.e., there is a small probability that the observation indicates that the agent is in one of the states in the local neighborhood of its true state. Figure~\ref{fig:grid_noisy} shows the $9\times 9$ Gridworld used in the task, with a single agent (yellow), a single goal (purple), walls (white) and noise parameter $\eta = 0.2$. As explained in the main text, this $\eta$ value means that $80\%$ of the time the agent gets an observation that matches its true hidden state (true location), but $20\%$ of the time the observation is a nearby state. Figure~\ref{fig:origin_noisy} shows the successor representation of states with respect to the initial state: states that are closer to the initial state show up brighter in the figure due to their larger entries in the successor matrix. Figure~\ref{fig:macro_noisy} shows the macro states that are learned by the model ($4$ in number). Figure~\ref{fig:cluster_noisy} shows a visualization of the $4$ macro state clusters in terms of projections of the micro states into an embedding space. These results show that for a low amount of noise, the method results in good clusters similar to the MDP case. If we significantly increase the noise to $\eta = 0.8$, the clustering process does get affected and the clusters are not contiguous anymore (Figures~\ref{fig:macro_bad} and~\ref{fig:cluster_bad}). In this case, we found that the agent struggles to learn good macro actions, either getting stuck near a wall or entering into loops at the boundaries between the clusters.

\section{Smooth Successor Matrices for Clustering}
\label{sec:smooth_M}
In the case of the experiments with continuous state spaces, often the exploration is not sufficient enough to get smooth clustering without outliers or discontinuities, which can affect the generation of macro actions and planning in the higher level. To address this issue, we use a spatial RBF kernel as a form of regularizer on the matrix described as follows.

\subsection{Spatial RBF Kernel}

We compute pairwise squared Euclidean distances between states in normalised Cartesian coordinates:
\begin{equation}
D^2_{(i,j),(k,l)} = \left\|\mathbf{x}_{i,j} - \mathbf{x}_{k,l}\right\|^2
\end{equation}
and form a radial basis function (Gaussian) kernel:
\begin{equation}
K^{\text{spatial}}_{(i,j),(k,l)} = \exp\!\left(-\frac{D^2_{(i,j),(k,l)}}{2\sigma^2}\right)
\end{equation}
with bandwidth $\sigma = 1.0$. This kernel is symmetric, non-negative,
and equals~1 on the diagonal.

\subsection{Successor Representation Based Affinity}

The standard spectral clustering affinity is the symmetrised successor
matrix:
\begin{equation}
\mathbf{A}^{\text{SR}} = \max\!\left(M,\;M^\top\right)
\end{equation}
To blend it with the spatial kernel, we first normalise
$\mathbf{A}^{\text{SR}}$ to $[0, 1]$:
\begin{equation}
\widetilde{\mathbf{A}}^{\text{SR}} = \frac{\mathbf{A}^{\text{SR}} - \min\!\left(\mathbf{A}^{\text{SR}}\right)}{\max\!\left(\mathbf{A}^{\text{SR}}\right) - \min\!\left(\mathbf{A}^{\text{SR}}\right)}
\end{equation}

\subsection{Adaptive Blending}

A uniform blend weight $\alpha$ applies the spatial kernel equally to all
state pairs, regardless of how well $M$ has been estimated.
This can be problematic: in environments with walls or barriers, two
states may be spatially adjacent but dynamically disconnected. A uniform
spatial kernel would incorrectly boost their affinity, blurring the very
bottleneck structure that clustering should discover.

We address this with an \emph{adaptive} blend weight that varies per
state pair based on the confidence in $M$. The intuition is
simple: trust $M$ where it is reliable; fall back on the spatial
kernel where it is not.

\paragraph{Per-state confidence.}
The row sum $r_i = \sum_j M_{ij}$ reflects how well state~$i$ has been
explored: well-visited states accumulate large successor mass, while
rarely-visited states (e.g., near-goal boundary regions) have small row
sums. We normalise to $[0,1]$:
\begin{equation}
\hat{c}_i = \frac{r_i}{\max_k\, r_k}
\end{equation}

\paragraph{Per-pair confidence.}
For the pair $(i,j)$, we take the minimum of the two states'
confidences, so that the spatial kernel activates whenever \emph{either}
state is poorly estimated:
\begin{equation}
c_{ij} = \min\!\left(\hat{c}_i,\;\hat{c}_j\right)
\end{equation}

\paragraph{Adaptive blend weight.}
The effective spatial weight for pair $(i,j)$ is:
\begin{equation}
\alpha_{ij} = \alpha_{\max}\,(1 - c_{ij})
\end{equation}
where $\alpha_{\max}$ is the maximum blend weight (a hyperparameter;
default $0.3$ for Mountain Car, $0.15$ for PointMaze). The final
blended affinity becomes:
\begin{equation}
\boxed{A_{ij} = (1 - \alpha_{ij})\,\widetilde{A}^{\text{SR}}_{ij} + \alpha_{ij}\,K^{\text{spatial}}_{ij}}
\end{equation}

This gives the desired behaviour in three regimes:

\begin{itemize}
\item \textbf{Both states well-visited}
($\hat{c}_i, \hat{c}_j \approx 1$): $\alpha_{ij} \approx 0$.
The affinity is determined almost entirely by the SR, dynamical barriers
(walls, unreachable regions) are preserved.
\item \textbf{At least one state poorly-visited}
($\min(\hat{c}_i, \hat{c}_j) \approx 0$):
$\alpha_{ij} \approx \alpha_{\max}$. The spatial kernel fills in for the
unreliable SR, preventing sparse boundary states from splitting off as
outlier clusters.

\end{itemize}

\subsection{Scope of Application}

The blended affinity $\mathbf{A}$ is used exclusively for \emph{macro state
discovery}: it determines the spectral embedding for visualisation and
the spectral clustering for cluster assignments. The core micro-level
quantities, the learned successor matrix $M$, the transition
matrix $B$, and the value function
$V = M\,C$ remain unmodified.

Macro-level planning matrices ($B_{\text{macro}}$,
$M_{\text{macro}}$) are derived from raw transition counts
between the resulting clusters, so they are \emph{indirectly} shaped by
the blended clustering but do not incorporate the spatial kernel
themselves.

This separation is deliberate: the spatial kernel should produce
physically coherent macro state regions without distorting the learned
dynamics that drive action selection.

% Prior work \parencite{stachenfeld2017hippocampus} has showed that the
% eigenvectors of the successor representation matrix $\mathbf{M}$ are
% equivalent to proto-value functions, the eigenvectors of the graph
% Laplacian over the state-transition graph.  Building on this equivalence,
% Machado et~al.~\autocite{machado2018eigenoption} proposed an
% \emph{eigenoption discovery} framework that uses spectral decomposition
% of $\mathbf{M}$ to identify bottleneck states and construct temporally
% abstract actions (options) from raw pixels.  The earlier Laplacian
% framework~\autocite{machado2017laplacian} showed that eigenoptions
% correspond to eigenvectors associated with the largest eigenvalues of
% $\mathbf{M}$ (equivalently, the smallest non-trivial eigenvalues of the
% graph Laplacian), providing a principled connection between spectral
% methods and option discovery in reinforcement learning.  Our micro-level
% spectral clustering on $\mathbf{M}$ follows this established approach
% directly.

The idea of combining multiple affinity matrices appears extensively in
multi-view clustering \parencite{kumar2011co,chao2021survey}, where a unified affinity is learned from several
``views'' of the same data. Our convex combination of a dynamics-based
affinity ($\widetilde{\mathbf{A}}^{\text{SR}}$) with a geometry-based
kernel ($\mathbf{K}^{\text{spatial}}$) is conceptually analogous,
treating the successor representation and the physical state layout as
two complementary views of the same state space.

The adaptive confidence weighting makes the blending robust to environments 
containing dynamical barriers, where a uniform spatial kernel would incorrectly
smooth over true bottleneck structure.
To our knowledge, the specific combination of a learned successor
representation with a spatial RBF kernel via an adaptive,
confidence-weighted convex blend for macro state discovery has not
appeared in prior work. 

\section{PointMaze: Supplementary Details}
\label{sec:pointmaze_details}

\subsection{State and Action Discretization}

The continuous 2D position $(x, y)$ is discretized into a grid of $n_x \times n_y$ bins. For a maze with $c$ columns and $r$ rows, the physical coordinate ranges are $x \in [-c/2, c/2]$ and $y \in [-r/2, r/2]$ (each maze cell is 1 unit). A continuous position is mapped to bin indices $(i, j)$, and the flat state index is $s = i \cdot n_y + j$, giving $N = n_x \cdot n_y$ total states. Bins whose centers fall inside wall cells are excluded from learning and planning.

Velocity is discarded in the state representation. Unlike MountainCar where velocity is useful for distinguishing dynamically different states, the PointMaze point mass has negligible momentum at the discretization scale. The SR captures temporal dynamics implicitly: states from which the agent tends to move eastward will have high SR entries with eastern states, even without explicit velocity.

The continuous 2D force space is mapped to 8 discrete directions: the four cardinal (E, W, N, S) and four diagonal directions, each with unit or near-unit magnitude.

\paragraph{Smooth stepping.}
Because the point mass moves only about 0.0024 units per physics step, but each bin spans 0.25 units (for UMaze with 20 bins over 5 units), a single physics step rarely changes the discrete state. Each discrete action is repeated for up to $N_{\text{smooth}}$ consecutive physics steps, terminating early when the discrete state changes, the goal region is entered, or the environment signals termination. Training uses $N_{\text{smooth}}^{\text{train}} = 200$ to allow exploration, while testing uses $N_{\text{smooth}}^{\text{test}} = 100$ for finer control.

The SR is learned from diverse-start exploration (i.e., each episode begins from a uniformly random navigable position).
%with experience replay at half the learning rate for stabilization.
As in MountainCar, the affinity matrix for spectral clustering uses a light spatial RBF kernel blend (see Appendix~\ref{sec:smooth_M}) with $\alpha_{\text{blend}} = 0.15$ deliberately lower than for MountainCar, because the SR in a maze already captures wall-induced disconnections well.

\subsection{Hyperparameters}
Table~\ref{tab:PM_hyperparams} is a collection of all the hyperparameters used for the various PointMaze environments. The shared parameters across all $3$ variants are:
Learning rate $\alpha = 0.05$, discount factor $\gamma = 0.95$, goal reward $r_{\text{goal}} = 100$, default cost $c_{\text{default}} = -1$.
\begin{table}[!htbp]
\centering
\caption{Hyperparameter configuration for each PointMaze variant.} 
\label{tab:PM_hyperparams}
\begin{tabular}{lccc}
\toprule
\textbf{Parameter} & \textbf{UMaze} & \textbf{Medium} & \textbf{Large} \\
\midrule
Discretization bins ($n_x \times n_y$) & $20 \times 20$ & $32 \times 32$ & $48 \times 36$ \\
Total states $N$ & 400 & 1024 & 1728 \\
Navigable states $N_v$ & $\sim$220 & $\sim$416 & $\sim$736 \\
Number of clusters $k$ & 4 & 8 & 12 \\
Training episodes & 1500 & 5000 & 25000 \\
Max test steps & 5000 & 10000 & 20000 \\
\bottomrule
\end{tabular}
\end{table}

\subsection{Single-Goal Planning: Medium and Large}
\label{sec:PM_single_goal_extended}

\begin{figure}[!t]
    \centering
    \begin{subfigure}{0.48\textwidth}
        \centering
        \includegraphics[width=\textwidth]{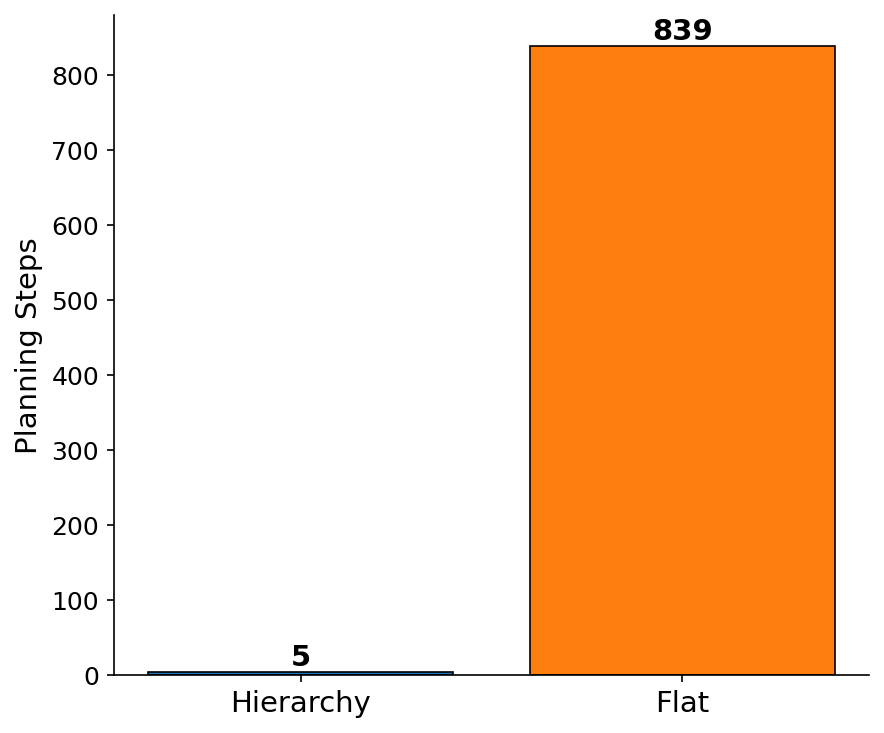}
        \caption{}
    \end{subfigure}
    \hfill
    \begin{subfigure}{0.48\textwidth}
        \centering
        \includegraphics[width=\textwidth]{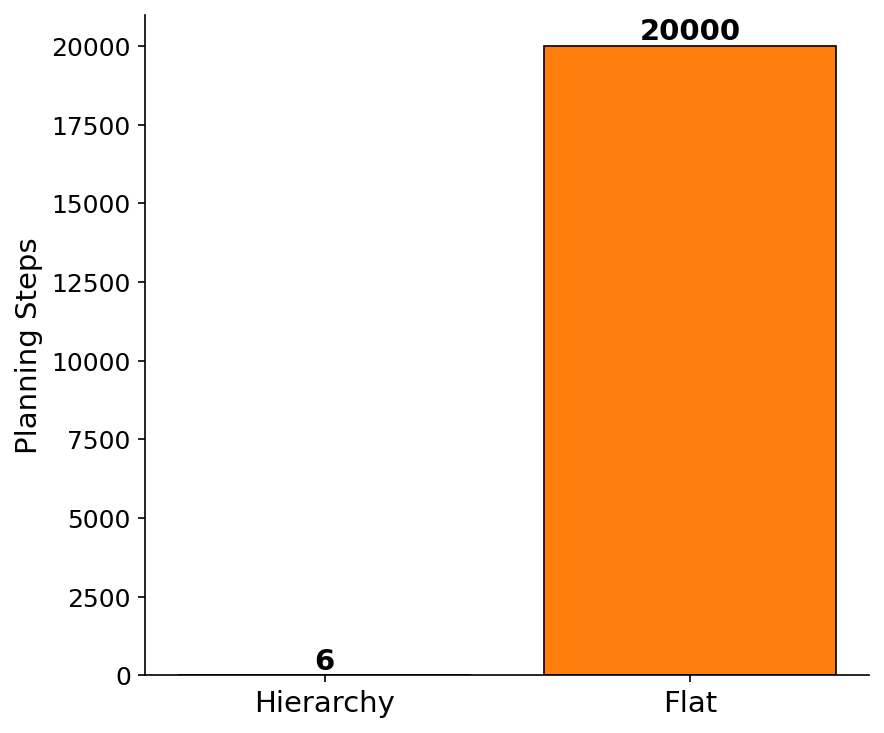}
        \caption{}
    \end{subfigure}
    \caption{{\bf Single-goal planning cost: Medium and Large.} (a) Medium: both agents succeed, but the flat agent requires $170\times$ more planning steps. (b) Large: the flat agent times out at 20{,}000 steps (the maximum) without reaching the goal, while the hierarchy succeeds in 6 steps.}
    \label{fig:PM_planning_extended}
\end{figure}

Figure~\ref{fig:PM_planning_extended} extends the single-goal planning comparison from UMaze above to the Medium and Large variants. On Medium, the hierarchical agent reaches the goal in 5 planning decisions while the flat agent requires 839 steps ($\sim$168$\times$ more); both succeed, but the efficiency gap widens substantially compared to UMaze. On Large, the hierarchical agent succeeds in just 6 planning decisions, while the flat agent times out at the 20{,}000-step limit without reaching the goal.

The root cause is vanishing value gradients in the flat representation $V = MC$. With $\gamma = 0.95$, the goal reward is attenuated by $\gamma^N$ over $N$ bin transitions. On UMaze, corridors span $\sim$12 bins per room ($\gamma^{12} \approx 0.54$), leaving value differences large enough for the greedy policy. On Medium, paths span about 100 bins ($\gamma^{100} \approx 0.006$), compressing start-region values to near-uniform. On Large, with 150+ bins ($\gamma^{150} \approx 0.0005$), the flat value function is effectively constant, making gradient-based planning infeasible. The hierarchy avoids this through multi-scale value computation: the macro level $V_{\text{macro}} = M_{\text{macro}} \cdot C_{\text{macro}}$ over $k$ macro states provides a clear room-to-room gradient, temporary bottleneck value functions create strong local gradients for each room transition, and the global value function is only needed within the goal cluster where values are large.

\subsection{Trajectory and Cluster Visualizations}
\begin{figure}[!b]
    \centering
    \includegraphics[width=0.76\textwidth]{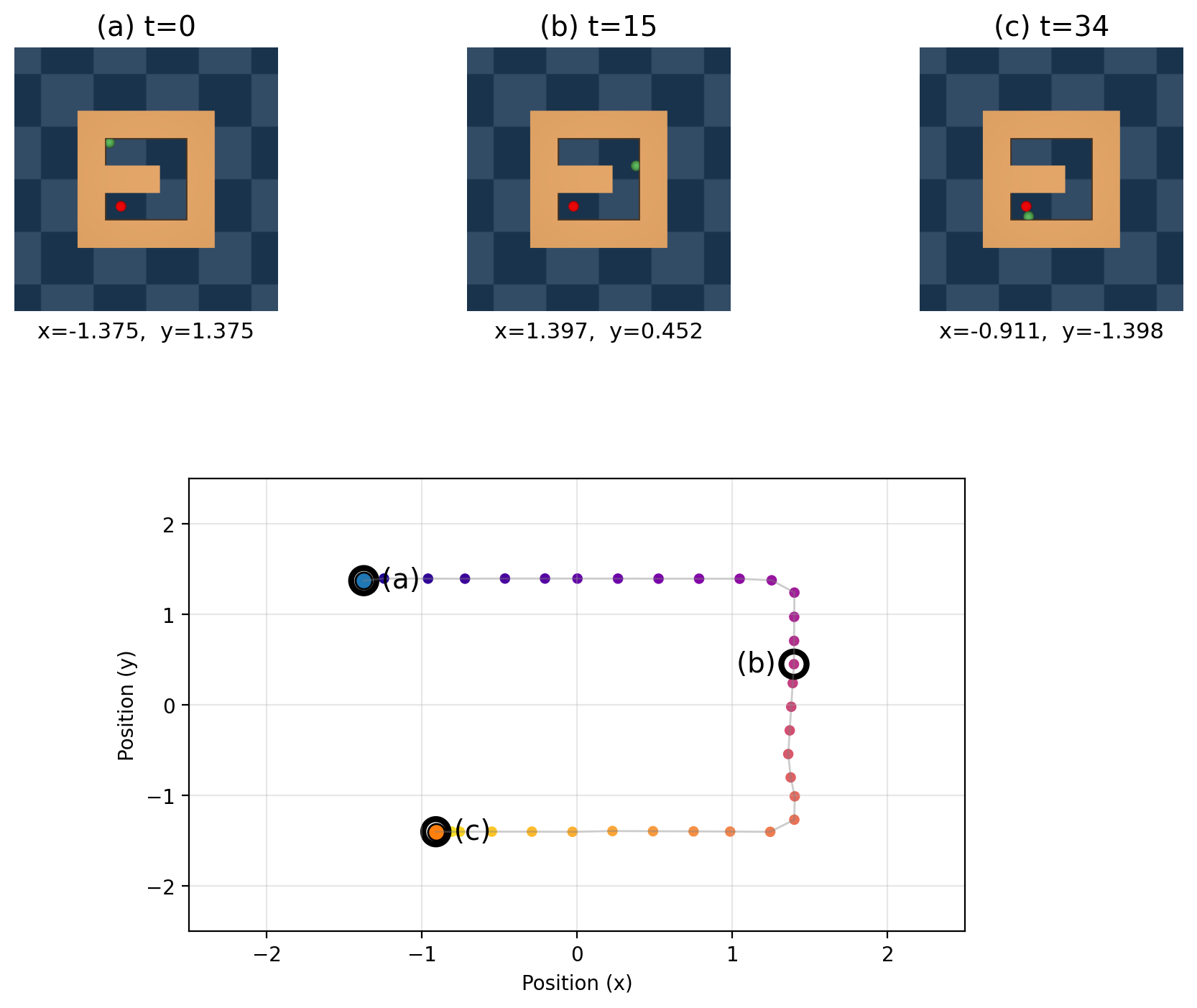}
    \caption{{\bf Trajectory for UMaze.} Rendered frames (top) and corresponding trajectory segments (bottom). (a) to (b) to (c) correspond to macro actions (room-to-room navigation).}
    \label{fig:PM_stages}
\end{figure}

\begin{figure}[!t]
    \centering
    \begin{subfigure}[t]{0.45\textwidth}
        \centering
        \includegraphics[width=\textwidth]{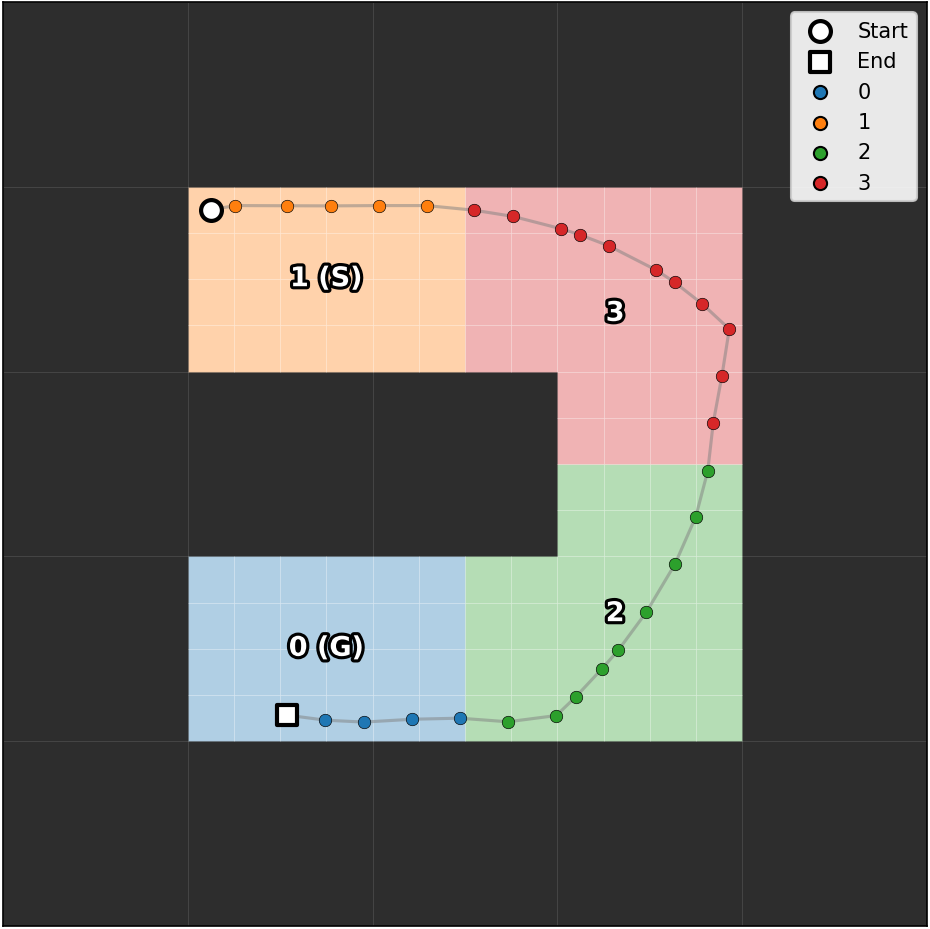}
        \caption{}
        \label{fig:PM_traj_state}
    \end{subfigure}
    \hfill
    \begin{subfigure}[t]{0.45\textwidth}
        \centering
        \includegraphics[width=\textwidth]{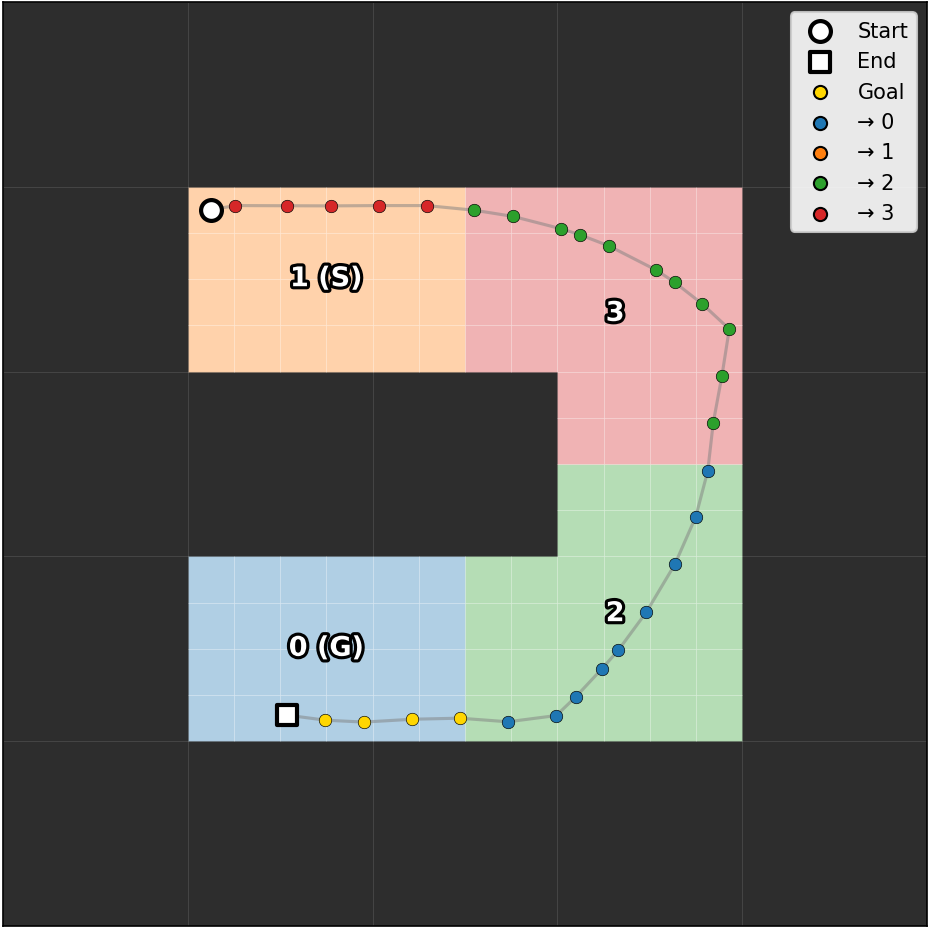}
        \caption{}
        \label{fig:PM_traj_action}
    \end{subfigure}
    \caption{{\bf Hierarchical trajectory on UMaze.} The agent navigates from start (top-left) to goal (bottom-left), overlaid on the cluster map (faded regions). (a) Trajectory colored by current macro state membership. (b) Trajectory colored by macro action target (the cluster being navigated toward); gold indicates the goal phase within the final cluster.}
    \label{fig:PM_trajectory}
\end{figure}

\begin{figure}[!b]
    \centering
    \begin{subfigure}{0.48\textwidth}
        \centering
        \includegraphics[width=\textwidth]{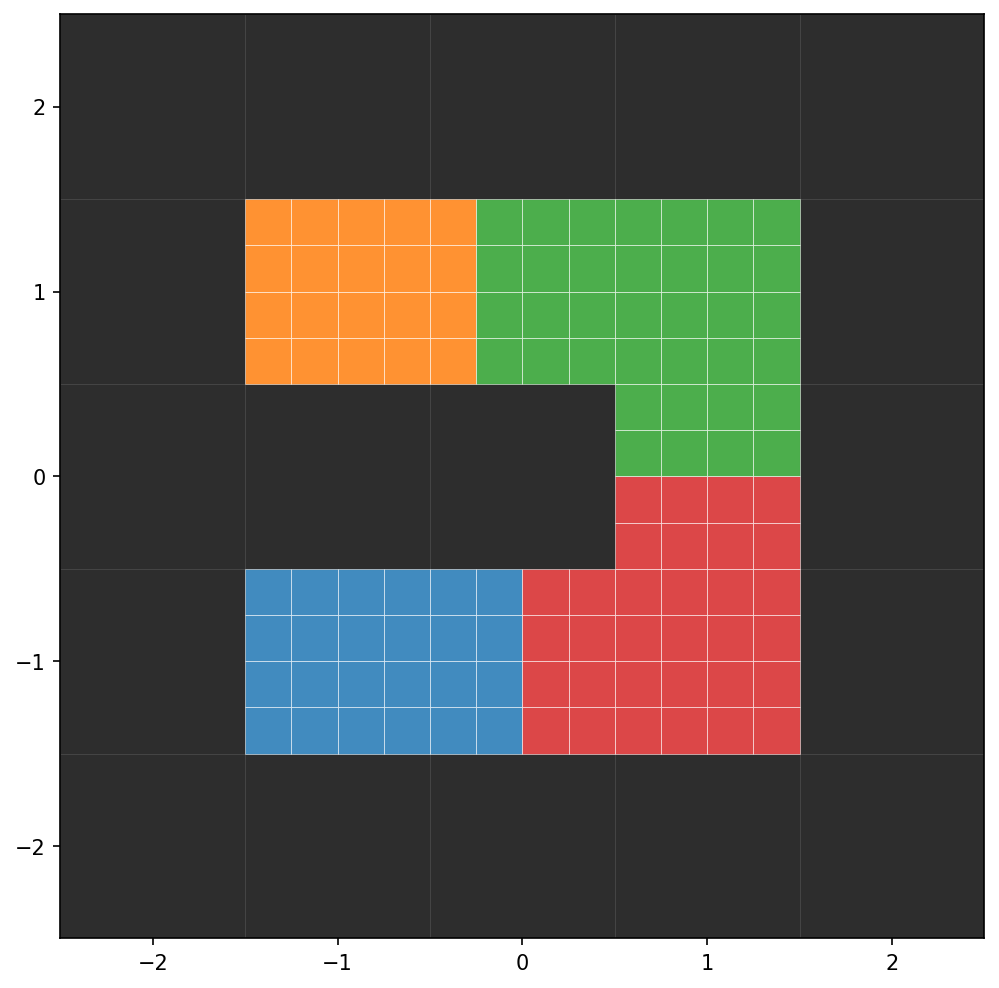}
        \caption{}
    \end{subfigure}
    \hfill
    \begin{subfigure}{0.48\textwidth}
        \centering
        \includegraphics[width=\textwidth]{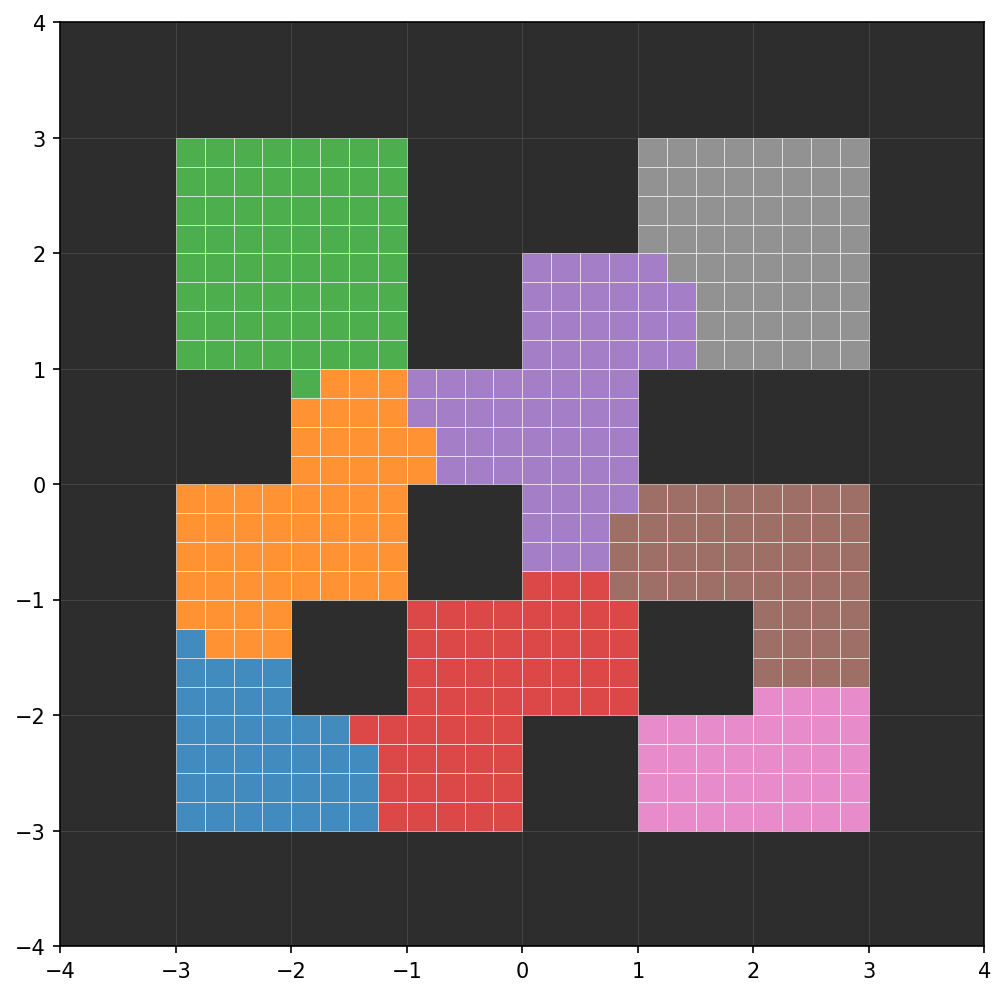}
        \caption{}
    \end{subfigure}

    \vspace{0.5em}

    \begin{subfigure}{0.60\textwidth}
        \centering
        \includegraphics[width=\textwidth]{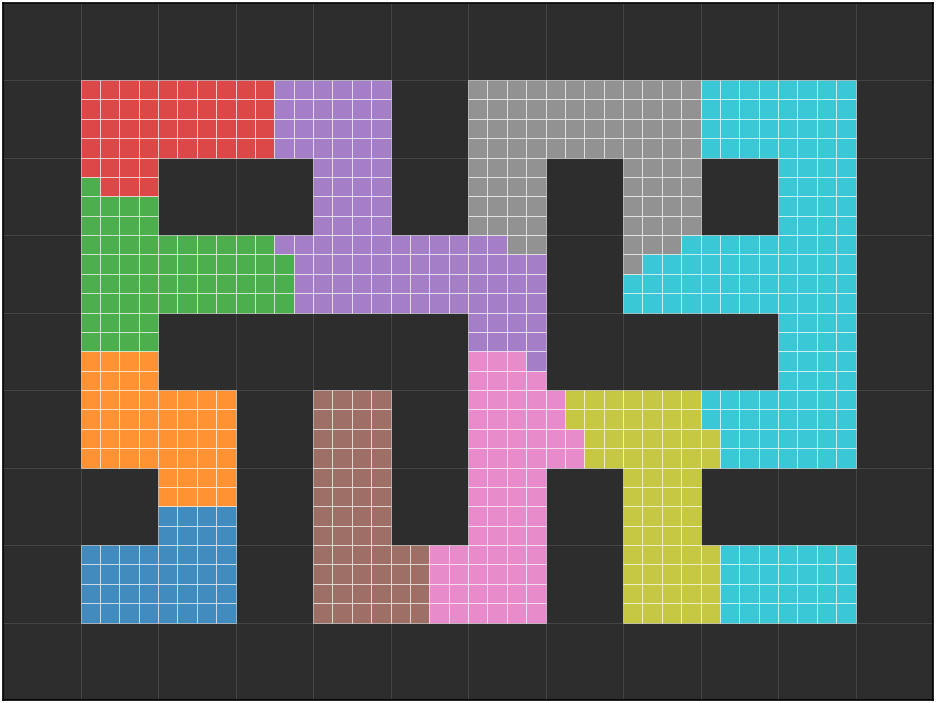}
        \caption{}
    \end{subfigure}
    \caption{{\bf Macro State Clusters for the Three Maze Variants.} (a) UMaze (4 clusters). (b) Medium (8 clusters). (c) Large (12 clusters). Colored regions show clusters discovered by spectral clustering of the successor representation. Cluster boundaries align with narrow corridors connecting rooms.}
    \label{fig:PM_replan_clusters}
\end{figure}

Here we show a few extra visualizations to help clarify how the agent navigates and clusters the PointMaze environment.
Figure~\ref{fig:PM_stages} shows the state trajectory alongside rendered frames.
Figure~\ref{fig:PM_trajectory} shows the hierarchical trajectory on UMaze, colored by macro state and macro action. Figure~\ref{fig:PM_replan_clusters} shows the macro state clusters for all three maze variants.

\clearpage
\printbibliography

@article{Rao-Ballard1999,
  title={Predictive coding in the visual cortex: a functional interpretation of some extra-classical receptive-field effects},
  author={Rao, Rajesh P. N. and Ballard, Dana H},
  journal={Nature neuroscience},
  volume={2},
  number={1},
  pages={79--87},
  year={1999},
  publisher={Nature Publishing Group}
}

@article{Pateria2021,
author = {Pateria, Shubham and Subagdja, Budhitama and Tan, Ah-hwee and Quek, Chai},
title = {Hierarchical Reinforcement Learning: A Comprehensive Survey},
year = {2021},
issue_date = {June 2022},
publisher = {Association for Computing Machinery},
address = {New York, NY, USA},
volume = {54},
number = {5},
issn = {0360-0300},
url = {https://doi.org/10.1145/3453160},
doi = {10.1145/3453160},
journal = {ACM Comput. Surv.},
month = jun,
articleno = {109},
numpages = {35}
}

@inproceedings{Dayan1992,
author = {Dayan, Peter and Hinton, Geoffrey E.},
title = {Feudal reinforcement learning},
year = {1992},
isbn = {1558602747},
publisher = {Morgan Kaufmann Publishers Inc.},
address = {San Francisco, CA, USA},
booktitle = {Proceedings of the 6th International Conference on Neural Information Processing Systems},
pages = {271--278},
numpages = {8},
location = {Denver, Colorado},
series = {NIPS'92}
}

@incollection{Kaufman1990,
  title = {Partitioning {Around} {Medoids} ({Program} {PAM})},
  booktitle = {Finding {Groups} in {Data}},
  publisher = {John Wiley \& Sons, Inc.},
  author = {Kaufman, Leonard and Rousseeuw, Peter J.},
  year = {1990},
  doi = {10.1002/9780470316801.ch2},
  pages = {68--125},
}

@article{Murtagh2012,
author = {Murtagh, Fionn and Contreras, Pedro},
title = {Algorithms for hierarchical clustering: an overview},
journal = {WIREs Data Mining and Knowledge Discovery},
volume = {2},
number = {1},
pages = {86-97},
doi = {10.1002/widm.53},
url = {https://wires.onlinelibrary.wiley.com/doi/abs/10.1002/widm.53},
eprint = {https://wires.onlinelibrary.wiley.com/doi/pdf/10.1002/widm.53},
year = {2012}
}

@inproceedings{ng2002spectral,
  title={On spectral clustering: Analysis and an algorithm},
  author={Ng, Andrew Y and Jordan, Michael I and Weiss, Yair},
  booktitle={Advances in neural information processing systems},
  pages={849--856},
  year={2002}
}

@book{Rao_Olshausen_ProbabilisticModels_2002,
  title={Probabilistic models of the brain: Perception and neural function},
  editor={Rao, Rajesh P. N. and Olshausen, Bruno A and Lewicki, Michael S},
  year={2002},
  publisher={MIT Press}
}

@book{Doya_etal_BayesianBrain_2011,
  title={Bayesian Brain: Probabilistic Approaches to Neural Coding},
  editor={Doya, Kenji and Ishii, Shin and Pouget, Alexandre and Rao, Rajesh P. N.},
  year={2007},
  publisher={MIT Press},
  address={Cambridge, MA}
}

@article{fountas2020deep,
  title={Deep active inference agents using {Monte-Carlo} methods},
  author={Fountas, Zafeirios and Sajid, Noor and Mediano, Pedro A. M. and Friston, Karl},
  journal={Advances in neural information processing systems},
  volume={33},
  pages={11662--11675},
  year={2020}
}

@article{ueltzhoffer2018deep,
  title={Deep active inference},
  author={Ueltzh{\"o}ffer, Kai},
  journal={Biological cybernetics},
  volume={112},
  number={6},
  pages={547--573},
  year={2018},
  publisher={Springer}
}

@article{stachenfeld2017hippocampus,
  title={The hippocampus as a predictive map},
  author={Stachenfeld, Kimberly L and Botvinick, Matthew M and Gershman, Samuel J},
  journal={Nature neuroscience},
  volume={20},
  number={11},
  pages={1643--1653},
  year={2017},
  publisher={Nature Publishing Group US New York}
}

@article{dayan1993improving,
  title={Improving generalization for temporal difference learning: The successor representation},
  author={Dayan, Peter},
  journal={Neural computation},
  volume={5},
  number={4},
  pages={613--624},
  year={1993},
  publisher={MIT Press}
}

@article{friston2010free,
  title={The free-energy principle: a unified brain theory?},
  author={Friston, Karl},
  journal={Nature reviews neuroscience},
  volume={11},
  number={2},
  pages={127--138},
  year={2010},
  publisher={Nature Publishing Group UK London}
}

@article{aitchison2017or,
  title={With or without you: predictive coding and {B}ayesian inference in the brain},
  author={Aitchison, Laurence and Lengyel, M{\'a}t{\'e}},
  journal={Current opinion in neurobiology},
  volume={46},
  pages={219--227},
  year={2017},
  publisher={Elsevier}
}

@inproceedings{millidge2023active,
  title={Active Inference Successor Representations},
  author={Millidge, Beren and Buckley, Christopher L},
  booktitle={Active Inference: Third International Workshop, IWAI 2022, Grenoble, France, September 19, 2022, Revised Selected Papers},
  pages={151--161},
  year={2023},
  organization={Springer}
}

@article{da2020active,
  title={Active inference on discrete state-spaces: A synthesis},
  author={Da Costa, Lancelot and Parr, Thomas and Sajid, Noor and Veselic, Sebastijan and Neacsu, Victorita and Friston, Karl},
  journal={Journal of Mathematical Psychology},
  volume={99},
  pages={102447},
  year={2020},
  publisher={Elsevier}
}

@article{friston2018deep,
  title={Deep temporal models and active inference},
  author={Friston, Karl J. and Rosch, Richard and Parr, Thomas and Price, Cathy and Bowman, Howard},
  journal={Neuroscience \& Biobehavioral Reviews},
  volume={90},
  pages={486--501},
  year={2018},
  publisher={Elsevier}
}

@article{friston2015active,
  title={Active inference and epistemic value},
  author={Friston, Karl and Rigoli, Francesco and Ognibene, Dimitri and Mathys, Christoph and Fitzgerald, Thomas and Pezzulo, Giovanni},
  journal={Cognitive neuroscience},
  volume={6},
  number={4},
  pages={187--214},
  year={2015},
  publisher={Taylor \& Francis}
}

@article{sharot2011optimism,
  title={The optimism bias},
  author={Sharot, Tali},
  journal={Current biology},
  volume={21},
  number={23},
  pages={R941--R945},
  year={2011},
  publisher={Elsevier}
}

@article{friston2006free,
  title={A free energy principle for the brain},
  author={Friston, Karl and Kilner, James and Harrison, Lee},
  journal={Journal of physiology-Paris},
  volume={100},
  number={1-3},
  pages={70--87},
  year={2006},
  publisher={Elsevier}
}

@article{huang2011predictive,
  title={Predictive coding},
  author={Huang, Yanping and Rao, Rajesh P. N.},
  journal={Wiley Interdisciplinary Reviews: Cognitive Science},
  volume={2},
  number={5},
  pages={580--593},
  year={2011},
  publisher={Wiley Online Library}
}

@article{Rao-APC-2024,
	title = {Active Predictive Coding: A Unifying Neural Model for Active Perception, Compositional Learning, and Hierarchical Planning},
 	journal = {Neural Computation},
	publisher = {MIT Press},
	author = {Rao, Rajesh P. N. and Gklezakos, Dimitrios C. and Sathish, Vishwas},
        volume = {36},
        number = {1},
        pages = {1--32},
	year = {2024}
}

@article{Rao-NN-2024,
	title = {A sensory-motor theory of the neocortex},
 	journal = {Nature Neuroscience},
	publisher = {Springer Nature},
	author = {Rao, Rajesh P. N.},
        volume = {27},
        pages = {1221--1235},
	year = {2024}
}

@article{knill2004bayesian,
  title={The {Bayesian} brain: the role of uncertainty in neural coding and computation},
  author={Knill, David C and Pouget, Alexandre},
  journal={TRENDS in Neurosciences},
  volume={27},
  number={12},
  pages={712--719},
  year={2004},
  publisher={Elsevier}
}

@article{tomov2020discovery,
  title={Discovery of hierarchical representations for efficient planning},
  author={Tomov, Momchil S. and Yagati, Samyukta and Kumar, Agni and Yang, Wanqian and Gershman, Samuel J.},
  journal={PLoS computational biology},
  volume={16},
  number={4},
  pages={e1007594},
  year={2020},
  publisher={Public Library of Science San Francisco, CA USA}
}

@article{momennejad2017successor,
  title={The successor representation in human reinforcement learning},
  author={Momennejad, Ida and Russek, Evan M and Cheong, Jin H and Botvinick, Matthew M and Daw, Nathaniel Douglass and Gershman, Samuel J},
  journal={Nature human behaviour},
  volume={1},
  number={9},
  pages={680--692},
  year={2017},
  publisher={Nature Publishing Group UK London}
}

@article{sutton1999between,
  title={Between {MDPs and semi-MDPs}: A framework for temporal abstraction in reinforcement learning},
  author={Sutton, Richard S and Precup, Doina and Singh, Satinder},
  journal={Artificial intelligence},
  volume={112},
  number={1-2},
  pages={181--211},
  year={1999},
  publisher={Elsevier}
}

@inproceedings{Ororbia2023,
  title={Active Predictive Coding: Brain-Inspired Reinforcement Learning for Sparse Reward Robotic Control Problems},
  author={Ororbia, Alexander and Mali, Ankur},
  booktitle={2023 IEEE International Conference on Robotics and Automation (ICRA)},
  pages={3015--3021},
  year={2023}
}

@ARTICLE{Friston-dirichlet-2015,
  
AUTHOR={FitzGerald, Thomas H. B.  and Dolan, Raymond J.  and Friston, Karl },
         
TITLE={Dopamine, reward learning, and active inference},
        
JOURNAL={Frontiers in Computational Neuroscience},
        
VOLUME={9},

YEAR={2015},

URL={https://www.frontiersin.org/journals/computational-neuroscience/articles/10.3389/fncom.2015.00136},

DOI={10.3389/fncom.2015.00136},

ISSN={1662-5188}

 }

@article{Friston-et-al2021,
    author = {Friston, Karl and Da Costa, Lancelot and Hafner, Danijar and Hesp, Casper and Parr, Thomas},
    title = {Sophisticated Inference},
    journal = {Neural Computation},
    volume = {33},
    number = {3},
    pages = {713-763},
    year = {2021},
    month = {03},
    doi = {10.1162/neco_a_01351},
    url = {https://doi.org/10.1162/neco_a_01351}
}

@article{Neacsu-et-al2022,
    doi = {10.1371/journal.pone.0277199},
    author = {Neacsu, Victorita AND Mirza, M. Berk AND Adams, Rick A. AND Friston, Karl J.},
    journal = {PLOS ONE},
    publisher = {Public Library of Science},
    title = {Structure learning enhances concept formation in synthetic Active Inference agents},
    year = {2022},
    month = {11},
    volume = {17},
    url = {https://doi.org/10.1371/journal.pone.0277199},
    pages = {1-34},
    number = {11}
}

@article{Friston-et-al2025,
    author = {Friston, Karl J. and Salvatori, Tommaso and Isomura, Takuya and Tschantz, Alexander and Kiefer, Alex and Verbelen, Tim and Koudahl, Magnus and Paul, Aswin and Parr, Thomas and Razi, Adeel and Kagan, Brett J. and Buckley, Christopher L. and Ramstead, Maxwell J. D.},
    title = {Active Inference and Intentional Behavior},
    journal = {Neural Computation},
    volume = {37},
    number = {4},
    pages = {666-700},
    year = {2025},
    month = {3},
    issn = {0899-7667},
    doi = {10.1162/neco_a_01738},
    url = {https://doi.org/10.1162/neco_a_01738},
    eprint = {https://direct.mit.edu/neco/article-pdf/37/4/666/2506531/neco_a_01738.pdf}
}

@ARTICLE{Friston-et-al-2025a,
  
AUTHOR={Friston, Karl  and Heins, Conor  and Verbelen, Tim  and Da Costa, Lancelot  and Salvatori, Tommaso  and Markovic, Dimitrije  and Tschantz, Alexander  and Koudahl, Magnus  and Buckley, Christopher  and Parr, Thomas },
         
TITLE={From pixels to planning: scale-free active inference},
        
JOURNAL={Frontiers in Network Physiology},
        
VOLUME={5},

YEAR={2025},

URL={https://www.frontiersin.org/journals/network-physiology/articles/10.3389/fnetp.2025.1521963},

DOI={10.3389/fnetp.2025.1521963},

ISSN={2674-0109},

}

@article{isomura2018vitro,
  title={In vitro neural networks minimise variational free energy},
  author={Isomura, Takuya and Friston, Karl},
  journal={Scientific reports},
  volume={8},
  number={1},
  pages={16926},
  year={2018},
  publisher={Nature Publishing Group UK London}
}

@article{friston2017active,
  title={Active inference: a process theory},
  author={Friston, Karl and FitzGerald, Thomas and Rigoli, Francesco and Schwartenbeck, Philipp and Pezzulo, Giovanni},
  journal={Neural computation},
  volume={29},
  number={1},
  pages={1--49},
  year={2017},
  publisher={MIT Press One Rogers Street, Cambridge, MA 02142-1209, USA journals-info~…}
}

@article{barreto2017successor,
  title={Successor features for transfer in reinforcement learning},
  author={Barreto, Andr{\'e} and Dabney, Will and Munos, R{\'e}mi and Hunt, Jonathan J and Schaul, Tom and van Hasselt, Hado P and Silver, David},
  journal={Advances in neural information processing systems},
  volume={30},
  year={2017}
}

@inproceedings{li2006towards,
  title={Towards a unified theory of state abstraction for {MDPs}},
  author={Li, Lihong and Walsh, Thomas J and Littman, Michael L},
  booktitle={Proceedings of the International Symposium on Artificial Intelligence and Mathematics (ISAIM)},
  year={2006}
}

@article{proietti2023active,
  title={An active inference model of hierarchical action understanding, learning and imitation},
  author={Proietti, Riccardo and Pezzulo, Giovanni and Tessari, Alessia},
  journal={Physics of Life Reviews},
  volume={46},
  pages={92--118},
  year={2023},
  publisher={Elsevier}
}

@article{pezzulo2018hierarchical,
  title={Hierarchical active inference: a theory of motivated control},
  author={Pezzulo, Giovanni and Rigoli, Francesco and Friston, Karl J},
  journal={Trends in cognitive sciences},
  volume={22},
  number={4},
  pages={294--306},
  year={2018},
  publisher={Elsevier}
}

@article{van2023bridging,
  title={A hierarchical active inference model of spatial alternation tasks and the hippocampal-prefrontal circuit},
  author={Van de Maele, Toon and Dhoedt, Bart and Verbelen, Tim and Pezzulo, Giovanni},
  journal={Nature Communications},
  volume={15},
  pages={9876},
  year={2024},
  doi={10.1038/s41467-024-54257-3}
}

@misc{hafner2022deephierarchicalplanningpixels,
      title={Deep Hierarchical Planning from Pixels}, 
      author={Danijar Hafner and Kuang-Huei Lee and Ian Fischer and Pieter Abbeel},
      year={2022},
      eprint={2206.04114},
      archivePrefix={arXiv},
      primaryClass={cs.AI},
      url={https://arxiv.org/abs/2206.04114}, 
}

@misc{brockman2016openaigym,
      title={OpenAI Gym}, 
      author={Greg Brockman and Vicki Cheung and Ludwig Pettersson and Jonas Schneider and John Schulman and Jie Tang and Wojciech Zaremba},
      year={2016},
      eprint={1606.01540},
      archivePrefix={arXiv},
      primaryClass={cs.LG},
      url={https://arxiv.org/abs/1606.01540}, 
}

@misc{kingma2022autoencodingvariationalbayes,
      title={Auto-Encoding Variational Bayes},
      author={Diederik P Kingma and Max Welling},
      year={2013},
      eprint={1312.6114},
      archivePrefix={arXiv},
      primaryClass={stat.ML},
      url={https://arxiv.org/abs/1312.6114}, 
}

@article{whittington2020tolman,
  title={The {Tolman-Eichenbaum} machine: unifying space and relational memory through generalization in the hippocampal formation},
  author={Whittington, James CR and Muller, Timothy H and Mark, Shirley and Chen, Guifen and Barry, Caswell and Burgess, Neil and Behrens, Timothy EJ},
  journal={Cell},
  volume={183},
  number={5},
  pages={1249--1263},
  year={2020},
  publisher={Elsevier}
}

@article{gershman2018successor,
  title={The successor representation: its computational logic and neural substrates},
  author={Gershman, Samuel J},
  journal={Journal of Neuroscience},
  volume={38},
  number={33},
  pages={7193--7200},
  year={2018},
  publisher={Soc Neuroscience}
}

@article{momennejad2020learning,
  title={Learning structures: predictive representations, replay, and generalization},
  author={Momennejad, Ida},
  journal={Current Opinion in Behavioral Sciences},
  volume={32},
  pages={155--166},
  year={2020},
  publisher={Elsevier}
}

@article{stoewer2022neural,
  title={Neural network based successor representations to form cognitive maps of space and language},
  author={Stoewer, Paul and Schlieker, Christian and Schilling, Achim and Metzner, Claus and Maier, Andreas and Krauss, Patrick},
  journal={Scientific Reports},
  volume={12},
  number={1},
  pages={11233},
  year={2022},
  publisher={Nature Publishing Group UK London}
}

@article{millidge2020deep,
  title={Deep active inference as variational policy gradients},
  author={Millidge, Beren},
  journal={Journal of Mathematical Psychology},
  volume={96},
  pages={102348},
  year={2020},
  publisher={Elsevier}
}

@article{smith2022step,
  title={A step-by-step tutorial on active inference and its application to empirical data},
  author={Smith, Ryan and Friston, Karl J and Whyte, Christopher J},
  journal={Journal of mathematical psychology},
  volume={107},
  pages={102632},
  year={2022},
  publisher={Elsevier}
}

@article{russek2017predictive,
  title={Predictive representations can link model-based reinforcement learning to model-free mechanisms},
  author={Russek, Evan M and Momennejad, Ida and Botvinick, Matthew M and Gershman, Samuel J and Daw, Nathaniel D},
  journal={PLoS computational biology},
  volume={13},
  number={9},
  pages={e1005768},
  year={2017},
  publisher={Public Library of Science San Francisco, CA USA}
}

@book{sutton-barto-book,
  title={Reinforcement learning: An introduction},
  author={Sutton, Richard S and Barto, Andrew G},
  year={2018},
  publisher={MIT press}
}

@article{scikit-learn,
  author  = {Pedregosa, F. and Varoquaux, G. and Gramfort, A. and Michel, V. and Thirion, B. and Grisel, O. and Blondel, M. and Prettenhofer, P. and Weiss, R. and Dubourg, V. and Vanderplas, J. and Passos, A. and Cournapeau, D. and Brucher, M. and Perrot, M. and Duchesnay, E.},
  title   = {Scikit-learn: Machine Learning in {P}ython},
  journal = {Journal of Machine Learning Research},
  volume  = {12},
  pages   = {2825--2830},
  year    = {2011}
}

@misc{gymnasium_robotics,
  title = {Gymnasium-Robotics: {PointMaze} Environments},
  author = {{Farama Foundation}},
  year = {2023},
  url = {https://robotics.farama.org/envs/maze/point_maze/},
  note = {Accessed: 2024}
}

@inproceedings{kumar2011co,
  title={Co-regularized Multi-view Spectral Clustering},
  author={Kumar, Abhishek and Rai, Piyush and Daume, Hal},
  booktitle={Advances in Neural Information Processing Systems},
  volume={24},
  pages={1413--1421},
  year={2011}
}

@article{chao2021survey,
  title={A Survey on Multiview Clustering},
  author={Chao, Guoqing and Sun, Shiliang and Bi, Jinbo},
  journal={IEEE Transactions on Artificial Intelligence},
  volume={2},
  number={2},
  pages={146--168},
  year={2021},
  publisher={IEEE}
}
\end{document}